\documentclass[letterpaper]{article} 
\usepackage[preprint]{aaai2027}
\usepackage[hyphens]{url}  
\usepackage{graphicx} 
\usepackage{natbib}  
\usepackage{caption} 
\usepackage{algorithm}
\usepackage{algorithmic}
\usepackage{booktabs}
\usepackage{array}
\usepackage{multirow}
\usepackage{amsmath}
\usepackage{amssymb}
\usepackage{mathtools}
\usepackage{amsthm}
\usepackage{tikz}
\usetikzlibrary{arrows.meta,positioning,fit,calc,backgrounds,shapes.geometric}
\usepackage{pgfplots}
\pgfplotsset{compat=1.16}
\usepgfplotslibrary{groupplots}
\usepackage[capitalize,noabbrev]{cleveref}
\theoremstyle{plain}
\newtheorem{theorem}{Theorem}[section]
\newtheorem{proposition}[theorem]{Proposition}
\newtheorem{lemma}[theorem]{Lemma}
\newtheorem{corollary}[theorem]{Corollary}
\theoremstyle{definition}
\newtheorem{definition}[theorem]{Definition}

\theoremstyle{remark}

\newcommand{\R}{\mathbb{R}}
\newcommand{\E}{\mathbb{E}}
\newcommand{\TV}{\mathrm{TV}}
\newcommand{\KL}{\mathrm{KL}}
\newcommand{\kl}{\mathrm{kl}}
\newcommand{\Dref}{D_{\mathrm{ref}}}
\newcommand{\Dnew}{D_{\mathrm{new}}}
\newcommand{\pold}{\pi_{\mathrm{old}}}
\newcommand{\pnew}{\pi_{\mathrm{new}}}
\newcommand{\Fire}{\mathsf{F}}
\renewcommand{\lg}{\ell_{\mathrm{g}}}
\newcommand{\ind}{\mathbf{1}}
\newcommand{\method}{FireGate}
\newcommand{\methodp}{FireGate-prompt}
\newcommand{\methoda}{FireGate-anytime}

\newcommand{\resdir}{latex_v2_results}

\newcommand{\placeholder}[1]{\begin{center}\fbox{\parbox[c][1.6cm][c]{0.9\linewidth}{\centering\small
  \textcolor{blue!60!black}{#1}}}\end{center}}
\newcommand{\inputor}[2]{\IfFileExists{#1}{\input{#1}}{\placeholder{#2}}}
\makeatletter
\newcommand{\inputwide}[2]{\begingroup\renewenvironment{table}[1][]{\@dblfloat{table}[t]}{\end@dblfloat}%
  \inputor{#1}{#2}\endgroup}
\makeatother
\newcommand{\resActMin}{100.0}

\newcommand{\resAvgKl}{55.9}

\newcommand{\resAvgNull}{58.6}

\newcommand{\resAvgOlora}{59.8}

\newcommand{\resAvgRefkl}{47.4}

\newcommand{\resAvgReplay}{58.2}

\newcommand{\resAvgRouter}{57.1}

\newcommand{\resAvgScale}{54.9}

\newcommand{\resAvgSeq}{57.2}
\newcommand{\resAvgSeqSd}{4.9}
\newcommand{\resAvgTw}{59.1}
\newcommand{\resAvgTwSd}{0.9}
\newcommand{\resAvgTwp}{58.8}
\newcommand{\resAvgTwpSd}{1.6}
\newcommand{\resAvgTws}{57.8}

\newcommand{\resBwtSeq}{$-$4.5}

\newcommand{\resBwtTw}{+0.1}

\newcommand{\resBwtTwp}{+0.0}

\newcommand{\resCertAdv}{1.2}

\newcommand{\resCertDolly}{2.5}

\newcommand{\resCertMMLU}{12.2}

\newcommand{\resCertPAdv}{1.2}

\newcommand{\resCertPDolly}{5.7}

\newcommand{\resCertPMMLU}{10.5}

\newcommand{\resCertPXS}{0.7}

\newcommand{\resCertXS}{0.7}

\newcommand{\resChgKlDolly}{90.8}

\newcommand{\resChgKlMmlu}{33.5}

\newcommand{\resChgNullDolly}{74.2}

\newcommand{\resChgOloraDolly}{90.9}

\newcommand{\resChgOloraMmlu}{36.4}

\newcommand{\resChgRefklDolly}{94.4}

\newcommand{\resChgRefklMmlu}{38.7}

\newcommand{\resChgReplayDolly}{90.2}

\newcommand{\resChgReplayMmlu}{37.5}

\newcommand{\resChgRouterDolly}{0.0}

\newcommand{\resChgScaleDolly}{84.8}

\newcommand{\resChgSeqDolly}{93.4}

\newcommand{\resChgSeqMmlu}{38.8}

\newcommand{\resChgSeqXs}{65.8}

\newcommand{\resChgTwDolly}{1.0}

\newcommand{\resChgTwpDolly}{2.6}

\newcommand{\resChgTwsDolly}{2.2}

\newcommand{\resExactN}{19853}
\newcommand{\resExactPN}{18714}
\newcommand{\resExactPViol}{0}
\newcommand{\resExactViol}{0}

\newcommand{\resFireRouterDolly}{0.0}

\newcommand{\resFireRouterMmlu}{0.0}

\newcommand{\resFireTwDolly}{1.2}

\newcommand{\resFireTwMmlu}{9.8}

\newcommand{\resFireTwpDolly}{3.8}

\newcommand{\resFireTwpMmlu}{8.2}

\newcommand{\resRetain}{96}

\newcommand{\resSeeds}{3}

\newcommand{\resSeqLeak}{1.5}

\newcommand{\resTwLeak}{0.3}

\newcommand{\ovlKlAChg}{87.8}

\newcommand{\ovlKlAGain}{+4.9}

\newcommand{\ovlKlAParam}{0.01}

\newcommand{\ovlKlBChg}{73.3}

\newcommand{\ovlKlBGain}{+3.5}

\newcommand{\ovlKlCChg}{64.2}

\newcommand{\ovlKlCGain}{+1.9}

\newcommand{\ovlKlDAdv}{+0.0}

\newcommand{\ovlKlDChg}{56.0}

\newcommand{\ovlKlDGain}{+0.7}

\newcommand{\ovlKlDParam}{1}

\newcommand{\ovlNullAChg}{97.1}

\newcommand{\ovlNullAGain}{+4.8}

\newcommand{\ovlNullBAdv}{$-$9.1}

\newcommand{\ovlNullBChg}{98.8}

\newcommand{\ovlNullBGain}{+5.5}

\newcommand{\ovlNullCChg}{95.2}

\newcommand{\ovlNullrfAChg}{93.7}

\newcommand{\ovlNullrfAGain}{+5.3}

\newcommand{\ovlOracleGain}{+5.2}

\newcommand{\ovlRefklAChg}{82.9}

\newcommand{\ovlRefklAGain}{+5.5}

\newcommand{\ovlRefklBChg}{86.9}

\newcommand{\ovlRefklBGain}{+2.3}

\newcommand{\ovlRefklCAdv}{$-$0.3}

\newcommand{\ovlRefklCChg}{86.5}

\newcommand{\ovlRefklCGain}{+1.3}

\newcommand{\ovlRefklCMmlu}{+1.9}

\newcommand{\ovlRefklCXs}{$-$7.5}

\newcommand{\ovlRefklsAChg}{73.2}

\newcommand{\ovlRefklsAGain}{+0.6}

\newcommand{\ovlRefklsBGain}{+0.9}

\newcommand{\ovlRouterChg}{6.8}

\newcommand{\ovlRouterFireDolly}{6.8}
\newcommand{\ovlRouterFireNq}{97.4}
\newcommand{\ovlRouterGain}{+5.4}

\newcommand{\ovlRouterXs}{$-$5.6}

\newcommand{\ovlScaleBChg}{86.4}

\newcommand{\ovlScaleBGain}{+4.6}

\newcommand{\ovlSeeds}{3}
\newcommand{\ovlSeqAdv}{$-$8.8}

\newcommand{\ovlSeqChg}{94.2}

\newcommand{\ovlSeqGain}{+5.2}

\newcommand{\ovlSeqLeak}{+0.0}
\newcommand{\ovlSeqMmlu}{+0.2}
\newcommand{\ovlSeqXs}{$-$50.3}

\newcommand{\ovlTwAAuroc}{0.83}

\newcommand{\ovlTwACertMax}{3.9}
\newcommand{\ovlTwAChg}{0.9}

\newcommand{\ovlTwAFireNq}{11.7}
\newcommand{\ovlTwAGain}{+0.3}

\newcommand{\ovlTwAParam}{0.03}

\newcommand{\ovlTwBCertMax}{11.4}
\newcommand{\ovlTwBChg}{6.4}

\newcommand{\ovlTwBCondGain}{+3.0}

\newcommand{\ovlTwBFireNq}{39.8}
\newcommand{\ovlTwBGain}{+1.2}

\newcommand{\ovlTwBInPrompt}{67.7}

\newcommand{\ovlTwBParam}{0.1}

\newcommand{\ovlTwCCertMax}{30.9}
\newcommand{\ovlTwCChg}{23.3}

\newcommand{\ovlTwCCondGain}{+4.0}

\newcommand{\ovlTwCFireNq}{78.3}
\newcommand{\ovlTwCGain}{+3.1}

\newcommand{\ovlTwCParam}{0.3}

\newcommand{\ovlTwpAAuroc}{0.99}

\newcommand{\ovlTwpACertMax}{5.4}
\newcommand{\ovlTwpAChg}{3.5}

\newcommand{\ovlTwpAFireNq}{90.8}
\newcommand{\ovlTwpAGain}{+4.0}

\newcommand{\ovlTwpAParam}{0.03}
\newcommand{\ovlTwpAXs}{$-$0.8}

\newcommand{\ovlTwpBCertMax}{12.3}
\newcommand{\ovlTwpBChg}{9.4}

\newcommand{\ovlTwpBCondGain}{+5.4}

\newcommand{\ovlTwpBFireNq}{97.8}
\newcommand{\ovlTwpBGain}{+5.3}

\newcommand{\ovlTwpBParam}{0.1}
\newcommand{\ovlTwpBXs}{$-$11.9}

\newcommand{\ovlTwpCCertMax}{33.0}
\newcommand{\ovlTwpCChg}{28.9}

\newcommand{\ovlTwpCFireNq}{99.9}
\newcommand{\ovlTwpCGain}{+5.9}

\newcommand{\ovlTwpCParam}{0.3}
\newcommand{\ovlTwpCXs}{$-$34.8}

\newcommand{\resXSampChgSeq}{95.8}

\newcommand{\resXSampNonfired}{18543}
\newcommand{\resXSampViol}{0}

\newcommand{\resXSysFireTwp}{0.8}

\newcommand{\resXTurnFireTwp}{1.5}

\newcommand{\resXWrapFireTw}{75.0}
\newcommand{\resXWrapFireTwp}{100.0}

\newcommand{\resXWrapRefSeq}{$-$36.0}
\newcommand{\resXWrapRefTw}{$-$35.0}
\newcommand{\resXWrapRefTwp}{$-$16.1}

\newcommand{\ovlConcisectxSeqAdv}{$-$11.5}

\newcommand{\ovlConcisectxSeqChg}{96.1}

\newcommand{\ovlConcisectxSeqGain}{+65.2}

\newcommand{\ovlConcisectxSeqMmlu}{$-$3.5}
\newcommand{\ovlConcisectxSeqXs}{$-$51.2}

\newcommand{\ovlConcisectxTwpAAuroc}{0.84}

\newcommand{\ovlConcisectxTwpAFireNq}{3.4}
\newcommand{\ovlConcisectxTwpAGain}{+1.5}

\newcommand{\inputif}[1]{\IfFileExists{#1}{\input{#1}}{}}

\newcommand{\anNqRatioCalNDolly}{11/700}

\newcommand{\anNqRatioHeldNDolly}{13/400}

\newcommand{\anNqRatioPexchDolly}{0.06}

\newcommand{\anNqRatioPfisherDolly}{0.09}

\newcommand{\anNqSweepMae}{0.3}
\newcommand{\anNqSweepMax}{0.7}

\newcommand{\rvNqDiffRouterVsTwp}{+1.4}
\newcommand{\rvNqDiffRouterVsTwpCi}{[$-$0.9, +3.9]}

\newcommand{\rvNqDiffSeqswVsSeq}{$-$0.5}
\newcommand{\rvNqDiffSeqswVsSeqCi}{[$-$1.6, +0.4]}

\newcommand{\rvNqDiffTwpVsSeq}{$-$1.2}
\newcommand{\rvNqDiffTwpVsSeqCi}{[$-$3.5, +1.1]}

\newcommand{\rvNqRcFull}{97.2}

\newcommand{\rvNqRcSmallQ}{83.1}
\newcommand{\rvNqRcSmallSkew}{79.3}
\newcommand{\rvNqRdDollyAbove}{55.0}
\newcommand{\rvNqRdDollyMax}{4.8}
\newcommand{\rvNqRdDollyMean}{3.1}
\newcommand{\rvNqRdGainMean}{+4.8}
\newcommand{\rvNqRdGainSd}{0.1}
\newcommand{\rvNqRdN}{20}
\newcommand{\rvNqRdPowerMean}{92.2}
\newcommand{\rvNqRdPowerSd}{1.6}

\newcommand{\anIfcSweepMax}{4.1}

\newcommand{\anConcisectxSweepMax}{3.3}

\newcommand{\anSevenSweepMae}{0.4}

\newcommand{\anOlmoSweepMae}{0.3}

\newcommand{\ovgConcisectxRatioACpDolly}{3.3}

\newcommand{\ovgConcisectxRatioACpMmlu}{0.6}

\newcommand{\ovgConcisectxRatioADolly}{1.8}

\newcommand{\ovgConcisectxRatioAMmlu}{0.0}
\newcommand{\ovgConcisectxRatioAPower}{3.8}

\newcommand{\ovgConcisectxTwoACpDolly}{2.9}

\newcommand{\ovgConcisectxTwoACpMmlu}{0.6}

\newcommand{\ovgConcisectxTwoADolly}{1.5}

\newcommand{\ovgConcisectxTwoAMmlu}{0.0}
\newcommand{\ovgConcisectxTwoAPower}{3.7}

\newcommand{\ovgIfcRatioACpDolly}{2.9}

\newcommand{\ovgIfcRatioACpMmlu}{99.8}

\newcommand{\ovgIfcRatioADolly}{1.5}
\newcommand{\ovgIfcRatioAJb}{34.0}

\newcommand{\ovgIfcRatioAMmlu}{99.4}
\newcommand{\ovgIfcRatioAPower}{100.0}

\newcommand{\ovgIfcTwoACpDolly}{0.7}

\newcommand{\ovgIfcTwoACpMmlu}{0.6}

\newcommand{\ovgIfcTwoADolly}{0.0}

\newcommand{\ovgIfcTwoAMmlu}{0.0}
\newcommand{\ovgIfcTwoAPower}{98.0}

\newcommand{\ovgIfcsevenRatioACpDolly}{3.3}

\newcommand{\ovgIfcsevenRatioACpMmlu}{100.0}

\newcommand{\ovgIfcsevenRatioADolly}{1.8}
\newcommand{\ovgIfcsevenRatioAJb}{49.7}

\newcommand{\ovgIfcsevenRatioAMmlu}{100.0}
\newcommand{\ovgIfcsevenRatioAPower}{100.0}

\newcommand{\ovgIfcsevenTwoACpDolly}{0.7}

\newcommand{\ovgIfcsevenTwoACpMmlu}{0.6}

\newcommand{\ovgIfcsevenTwoADolly}{0.0}

\newcommand{\ovgIfcsevenTwoAMmlu}{0.0}
\newcommand{\ovgIfcsevenTwoAPower}{98.0}

\newcommand{\ovgNqRatioACpDolly}{5.1}

\newcommand{\ovgNqRatioACpMmlu}{0.6}

\newcommand{\ovgNqRatioADolly}{3.2}

\newcommand{\ovgNqRatioAMmlu}{0.0}
\newcommand{\ovgNqRatioAPower}{92.2}

\newcommand{\ovgNqTwoACpDolly}{4.5}

\newcommand{\ovgNqTwoACpMmlu}{0.6}

\newcommand{\ovgNqTwoADolly}{2.8}

\newcommand{\ovgNqTwoAMmlu}{0.0}
\newcommand{\ovgNqTwoAPower}{90.7}

\newcommand{\ovgNqolmoRatioACpDolly}{3.9}

\newcommand{\ovgNqolmoRatioACpMmlu}{0.6}

\newcommand{\ovgNqolmoRatioADolly}{2.2}

\newcommand{\ovgNqolmoRatioAMmlu}{0.0}
\newcommand{\ovgNqolmoRatioAPower}{96.8}

\newcommand{\ovgNqolmoTwoACpDolly}{3.9}

\newcommand{\ovgNqolmoTwoACpMmlu}{0.6}

\newcommand{\ovgNqolmoTwoADolly}{2.2}

\newcommand{\ovgNqolmoTwoAMmlu}{0.0}
\newcommand{\ovgNqolmoTwoAPower}{95.0}

\newcommand{\ovgNqsevenRatioACpDolly}{4.5}

\newcommand{\ovgNqsevenRatioACpMmlu}{0.6}

\newcommand{\ovgNqsevenRatioADolly}{2.8}

\newcommand{\ovgNqsevenRatioAMmlu}{0.0}
\newcommand{\ovgNqsevenRatioAPower}{97.9}

\newcommand{\ovgNqsevenTwoACpDolly}{3.6}

\newcommand{\ovgNqsevenTwoACpMmlu}{0.6}

\newcommand{\ovgNqsevenTwoADolly}{2.0}

\newcommand{\ovgNqsevenTwoAMmlu}{0.0}
\newcommand{\ovgNqsevenTwoAPower}{96.1}

\newcommand{\ovgStreamRatioACpDolly}{6.0}

\newcommand{\ovgStreamRatioACpMmlu}{10.9}

\newcommand{\ovgStreamRatioADolly}{4.0}

\newcommand{\ovgStreamRatioAMmlu}{8.6}
\newcommand{\ovgStreamRatioAPower}{100.0}

\newcommand{\ovgStreamTwoACpDolly}{0.7}

\newcommand{\ovgStreamTwoACpMmlu}{0.6}

\newcommand{\ovgStreamTwoADolly}{0.0}

\newcommand{\ovgStreamTwoAMmlu}{0.0}
\newcommand{\ovgStreamTwoAPower}{98.0}

\newcommand{\ovgStreamfiveRatioACpDolly}{8.3}

\newcommand{\ovgStreamfiveRatioACpMmlu}{53.8}

\newcommand{\ovgStreamfiveRatioADolly}{6.0}

\newcommand{\ovgStreamfiveRatioAMmlu}{50.0}
\newcommand{\ovgStreamfiveRatioAPower}{100.0}

\newcommand{\ovgStreamfiveTwoACpDolly}{0.7}

\newcommand{\ovgStreamfiveTwoACpMmlu}{0.6}

\newcommand{\ovgStreamfiveTwoADolly}{0.0}

\newcommand{\ovgStreamfiveTwoAMmlu}{0.0}
\newcommand{\ovgStreamfiveTwoAPower}{98.0}

\newcommand{\ovgStreamsevenRatioACpDolly}{4.8}

\newcommand{\ovgStreamsevenRatioACpMmlu}{42.7}

\newcommand{\ovgStreamsevenRatioADolly}{3.0}

\newcommand{\ovgStreamsevenRatioAMmlu}{39.0}
\newcommand{\ovgStreamsevenRatioAPower}{100.0}

\newcommand{\ovgStreamsevenTwoACpDolly}{0.7}

\newcommand{\ovgStreamsevenTwoACpMmlu}{0.6}

\newcommand{\ovgStreamsevenTwoADolly}{0.0}

\newcommand{\ovgStreamsevenTwoAMmlu}{0.0}
\newcommand{\ovgStreamsevenTwoAPower}{98.0}

\newcommand{\ovwConcisectxSeqOwnGain}{+65.2}

\newcommand{\ovwConcisectxSeqRatioAGain}{+2.6}

\newcommand{\ovwConcisectxSeqRatioBChg}{7.3}
\newcommand{\ovwConcisectxSeqRatioBGain}{+17.7}

\newcommand{\ovwConcisectxSeqRatioCChg}{26.8}
\newcommand{\ovwConcisectxSeqRatioCGain}{+54.4}

\newcommand{\rvbAllCovErrMax}{0.5}
\newcommand{\rvbAllCovN}{92}
\newcommand{\rvbAllCovZero}{91}

\newcommand{\decNonfired}{0}
\newcommand{\decNonfiredChecked}{13{,}947}

\providecommand{\decNonfired}{--}\providecommand{\decNonfiredChecked}{--}

\providecommand{\rvbAllCovN}{92}\providecommand{\rvbAllCovZero}{91}\providecommand{\rvbAllCovErrMax}{0.5}

\makeatletter
\def\tw@inputprefixed#1#2{\begingroup
  \def\newcommand##1##2{\expandafter\xdef\csname #1\expandafter\@gobble\string##1\endcsname{\unexpanded{##2}}}%
  \input{#2}\endgroup}
\def\tw@inputrelabel#1#2{\begingroup\let\origlabel\label\def\label##1{\origlabel{#1}}\input{#2}\endgroup}
\newcommand{\inputprefixed}[2]{\IfFileExists{#2}{\tw@inputprefixed{#1}{#2}}{}}
\newcommand{\inputrelabel}[3]{\IfFileExists{#2}{\tw@inputrelabel{#1}{#2}}{\placeholder{#3}}}
\makeatother
\expandafter\gdef\csname sevenresActMin\endcsname{100.0}
\expandafter\gdef\csname sevenresActMinSd\endcsname{0.0}
\expandafter\gdef\csname sevenresAvgAgem\endcsname{--}
\expandafter\gdef\csname sevenresAvgEwc\endcsname{--}
\expandafter\gdef\csname sevenresAvgKl\endcsname{77.8}
\expandafter\gdef\csname sevenresAvgKlSd\endcsname{0.0}
\expandafter\gdef\csname sevenresAvgLtwo\endcsname{--}
\expandafter\gdef\csname sevenresAvgMas\endcsname{--}
\expandafter\gdef\csname sevenresAvgNull\endcsname{--}
\expandafter\gdef\csname sevenresAvgOlora\endcsname{--}
\expandafter\gdef\csname sevenresAvgRefkl\endcsname{72.9}
\expandafter\gdef\csname sevenresAvgRefklSd\endcsname{1.0}
\expandafter\gdef\csname sevenresAvgReplay\endcsname{--}
\expandafter\gdef\csname sevenresAvgRouter\endcsname{75.6}
\expandafter\gdef\csname sevenresAvgRouterSd\endcsname{0.0}
\expandafter\gdef\csname sevenresAvgScale\endcsname{--}
\expandafter\gdef\csname sevenresAvgSeq\endcsname{79.9}
\expandafter\gdef\csname sevenresAvgSeqSd\endcsname{2.7}
\expandafter\gdef\csname sevenresAvgTw\endcsname{78.7}
\expandafter\gdef\csname sevenresAvgTwSd\endcsname{0.0}
\expandafter\gdef\csname sevenresAvgTwp\endcsname{79.6}
\expandafter\gdef\csname sevenresAvgTwpSd\endcsname{0.8}
\expandafter\gdef\csname sevenresAvgTws\endcsname{70.7}
\expandafter\gdef\csname sevenresAvgTwsSd\endcsname{15.3}
\expandafter\gdef\csname sevenresBwtAgem\endcsname{--}
\expandafter\gdef\csname sevenresBwtEwc\endcsname{--}
\expandafter\gdef\csname sevenresBwtKl\endcsname{$-$1.0}
\expandafter\gdef\csname sevenresBwtKlSd\endcsname{0.0}
\expandafter\gdef\csname sevenresBwtLtwo\endcsname{--}
\expandafter\gdef\csname sevenresBwtMas\endcsname{--}
\expandafter\gdef\csname sevenresBwtNull\endcsname{--}
\expandafter\gdef\csname sevenresBwtOlora\endcsname{--}
\expandafter\gdef\csname sevenresBwtRefkl\endcsname{$-$2.6}
\expandafter\gdef\csname sevenresBwtRefklSd\endcsname{1.8}
\expandafter\gdef\csname sevenresBwtReplay\endcsname{--}
\expandafter\gdef\csname sevenresBwtRouter\endcsname{+0.0}
\expandafter\gdef\csname sevenresBwtRouterSd\endcsname{0.0}
\expandafter\gdef\csname sevenresBwtScale\endcsname{--}
\expandafter\gdef\csname sevenresBwtSeq\endcsname{+0.7}
\expandafter\gdef\csname sevenresBwtSeqSd\endcsname{1.2}
\expandafter\gdef\csname sevenresBwtTw\endcsname{+0.0}
\expandafter\gdef\csname sevenresBwtTwSd\endcsname{0.0}
\expandafter\gdef\csname sevenresBwtTwp\endcsname{+0.0}
\expandafter\gdef\csname sevenresBwtTwpSd\endcsname{0.1}
\expandafter\gdef\csname sevenresBwtTws\endcsname{+0.1}
\expandafter\gdef\csname sevenresBwtTwsSd\endcsname{0.1}
\expandafter\gdef\csname sevenresCertAdv\endcsname{1.2}
\expandafter\gdef\csname sevenresCertAdvSd\endcsname{0.0}
\expandafter\gdef\csname sevenresCertDolly\endcsname{2.6}
\expandafter\gdef\csname sevenresCertDollySd\endcsname{0.0}
\expandafter\gdef\csname sevenresCertMMLU\endcsname{8.5}
\expandafter\gdef\csname sevenresCertMMLUSd\endcsname{0.0}
\expandafter\gdef\csname sevenresCertPAdv\endcsname{1.2}
\expandafter\gdef\csname sevenresCertPAdvSd\endcsname{0.0}
\expandafter\gdef\csname sevenresCertPDolly\endcsname{4.5}
\expandafter\gdef\csname sevenresCertPDollySd\endcsname{0.0}
\expandafter\gdef\csname sevenresCertPMMLU\endcsname{42.3}
\expandafter\gdef\csname sevenresCertPMMLUSd\endcsname{0.0}
\expandafter\gdef\csname sevenresCertPXS\endcsname{0.7}
\expandafter\gdef\csname sevenresCertPXSSd\endcsname{0.0}
\expandafter\gdef\csname sevenresCertXS\endcsname{0.7}
\expandafter\gdef\csname sevenresCertXSSd\endcsname{0.0}
\expandafter\gdef\csname sevenresChgAgemAdv\endcsname{--}
\expandafter\gdef\csname sevenresChgAgemDolly\endcsname{--}
\expandafter\gdef\csname sevenresChgAgemMmlu\endcsname{--}
\expandafter\gdef\csname sevenresChgAgemXs\endcsname{--}
\expandafter\gdef\csname sevenresChgEwcAdv\endcsname{--}
\expandafter\gdef\csname sevenresChgEwcDolly\endcsname{--}
\expandafter\gdef\csname sevenresChgEwcMmlu\endcsname{--}
\expandafter\gdef\csname sevenresChgEwcXs\endcsname{--}
\expandafter\gdef\csname sevenresChgKlAdv\endcsname{96.0}
\expandafter\gdef\csname sevenresChgKlAdvSd\endcsname{0.0}
\expandafter\gdef\csname sevenresChgKlDolly\endcsname{88.5}
\expandafter\gdef\csname sevenresChgKlDollySd\endcsname{0.0}
\expandafter\gdef\csname sevenresChgKlMmlu\endcsname{1.4}
\expandafter\gdef\csname sevenresChgKlMmluSd\endcsname{0.0}
\expandafter\gdef\csname sevenresChgKlXs\endcsname{97.8}
\expandafter\gdef\csname sevenresChgKlXsSd\endcsname{0.0}
\expandafter\gdef\csname sevenresChgLtwoAdv\endcsname{--}
\expandafter\gdef\csname sevenresChgLtwoDolly\endcsname{--}
\expandafter\gdef\csname sevenresChgLtwoMmlu\endcsname{--}
\expandafter\gdef\csname sevenresChgLtwoXs\endcsname{--}
\expandafter\gdef\csname sevenresChgMasAdv\endcsname{--}
\expandafter\gdef\csname sevenresChgMasDolly\endcsname{--}
\expandafter\gdef\csname sevenresChgMasMmlu\endcsname{--}
\expandafter\gdef\csname sevenresChgMasXs\endcsname{--}
\expandafter\gdef\csname sevenresChgNullAdv\endcsname{--}
\expandafter\gdef\csname sevenresChgNullDolly\endcsname{--}
\expandafter\gdef\csname sevenresChgNullMmlu\endcsname{--}
\expandafter\gdef\csname sevenresChgNullXs\endcsname{--}
\expandafter\gdef\csname sevenresChgOloraAdv\endcsname{--}
\expandafter\gdef\csname sevenresChgOloraDolly\endcsname{--}
\expandafter\gdef\csname sevenresChgOloraMmlu\endcsname{--}
\expandafter\gdef\csname sevenresChgOloraXs\endcsname{--}
\expandafter\gdef\csname sevenresChgRefklAdv\endcsname{100.0}
\expandafter\gdef\csname sevenresChgRefklAdvSd\endcsname{0.0}
\expandafter\gdef\csname sevenresChgRefklDolly\endcsname{95.8}
\expandafter\gdef\csname sevenresChgRefklDollySd\endcsname{0.3}
\expandafter\gdef\csname sevenresChgRefklMmlu\endcsname{6.4}
\expandafter\gdef\csname sevenresChgRefklMmluSd\endcsname{1.0}
\expandafter\gdef\csname sevenresChgRefklXs\endcsname{99.5}
\expandafter\gdef\csname sevenresChgRefklXsSd\endcsname{0.2}
\expandafter\gdef\csname sevenresChgReplayAdv\endcsname{--}
\expandafter\gdef\csname sevenresChgReplayDolly\endcsname{--}
\expandafter\gdef\csname sevenresChgReplayMmlu\endcsname{--}
\expandafter\gdef\csname sevenresChgReplayXs\endcsname{--}
\expandafter\gdef\csname sevenresChgRouterAdv\endcsname{0.0}
\expandafter\gdef\csname sevenresChgRouterAdvSd\endcsname{0.0}
\expandafter\gdef\csname sevenresChgRouterDolly\endcsname{0.0}
\expandafter\gdef\csname sevenresChgRouterDollySd\endcsname{0.0}
\expandafter\gdef\csname sevenresChgRouterMmlu\endcsname{0.0}
\expandafter\gdef\csname sevenresChgRouterMmluSd\endcsname{0.0}
\expandafter\gdef\csname sevenresChgRouterXs\endcsname{0.0}
\expandafter\gdef\csname sevenresChgRouterXsSd\endcsname{0.0}
\expandafter\gdef\csname sevenresChgScaleAdv\endcsname{--}
\expandafter\gdef\csname sevenresChgScaleDolly\endcsname{--}
\expandafter\gdef\csname sevenresChgScaleMmlu\endcsname{--}
\expandafter\gdef\csname sevenresChgScaleXs\endcsname{--}
\expandafter\gdef\csname sevenresChgSeqAdv\endcsname{99.3}
\expandafter\gdef\csname sevenresChgSeqAdvSd\endcsname{0.8}
\expandafter\gdef\csname sevenresChgSeqDolly\endcsname{93.5}
\expandafter\gdef\csname sevenresChgSeqDollySd\endcsname{1.5}
\expandafter\gdef\csname sevenresChgSeqMmlu\endcsname{3.2}
\expandafter\gdef\csname sevenresChgSeqMmluSd\endcsname{1.7}
\expandafter\gdef\csname sevenresChgSeqXs\endcsname{98.6}
\expandafter\gdef\csname sevenresChgSeqXsSd\endcsname{0.6}
\expandafter\gdef\csname sevenresChgTwAdv\endcsname{0.0}
\expandafter\gdef\csname sevenresChgTwAdvSd\endcsname{0.0}
\expandafter\gdef\csname sevenresChgTwDolly\endcsname{0.8}
\expandafter\gdef\csname sevenresChgTwDollySd\endcsname{0.0}
\expandafter\gdef\csname sevenresChgTwMmlu\endcsname{0.0}
\expandafter\gdef\csname sevenresChgTwMmluSd\endcsname{0.0}
\expandafter\gdef\csname sevenresChgTwXs\endcsname{0.0}
\expandafter\gdef\csname sevenresChgTwXsSd\endcsname{0.0}
\expandafter\gdef\csname sevenresChgTwpAdv\endcsname{0.0}
\expandafter\gdef\csname sevenresChgTwpAdvSd\endcsname{0.0}
\expandafter\gdef\csname sevenresChgTwpDolly\endcsname{2.2}
\expandafter\gdef\csname sevenresChgTwpDollySd\endcsname{0.1}
\expandafter\gdef\csname sevenresChgTwpMmlu\endcsname{0.1}
\expandafter\gdef\csname sevenresChgTwpMmluSd\endcsname{0.1}
\expandafter\gdef\csname sevenresChgTwpXs\endcsname{0.0}
\expandafter\gdef\csname sevenresChgTwpXsSd\endcsname{0.0}
\expandafter\gdef\csname sevenresChgTwsAdv\endcsname{0.0}
\expandafter\gdef\csname sevenresChgTwsAdvSd\endcsname{0.0}
\expandafter\gdef\csname sevenresChgTwsDolly\endcsname{2.5}
\expandafter\gdef\csname sevenresChgTwsDollySd\endcsname{0.3}
\expandafter\gdef\csname sevenresChgTwsMmlu\endcsname{1.5}
\expandafter\gdef\csname sevenresChgTwsMmluSd\endcsname{0.6}
\expandafter\gdef\csname sevenresChgTwsXs\endcsname{0.0}
\expandafter\gdef\csname sevenresChgTwsXsSd\endcsname{0.0}
\expandafter\gdef\csname sevenresExactN\endcsname{6698}
\expandafter\gdef\csname sevenresExactPN\endcsname{18150}
\expandafter\gdef\csname sevenresExactPViol\endcsname{0}
\expandafter\gdef\csname sevenresExactViol\endcsname{0}
\expandafter\gdef\csname sevenresFireRouterAdv\endcsname{0.0}
\expandafter\gdef\csname sevenresFireRouterAdvSd\endcsname{0.0}
\expandafter\gdef\csname sevenresFireRouterDolly\endcsname{0.0}
\expandafter\gdef\csname sevenresFireRouterDollySd\endcsname{0.0}
\expandafter\gdef\csname sevenresFireRouterMmlu\endcsname{0.0}
\expandafter\gdef\csname sevenresFireRouterMmluSd\endcsname{0.0}
\expandafter\gdef\csname sevenresFireRouterXs\endcsname{0.0}
\expandafter\gdef\csname sevenresFireRouterXsSd\endcsname{0.0}
\expandafter\gdef\csname sevenresFireTwAdv\endcsname{0.0}
\expandafter\gdef\csname sevenresFireTwAdvSd\endcsname{0.0}
\expandafter\gdef\csname sevenresFireTwDolly\endcsname{1.2}
\expandafter\gdef\csname sevenresFireTwDollySd\endcsname{0.0}
\expandafter\gdef\csname sevenresFireTwMmlu\endcsname{6.4}
\expandafter\gdef\csname sevenresFireTwMmluSd\endcsname{0.0}
\expandafter\gdef\csname sevenresFireTwXs\endcsname{0.0}
\expandafter\gdef\csname sevenresFireTwXsSd\endcsname{0.0}
\expandafter\gdef\csname sevenresFireTwpAdv\endcsname{0.0}
\expandafter\gdef\csname sevenresFireTwpAdvSd\endcsname{0.0}
\expandafter\gdef\csname sevenresFireTwpDolly\endcsname{2.8}
\expandafter\gdef\csname sevenresFireTwpDollySd\endcsname{0.0}
\expandafter\gdef\csname sevenresFireTwpMmlu\endcsname{38.6}
\expandafter\gdef\csname sevenresFireTwpMmluSd\endcsname{0.0}
\expandafter\gdef\csname sevenresFireTwpXs\endcsname{0.0}
\expandafter\gdef\csname sevenresFireTwpXsSd\endcsname{0.0}
\expandafter\gdef\csname sevenresFireTwsAdv\endcsname{0.0}
\expandafter\gdef\csname sevenresFireTwsAdvSd\endcsname{0.0}
\expandafter\gdef\csname sevenresFireTwsDolly\endcsname{2.8}
\expandafter\gdef\csname sevenresFireTwsDollySd\endcsname{0.0}
\expandafter\gdef\csname sevenresFireTwsMmlu\endcsname{38.6}
\expandafter\gdef\csname sevenresFireTwsMmluSd\endcsname{0.0}
\expandafter\gdef\csname sevenresFireTwsXs\endcsname{0.0}
\expandafter\gdef\csname sevenresFireTwsXsSd\endcsname{0.0}
\expandafter\gdef\csname sevenresRetain\endcsname{89}
\expandafter\gdef\csname sevenresRetainSd\endcsname{0.0}
\expandafter\gdef\csname sevenresSeeds\endcsname{3}
\expandafter\gdef\csname sevenresSeqAdvDelta\endcsname{$-$0.3}
\expandafter\gdef\csname sevenresSeqAdvDeltaSd\endcsname{0.8}
\expandafter\gdef\csname sevenresSeqChangeDolly\endcsname{93.5}
\expandafter\gdef\csname sevenresSeqChangeDollySd\endcsname{1.5}
\expandafter\gdef\csname sevenresSeqLeak\endcsname{0.2}
\expandafter\gdef\csname sevenresSeqLeakSd\endcsname{0.3}
\expandafter\gdef\csname sevenresTwLeak\endcsname{0.0}
\expandafter\gdef\csname sevenresTwLeakSd\endcsname{0.0}
\expandafter\gdef\csname sevenresXAdvSysRefSeq\endcsname{+0.0}
\expandafter\gdef\csname sevenresXAdvSysRefTw\endcsname{+0.0}
\expandafter\gdef\csname sevenresXSampChgSeq\endcsname{93.4}
\expandafter\gdef\csname sevenresXSampChgTw\endcsname{0.4}
\expandafter\gdef\csname sevenresXSampNonfired\endcsname{3130}
\expandafter\gdef\csname sevenresXSampViol\endcsname{0}
\expandafter\gdef\csname sevenresXSysChgSeq\endcsname{82.5}
\expandafter\gdef\csname sevenresXSysChgTw\endcsname{1.0}
\expandafter\gdef\csname sevenresXSysFireSeq\endcsname{0.0}
\expandafter\gdef\csname sevenresXSysFireTw\endcsname{1.0}
\expandafter\gdef\csname sevenresXTurnChgSeq\endcsname{93.8}
\expandafter\gdef\csname sevenresXTurnChgTw\endcsname{0.5}
\expandafter\gdef\csname sevenresXTurnFireSeq\endcsname{0.0}
\expandafter\gdef\csname sevenresXTurnFireTw\endcsname{1.0}
\expandafter\gdef\csname sevenresXWrapFireSeq\endcsname{0.0}
\expandafter\gdef\csname sevenresXWrapFireTw\endcsname{66.9}
\expandafter\gdef\csname sevenresXWrapRefMinSeq\endcsname{$-$4.0}
\expandafter\gdef\csname sevenresXWrapRefMinTw\endcsname{$-$1.2}
\expandafter\gdef\csname sevenresXWrapRefSeq\endcsname{+2.1}
\expandafter\gdef\csname sevenresXWrapRefTw\endcsname{$-$0.7}
\expandafter\gdef\csname fiveresActMin\endcsname{--}
\expandafter\gdef\csname fiveresAvgAgem\endcsname{--}
\expandafter\gdef\csname fiveresAvgEwc\endcsname{--}
\expandafter\gdef\csname fiveresAvgKl\endcsname{--}
\expandafter\gdef\csname fiveresAvgLtwo\endcsname{--}
\expandafter\gdef\csname fiveresAvgMas\endcsname{--}
\expandafter\gdef\csname fiveresAvgNull\endcsname{--}
\expandafter\gdef\csname fiveresAvgOlora\endcsname{--}
\expandafter\gdef\csname fiveresAvgRefkl\endcsname{--}
\expandafter\gdef\csname fiveresAvgReplay\endcsname{--}
\expandafter\gdef\csname fiveresAvgRouter\endcsname{--}
\expandafter\gdef\csname fiveresAvgScale\endcsname{--}
\expandafter\gdef\csname fiveresAvgSeq\endcsname{46.5}
\expandafter\gdef\csname fiveresAvgSeqSd\endcsname{1.7}
\expandafter\gdef\csname fiveresAvgTw\endcsname{--}
\expandafter\gdef\csname fiveresAvgTwp\endcsname{49.6}
\expandafter\gdef\csname fiveresAvgTwpSd\endcsname{2.9}
\expandafter\gdef\csname fiveresAvgTws\endcsname{50.0}
\expandafter\gdef\csname fiveresAvgTwsSd\endcsname{1.1}
\expandafter\gdef\csname fiveresBwtAgem\endcsname{--}
\expandafter\gdef\csname fiveresBwtEwc\endcsname{--}
\expandafter\gdef\csname fiveresBwtKl\endcsname{--}
\expandafter\gdef\csname fiveresBwtLtwo\endcsname{--}
\expandafter\gdef\csname fiveresBwtMas\endcsname{--}
\expandafter\gdef\csname fiveresBwtNull\endcsname{--}
\expandafter\gdef\csname fiveresBwtOlora\endcsname{--}
\expandafter\gdef\csname fiveresBwtRefkl\endcsname{--}
\expandafter\gdef\csname fiveresBwtReplay\endcsname{--}
\expandafter\gdef\csname fiveresBwtRouter\endcsname{--}
\expandafter\gdef\csname fiveresBwtScale\endcsname{--}
\expandafter\gdef\csname fiveresBwtSeq\endcsname{$-$3.5}
\expandafter\gdef\csname fiveresBwtSeqSd\endcsname{0.7}
\expandafter\gdef\csname fiveresBwtTw\endcsname{--}
\expandafter\gdef\csname fiveresBwtTwp\endcsname{+0.0}
\expandafter\gdef\csname fiveresBwtTwpSd\endcsname{0.0}
\expandafter\gdef\csname fiveresBwtTws\endcsname{+0.0}
\expandafter\gdef\csname fiveresBwtTwsSd\endcsname{0.0}
\expandafter\gdef\csname fiveresCertAdv\endcsname{--}
\expandafter\gdef\csname fiveresCertDolly\endcsname{--}
\expandafter\gdef\csname fiveresCertMMLU\endcsname{--}
\expandafter\gdef\csname fiveresCertPAdv\endcsname{1.2}
\expandafter\gdef\csname fiveresCertPAdvSd\endcsname{0.0}
\expandafter\gdef\csname fiveresCertPDolly\endcsname{8.0}
\expandafter\gdef\csname fiveresCertPDollySd\endcsname{0.0}
\expandafter\gdef\csname fiveresCertPMMLU\endcsname{55.4}
\expandafter\gdef\csname fiveresCertPMMLUSd\endcsname{0.0}
\expandafter\gdef\csname fiveresCertPXS\endcsname{0.7}
\expandafter\gdef\csname fiveresCertPXSSd\endcsname{0.0}
\expandafter\gdef\csname fiveresCertXS\endcsname{--}
\expandafter\gdef\csname fiveresChgAgemAdv\endcsname{--}
\expandafter\gdef\csname fiveresChgAgemDolly\endcsname{--}
\expandafter\gdef\csname fiveresChgAgemMmlu\endcsname{--}
\expandafter\gdef\csname fiveresChgAgemXs\endcsname{--}
\expandafter\gdef\csname fiveresChgEwcAdv\endcsname{--}
\expandafter\gdef\csname fiveresChgEwcDolly\endcsname{--}
\expandafter\gdef\csname fiveresChgEwcMmlu\endcsname{--}
\expandafter\gdef\csname fiveresChgEwcXs\endcsname{--}
\expandafter\gdef\csname fiveresChgKlAdv\endcsname{--}
\expandafter\gdef\csname fiveresChgKlDolly\endcsname{--}
\expandafter\gdef\csname fiveresChgKlMmlu\endcsname{--}
\expandafter\gdef\csname fiveresChgKlXs\endcsname{--}
\expandafter\gdef\csname fiveresChgLtwoAdv\endcsname{--}
\expandafter\gdef\csname fiveresChgLtwoDolly\endcsname{--}
\expandafter\gdef\csname fiveresChgLtwoMmlu\endcsname{--}
\expandafter\gdef\csname fiveresChgLtwoXs\endcsname{--}
\expandafter\gdef\csname fiveresChgMasAdv\endcsname{--}
\expandafter\gdef\csname fiveresChgMasDolly\endcsname{--}
\expandafter\gdef\csname fiveresChgMasMmlu\endcsname{--}
\expandafter\gdef\csname fiveresChgMasXs\endcsname{--}
\expandafter\gdef\csname fiveresChgNullAdv\endcsname{--}
\expandafter\gdef\csname fiveresChgNullDolly\endcsname{--}
\expandafter\gdef\csname fiveresChgNullMmlu\endcsname{--}
\expandafter\gdef\csname fiveresChgNullXs\endcsname{--}
\expandafter\gdef\csname fiveresChgOloraAdv\endcsname{--}
\expandafter\gdef\csname fiveresChgOloraDolly\endcsname{--}
\expandafter\gdef\csname fiveresChgOloraMmlu\endcsname{--}
\expandafter\gdef\csname fiveresChgOloraXs\endcsname{--}
\expandafter\gdef\csname fiveresChgRefklAdv\endcsname{--}
\expandafter\gdef\csname fiveresChgRefklDolly\endcsname{--}
\expandafter\gdef\csname fiveresChgRefklMmlu\endcsname{--}
\expandafter\gdef\csname fiveresChgRefklXs\endcsname{--}
\expandafter\gdef\csname fiveresChgReplayAdv\endcsname{--}
\expandafter\gdef\csname fiveresChgReplayDolly\endcsname{--}
\expandafter\gdef\csname fiveresChgReplayMmlu\endcsname{--}
\expandafter\gdef\csname fiveresChgReplayXs\endcsname{--}
\expandafter\gdef\csname fiveresChgRouterAdv\endcsname{--}
\expandafter\gdef\csname fiveresChgRouterDolly\endcsname{--}
\expandafter\gdef\csname fiveresChgRouterMmlu\endcsname{--}
\expandafter\gdef\csname fiveresChgRouterXs\endcsname{--}
\expandafter\gdef\csname fiveresChgScaleAdv\endcsname{--}
\expandafter\gdef\csname fiveresChgScaleDolly\endcsname{--}
\expandafter\gdef\csname fiveresChgScaleMmlu\endcsname{--}
\expandafter\gdef\csname fiveresChgScaleXs\endcsname{--}
\expandafter\gdef\csname fiveresChgSeqAdv\endcsname{9.2}
\expandafter\gdef\csname fiveresChgSeqAdvSd\endcsname{1.4}
\expandafter\gdef\csname fiveresChgSeqDolly\endcsname{91.8}
\expandafter\gdef\csname fiveresChgSeqDollySd\endcsname{1.6}
\expandafter\gdef\csname fiveresChgSeqMmlu\endcsname{34.6}
\expandafter\gdef\csname fiveresChgSeqMmluSd\endcsname{8.2}
\expandafter\gdef\csname fiveresChgSeqXs\endcsname{68.6}
\expandafter\gdef\csname fiveresChgSeqXsSd\endcsname{2.2}
\expandafter\gdef\csname fiveresChgTwAdv\endcsname{--}
\expandafter\gdef\csname fiveresChgTwDolly\endcsname{--}
\expandafter\gdef\csname fiveresChgTwMmlu\endcsname{--}
\expandafter\gdef\csname fiveresChgTwXs\endcsname{--}
\expandafter\gdef\csname fiveresChgTwpAdv\endcsname{0.0}
\expandafter\gdef\csname fiveresChgTwpAdvSd\endcsname{0.0}
\expandafter\gdef\csname fiveresChgTwpDolly\endcsname{3.3}
\expandafter\gdef\csname fiveresChgTwpDollySd\endcsname{0.1}
\expandafter\gdef\csname fiveresChgTwpMmlu\endcsname{5.2}
\expandafter\gdef\csname fiveresChgTwpMmluSd\endcsname{0.3}
\expandafter\gdef\csname fiveresChgTwpXs\endcsname{0.0}
\expandafter\gdef\csname fiveresChgTwpXsSd\endcsname{0.0}
\expandafter\gdef\csname fiveresChgTwsAdv\endcsname{0.0}
\expandafter\gdef\csname fiveresChgTwsAdvSd\endcsname{0.0}
\expandafter\gdef\csname fiveresChgTwsDolly\endcsname{3.4}
\expandafter\gdef\csname fiveresChgTwsDollySd\endcsname{0.1}
\expandafter\gdef\csname fiveresChgTwsMmlu\endcsname{7.8}
\expandafter\gdef\csname fiveresChgTwsMmluSd\endcsname{1.9}
\expandafter\gdef\csname fiveresChgTwsXs\endcsname{0.0}
\expandafter\gdef\csname fiveresChgTwsXsSd\endcsname{0.0}
\expandafter\gdef\csname fiveresExactN\endcsname{--}
\expandafter\gdef\csname fiveresExactPN\endcsname{42582}
\expandafter\gdef\csname fiveresExactPViol\endcsname{0}
\expandafter\gdef\csname fiveresExactViol\endcsname{--}
\expandafter\gdef\csname fiveresFireRouterAdv\endcsname{--}
\expandafter\gdef\csname fiveresFireRouterDolly\endcsname{--}
\expandafter\gdef\csname fiveresFireRouterMmlu\endcsname{--}
\expandafter\gdef\csname fiveresFireRouterXs\endcsname{--}
\expandafter\gdef\csname fiveresFireTwAdv\endcsname{--}
\expandafter\gdef\csname fiveresFireTwDolly\endcsname{--}
\expandafter\gdef\csname fiveresFireTwMmlu\endcsname{--}
\expandafter\gdef\csname fiveresFireTwXs\endcsname{--}
\expandafter\gdef\csname fiveresFireTwpAdv\endcsname{0.0}
\expandafter\gdef\csname fiveresFireTwpAdvSd\endcsname{0.0}
\expandafter\gdef\csname fiveresFireTwpDolly\endcsname{5.8}
\expandafter\gdef\csname fiveresFireTwpDollySd\endcsname{0.0}
\expandafter\gdef\csname fiveresFireTwpMmlu\endcsname{51.6}
\expandafter\gdef\csname fiveresFireTwpMmluSd\endcsname{0.0}
\expandafter\gdef\csname fiveresFireTwpXs\endcsname{0.0}
\expandafter\gdef\csname fiveresFireTwpXsSd\endcsname{0.0}
\expandafter\gdef\csname fiveresFireTwsAdv\endcsname{0.0}
\expandafter\gdef\csname fiveresFireTwsAdvSd\endcsname{0.0}
\expandafter\gdef\csname fiveresFireTwsDolly\endcsname{5.8}
\expandafter\gdef\csname fiveresFireTwsDollySd\endcsname{0.0}
\expandafter\gdef\csname fiveresFireTwsMmlu\endcsname{51.6}
\expandafter\gdef\csname fiveresFireTwsMmluSd\endcsname{0.0}
\expandafter\gdef\csname fiveresFireTwsXs\endcsname{0.0}
\expandafter\gdef\csname fiveresFireTwsXsSd\endcsname{0.0}
\expandafter\gdef\csname fiveresRetain\endcsname{--}
\expandafter\gdef\csname fiveresSeeds\endcsname{3}
\expandafter\gdef\csname fiveresSeqAdvDelta\endcsname{$-$1.6}
\expandafter\gdef\csname fiveresSeqAdvDeltaSd\endcsname{0.8}
\expandafter\gdef\csname fiveresSeqChangeDolly\endcsname{91.7}
\expandafter\gdef\csname fiveresSeqChangeDollySd\endcsname{1.7}
\expandafter\gdef\csname fiveresSeqLeak\endcsname{0.9}
\expandafter\gdef\csname fiveresSeqLeakSd\endcsname{1.0}
\expandafter\gdef\csname fiveresTwLeak\endcsname{--}
\newcommand{\sev}[1]{\ifcsname seven#1\endcsname\csname seven#1\endcsname\else--\fi}
\newcommand{\fiv}[1]{\ifcsname five#1\endcsname\csname five#1\endcsname\else--\fi}
\def\pv#1{\providecommand{#1}{--}}
\pv\resCertPDolly \pv\resCertPAdv \pv\resCertPXS \pv\resCertPMMLU \pv\resExactPN \pv\resExactPViol
\pv\resSeeds \pv\resExactN \pv\resExactViol \pv\resActMin \pv\resRetain
\pv\resCertDolly \pv\resCertAdv \pv\resCertXS \pv\resCertMMLU \pv\resSeqLeak \pv\resTwLeak
\pv\resAvgSeq \pv\resAvgKl \pv\resAvgReplay \pv\resAvgRefkl \pv\resAvgOlora \pv\resAvgNull \pv\resAvgScale
\pv\resAvgRouter \pv\resAvgTwp \pv\resAvgTw \pv\resAvgSeqSd \pv\resAvgTwSd \pv\resAvgTwpSd
\pv\resBwtSeq \pv\resBwtTw \pv\resBwtTwp
\pv\resChgSeqDolly \pv\resChgSeqXs \pv\resChgSeqMmlu \pv\resChgKlDolly \pv\resChgKlMmlu \pv\resChgReplayDolly
\pv\resChgReplayMmlu \pv\resChgRefklDolly \pv\resChgRefklMmlu \pv\resChgOloraDolly \pv\resChgNullDolly
\pv\resChgRouterDolly \pv\resChgTwDolly \pv\resChgTwpDolly
\pv\resFireTwDolly \pv\resFireTwMmlu \pv\resFireTwpDolly \pv\resFireTwpMmlu \pv\resFireRouterDolly \pv\resFireRouterMmlu
\pv\ovlSeeds \pv\ovlSeqGain \pv\ovlSeqChg \pv\ovlSeqAdv \pv\ovlSeqXs \pv\ovlSeqLeak
\pv\ovlKlAParam \pv\ovlKlAGain \pv\ovlKlAChg \pv\ovlKlDParam \pv\ovlKlDGain \pv\ovlKlDChg
\pv\ovlRefklBGain \pv\ovlRefklBChg \pv\ovlRefklCGain \pv\ovlRefklCChg
\pv\ovlNullBGain \pv\ovlNullBChg \pv\ovlScaleBGain \pv\ovlScaleBChg
\pv\ovlRouterGain \pv\ovlRouterChg \pv\ovlRouterFireDolly
\pv\ovlTwAParam \pv\ovlTwAGain \pv\ovlTwAChg \pv\ovlTwAFireNq \pv\ovlTwACertMax \pv\ovlTwAAuroc
\pv\ovlTwBParam \pv\ovlTwBGain \pv\ovlTwBChg \pv\ovlTwBFireNq \pv\ovlTwBCertMax \pv\ovlTwBCondGain \pv\ovlTwBInPrompt
\pv\ovlTwCParam \pv\ovlTwCGain \pv\ovlTwCChg \pv\ovlTwCFireNq \pv\ovlTwCCertMax \pv\ovlTwCCondGain
\pv\ovlTwpAGain \pv\ovlTwpAChg \pv\ovlTwpBGain \pv\ovlTwpBChg \pv\ovlTwpCGain \pv\ovlTwpCChg
\pv\ovlTwpAParam \pv\ovlTwpBParam \pv\ovlTwpCParam \pv\ovlTwpAFireNq \pv\ovlTwpBFireNq \pv\ovlTwpCFireNq
\pv\ovlTwpACertMax \pv\ovlTwpBCertMax \pv\ovlTwpCCertMax \pv\ovlTwpAXs \pv\ovlTwpBXs \pv\ovlTwpCXs \pv\ovlTwpAAuroc
\pv\ovlTwpBCondGain \pv\ovlKlBGain \pv\ovlKlBChg \pv\ovlKlCGain \pv\ovlKlCChg \pv\ovlRefklAGain \pv\ovlRefklAChg
\pv\ovlRefklCAdv \pv\ovlRefklCXs \pv\ovlRefklCMmlu \pv\ovlNullAGain \pv\ovlNullAChg \pv\ovlNullCChg \pv\ovlNullBAdv
\pv\ovlRouterFireNq \pv\ovlRouterXs \pv\ovlKlDAdv \pv\ovlSeqMmlu \pv\resChgOloraMmlu \pv\resAvgSeqSd
\pv\resChgScaleDolly \pv\ovlOracleGain \pv\resXSampViol \pv\resXSampNonfired \pv\resXSampChgSeq \pv\resXWrapFireTwp \pv\resXWrapFireTw
\pv\resXWrapRefTwp \pv\resXWrapRefTw \pv\resXWrapRefSeq \pv\resXSysFireTwp \pv\resXTurnFireTwp
\pv\ovlConcisectxTwpAAuroc \pv\ovlConcisectxSeqGain \pv\ovlConcisectxSeqChg \pv\ovlConcisectxSeqXs
\pv\ovlConcisectxSeqAdv \pv\ovlConcisectxSeqMmlu \pv\ovlConcisectxTwpAFireNq \pv\ovlConcisectxTwpAGain
\pv\ovlRefklsAGain \pv\ovlRefklsBGain \pv\ovlRefklsAChg \pv\ovlNullrfAGain \pv\ovlNullrfAChg
\pv\anNqSweepMae \pv\anNqSweepMax \pv\anSevenSweepMae \pv\anOlmoSweepMae \pv\anIfcSweepMax \pv\anConcisectxSweepMax
\pv\anNqRatioCalNDolly \pv\anNqRatioHeldNDolly \pv\anNqRatioPfisherDolly \pv\anNqRatioPexchDolly
\pv\rvNqRdPowerMean \pv\rvNqRdPowerSd \pv\rvNqRdGainMean \pv\rvNqRdGainSd \pv\rvNqRdDollyMean \pv\rvNqRdDollyMax
\pv\rvNqRdDollyAbove \pv\rvNqRdN \pv\rvNqDiffTwpVsSeq \pv\rvNqDiffTwpVsSeqCi \pv\rvNqDiffSeqswVsSeq
\pv\rvNqDiffSeqswVsSeqCi \pv\rvNqDiffRouterVsTwp \pv\rvNqDiffRouterVsTwpCi \pv\rvNqRcFull \pv\rvNqRcSmallQ
\pv\rvNqRcSmallSkew \pv\ovgIfcRatioAJb \pv\ovgIfcsevenRatioAJb \pv\ovgConcisectxRatioAPower
\pv\ovwConcisectxSeqRatioAGain \pv\ovwConcisectxSeqOwnGain \pv\ovwConcisectxSeqRatioBGain
\pv\ovwConcisectxSeqRatioCGain \pv\ovwConcisectxSeqRatioBChg \pv\ovwConcisectxSeqRatioCChg
\newcommand{\vtdir}{v3_results}

\newcommand{\vtable}[4]{\IfFileExists{\vtdir/v3_tables/#1}{\input{\vtdir/v3_tables/#1}}{\begin{#4}[t]\centering
  \caption{#2}\label{#3}\placeholder{Written by \texttt{bash slurm/plan.sh p3-tables} (PLAN\_PAPER3.md).}\end{#4}}}
\newcommand{\vtfile}[1]{\vtdir/v3_tables/#1}
\renewcommand{\vtable}[4]{\IfFileExists{\vtfile{#1}}{\input{\vtfile{#1}}}{}}
\newcommand{\vtkeep}[1]{#1}\newcommand{\vtdrop}[1]{}
\newcommand{\vtrow}[2]{\ifcsname #1\endcsname\expandafter\vtkeep\else\expandafter\vtdrop\fi{#2}}
\newif\ifvtmethods
\ifcsname vthreeConciseRefklProtectGain\endcsname\vtmethodstrue\fi
\ifcsname vthreeConciseSafemixGain\endcsname\vtmethodstrue\fi
\ifcsname vthreeConciseTripwirePromptProtectGain\endcsname\vtmethodstrue\fi
\ifcsname vthreeConcisectxRefklProtectGain\endcsname\vtmethodstrue\fi
\ifcsname vthreeConcisectxSafemixGain\endcsname\vtmethodstrue\fi
\ifcsname vthreeConcisectxTripwirePromptProtectGain\endcsname\vtmethodstrue\fi
\newif\ifvtledger
\newif\ifvtlattice
\newif\ifvtbudget
\newif\ifvtsize
\newif\ifvtstress
\newif\ifvtsquad
\newif\ifvtmethfiles

\title{Certified Policy Updates for Evolving Agents:\\ Bounding Behaviour Change Before Training in Continual Learning} 

\author{Md Anwar Hossen, Ali Jannesari}
\affiliations{Department of Computer Science, Iowa State University, Ames, IA, USA\\
\{manwar, jannesar\}@iastate.edu}

\begin{document}

\maketitle

\begin{abstract}
An AI agent that keeps learning after deployment faces the continual-learning problem. Its policy model is
fine-tuned on one task after another, and each update can silently change behaviour that should stay consistent,
such as earlier skills, general chat or safety refusals. Today such regressions are found only after training, and
only on the benchmarks someone chose to run. In this work, we bound them before training: the average change of every behaviour score at once, on the prompt
distributions that must stay protected.
We propose \method{}, which places each task's update, such as a new LoRA, behind a gate. The gate is a classifier on
the model's frozen lower layers that decides, for each prompt, whether the update is used. Wherever no gate fires,
the model computes exactly what the old model computes, and switching a gate off removes its update completely. As a
result, over a whole stream of updates, the average of every bounded behaviour score on any prompt distribution
changes by at most the gates' firing rate on that distribution. We calibrate each gate on unlabeled prompts whose
behaviour must not change. This fixes the firing rate before training, with a PAC guarantee (at most $\varepsilon$ with probability $1-\delta$ on each protected distribution) that holds for whatever weights training produces. The price of this guarantee is set by the gate's power, since no update with the same
budget can change the new task more often than the best test between new-task and protected prompts allows. On
reinforcement-learning streams of up to five tasks, gated training keeps the earlier tasks and changes only 3.3\% of
held-out chat responses, whereas ungated training changes 91.8\%. The gain kept behind the gate follows the gate's
firing rate on new-task prompts, which we measure before training. The certificate also covers whole agent episodes:
when tool-using agents are trained inside their tool loop, on a calculator task and on the API calls of a public
dialogue benchmark, gated training leaves every episode of the earlier skill identical, whereas ungated training
changes most of them and, on the benchmark, stops refusing most harmful requests.
\end{abstract}

\section{Introduction}
\label{sec:intro}

AI agents are increasingly expected to keep learning after deployment: they gain new skills, and the language model
that serves as their policy is fine-tuned again, for example with reinforcement learning (RL) on one task and later on
another \citep{gao2025survey}. An agent operating system manages an agent's context, memory, storage, tools and access
rights \citep{mei2024aios}. The policy model is one more resource that evolves, and each of its updates should be
handled like a software release: checked for regressions, monitored once deployed and rolled back if needed. Each
update must keep the earlier skills and the rest of the model's behaviour intact. However, this is checked only after
training, on whichever benchmarks were chosen, and the evidence is mixed. RL fine-tuning has been argued to forget less than supervised fine-tuning
\citep{shenfeld2026razor,chen2025retaining}, yet continual RL streams still forget \citep{luo2026cpo}, and benign
fine-tuning also erodes behaviour that nobody targeted, such as refusals \citep{qi2024finetuning}.

A check after training cannot cover the behaviours that nobody measured. This raises the question: \emph{can we bound
how much an update changes the deployed model before the update is trained, for the average of every behaviour
score at once, on the prompt distributions that must not change?}
Regularisers, replay and orthogonal low-rank adaptation (LoRA) change which model training finds, not which models it
could find, so they give no bound of this kind. To study this problem, we formulate the
following research questions:

\textbf{RQ1:} How can the behaviour change of an update be bounded before training, for every model that training may
produce, from unlabeled prompts only?

\textbf{RQ2:} What does such a bound cost on the new task, and can this cost be known before training?

\begin{figure}[t]
\centering
\includegraphics[width=\linewidth]{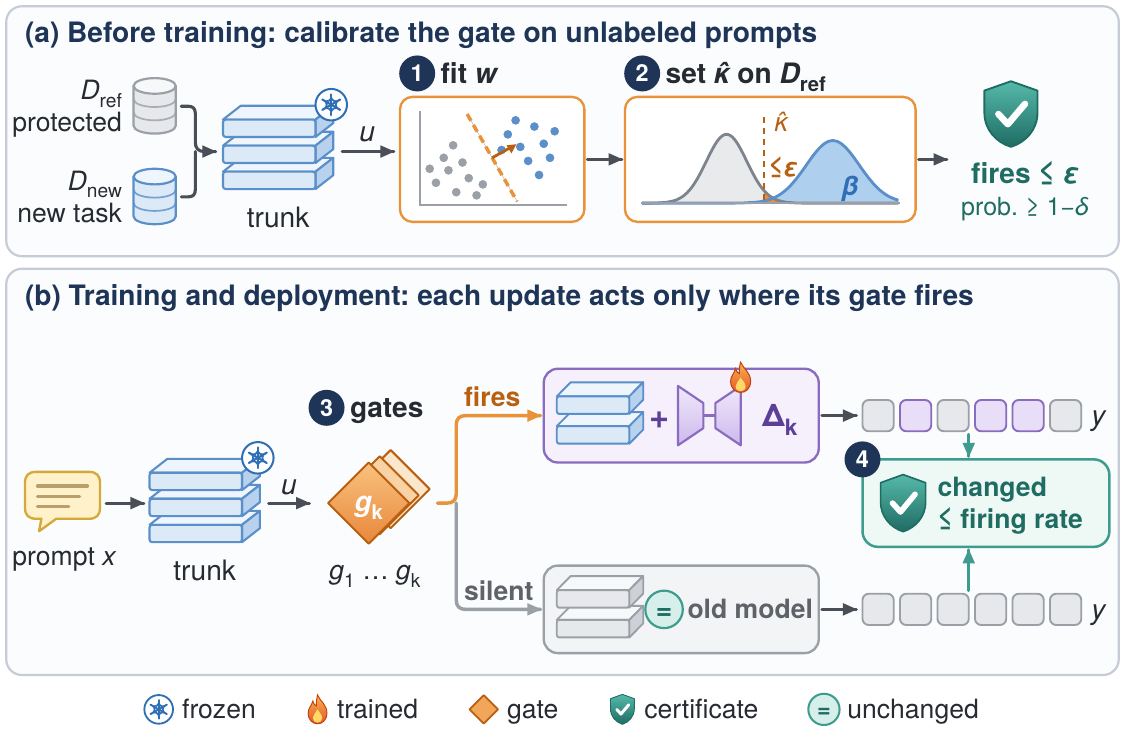}
\caption{Overview of \method{}. (a) Before training, the gate's classifier is fitted (1) and calibrated on unlabeled
prompts (2), which fixes its firing rate. (b) Each task's update acts only where its gate fires (3); elsewhere the model is
exactly the old one, so at most that share of outputs changes (4). The update can be a LoRA or a full-rank change of
the upper blocks, trained by RL or SFT, or a separately trained model behind a prompt-time gate.}
\label{fig:arch}
\end{figure}

To address these questions, we split the network into a frozen trunk, the lower blocks, and the
upper blocks. Each new task gets an update in the upper blocks, such as a LoRA, that is switched on only after a gate
fires; the gate is a classifier on trunk features that tests whether the prompt belongs to the new task
(\cref{fig:arch}). Until it fires, the updated model computes exactly what the old model computes. Therefore, the
total-variation (TV) distance between their outputs on any prompt distribution is at most the gate's firing rate
there. This rate does not depend on the update's weights, so we fix it before training: we set the gate's threshold
on unlabeled prompts of each component that must not change, as one sets the level of a Neyman--Pearson test
(\cref{thm:calib}). Forgetting on these components is then at most the gate's false-alarm rate. The bound concerns
averages over these prompt distributions, not each single prompt, and it covers only the distributions that are
sampled, so choosing them is the practitioner's main task.

The price is set by how well the new task can be told from the protected prompts: the update can act only on the
share $\beta$ of new-task prompts that the chosen gate accepts, its coverage, which is measured before training, and
no gate can exceed the power of the best test (\cref{thm:limit}). For an agent, each policy update thus becomes a release
with a regression bound fixed before training, a firing log that shows on which prompts the update can have acted, and
exact rollback by switching its gate off (\cref{sec:deploy}). The key contributions of our work are as follows:

We propose \method{}, a gated update with a certificate on the deployed model: on any prompt
distribution, the share of changed outputs is at most the gate's firing rate, for any update above the split and for
any separately trained model behind a prompt-time gate. We calibrate the gate on unlabeled prompts, which fixes this
rate before training with a PAC guarantee, extend the certificate to task streams, and show that the same gate supports
the release of an agent's updates: an audit of the fired prompts, anytime-valid monitoring of live traffic and exact
rollback (\cref{sec:method}).

We derive the price of the certificate: a Neyman--Pearson limit that binds every method,
gated or not, and the share of the gain that a model behind the gate keeps, which can be estimated before training
(\cref{sec:price}).

We evaluate \method{} on seven models from five families, two RL task streams and tasks whose
prompts look more and more like general chat, against eleven continual-learning baselines, on benchmarks written by
others, with predictions fixed before the runs, on a new task trained with every setting frozen, and inside
tool-using agents that are trained in their tool loop, on a calculator task and on the API calls of a public dialogue
benchmark (\cref{sec:experiments}).

\section{Problem Formulation and Motivation}
\label{sec:problem}

In this section, we define behaviour change and explain why a bound before training is hard to obtain.

A policy $\pi$ maps a prompt $x$ to a response $y=(y_1,y_2,\dots)$, generated token by token until an end-of-sequence
token or a length limit. Write $s=(x,y)$ for the full token sequence and $s_{\le t}$ for its first $t$ tokens. A
decoding rule turns the next-token distribution and fresh randomness $U\sim\mathrm{Unif}[0,1]$ into a token;
sampling at any temperature, top-$p$ sampling and greedy decoding are all of this form. For a prompt distribution $D$,
let $P^D_\pi$ be the law of $(x,y)$ when $x\sim D$ and $y$ is decoded from $\pi$.

In continual post-training, stage $k$ turns the model $\pi_{k-1}$ into $\pi_k$ using task $k$, which is given by
prompts and a reward $r_k\in[0,1]$. For a single update, $\pold$ and $\pnew$ denote the models before and after it,
and $\Dnew$ the new task's prompt distribution. The reference distribution $\Dref$ is a mixture of components, such
as general chat or harmful requests, on which behaviour should not change. The drift of an update on $D$ is
\begin{equation}
\TV\big(P^D_{\pnew},P^D_{\pold}\big)=\E_{x\sim D}\,\TV\big(\pnew(\cdot\mid x),\pold(\cdot\mid x)\big).
\label{eq:tv}
\end{equation}
A behaviour score is any function $f(x,y)$ with values in $[0,1]$, such as accuracy, refusal or a format check. The
mean of every such score differs between the two models by at most the drift, so a bound on the drift bounds all of
them, including the ones nobody measured. Under greedy decoding, the drift is the share of prompts whose response
changes.

\textbf{Challenge 1: The bound must hold for every trained model (RQ1).} Before training, the weights are unknown, so a
bound must hold for every model that training may produce. A penalty such as the KL divergence to the old model limits
only the weights that optimise it, and only their sampled outputs, while an unconstrained update can change almost
every response (\cref{prop:tight}). Therefore, regularisers, replay and orthogonal LoRA reduce the change measured after
training, but they give no bound before it.

\textbf{Challenge 2: No behaviour labels (RQ1).} We want to bound every behaviour score at once, so the bound cannot
rely on labels of any one behaviour. It may only use unlabeled prompts of what must not change.

\textbf{Challenge 3: The cost of the bound (RQ2).} A bound that forbids every change also forbids learning. We
therefore need to know, before training, how much of the new task an update may still change.

\section{Methodology}
\label{sec:method}

In this section, we introduce \method{} and its certificate. We first define a gated update and prove that it bounds
the drift (\cref{sec:certificate}), then calibrate the gate without labels (\cref{sec:calib}), extend it to task
streams (\cref{sec:stream}), derive its price (\cref{sec:price}) and show how it supports the release of updates
(\cref{sec:deploy}). \Cref{app:proofs} gives all proofs.

\subsection{Gated Updates and the Coupling Certificate}
\label{sec:certificate}

Let the network have $L$ blocks with causal attention and position-wise feed-forward layers. We fix a split depth
$\lg$. The trunk, blocks $1,\dots,\lg-1$, is frozen for the whole stream; the trunk feature $u_t(s)\in\R^d$ is
the residual stream entering block $\lg$ at position $t$. Because attention is causal, $u_t(s)$ depends only on
$s_{\le t}$.

\begin{definition}[Gated update]
\label{def:gated}
A gate is a map $g:\R^d\to\{0,1\}$ that reads the trunk feature at a position and returns 1 (it fires) or 0. It may
be randomised afresh per sequence, independently of input and decoding, and it may use only some positions, as long
as whether position $t$ is used depends only on $s_{\le t}$. The first firing time $T_g(s)$ is the first used
position $t$ with $g(u_t(s))=1$, or $\infty$ if there is none. The latch $m_t(s)$ is 1 from that position on and 0
before it. Let $\pold$ be an old policy and $\Delta$ an update that modifies the computation of blocks
$\lg,\dots,L$ position-wise, such as a LoRA on their projections or a full-rank change of their weights. The gated
policy $\pi_{\Delta,g}$ applies $\Delta$ at position $t$ if and only if the latch is on, $m_t(s)=1$. An anytime gate
may fire at any position after the chat-template prefix; a prompt-time gate only at the last prompt token.
\end{definition}

We say that the gate fires on $s$ if it fires at a position that can still influence the generated text, that is, if
$T_g(s)\le|s|-1$, and we write $\Fire_g$ for this event.

\begin{theorem}[Coupling certificate]
\label{thm:coupling}
For every prompt distribution $D$, every decoding rule, and every update $\Delta$,
\begin{equation*}
\TV\big(P^D_{\pi_{\Delta,g}},P^D_{\pold}\big)\;\le\;\varphi_g(D):=P^D_{\pold}\big(\Fire_g\big).
\end{equation*}
Moreover, with shared decoding randomness, the response of $\pi_{\Delta,g}$ equals that of $\pold$ on every run on
which the gate does not fire.
\end{theorem}

\emph{Proof idea:} We run both models with the same uniform variables. By induction over positions and blocks, the runs
share every hidden state until the gate first fires, and the firing decision depends only on that shared prefix.
Outside $\Fire_g$ the runs are identical, and the coupling inequality \citep{lindvall1992coupling} gives the bound. The
bound is on the average over $D$: it says nothing about any single prompt, about prompts outside $D$, or about how
responses change where the gate fires (\cref{tab:status}). It does not depend on $\Delta$, and no bound that holds for
every update can improve it by more than a factor $1-\eta$:

\begin{proposition}[Tightness]
\label{prop:tight}
Suppose an update can place all probability on a token $a^\dagger$ at the positions where it is active, and $\pold$
gives $a^\dagger$ probability at most $\eta$ at those positions. Then
$\TV(P^D_{\pi_{\Delta,g}},P^D_{\pold})\ge(1-\eta)\,\varphi_g(D)$.
\end{proposition}

Without a gate ($g\equiv1$), some update therefore drifts by nearly one, which is Challenge 1.

\textbf{A certified switch for any trained model:} A prompt-time gate decides once: its decision $G(x)=g(u_{|x|}(x))$
depends on $x$ alone, as the trunk is frozen and attention is causal.

\begin{corollary}[Certified switch]
\label{cor:switch}
Let $G$ be the decision of a prompt-time gate and $\pi'$ any policy. The switch $\pi_{\pi',G}$ answers a prompt $x$
with $\pi'$ if the gate fires, $G(x)=1$, and with $\pold$ otherwise. \Cref{thm:coupling} holds for the switch, with
the share of prompts on which the gate fires, $D(G=1)$, as its firing rate, and the switch's response equals
$\pold$'s on every prompt where the gate does not fire.
\end{corollary}

The certificate thus takes two forms. Trained under its gate, an update acts only where the gate fires, as in parameter
isolation \citep{aljundi2017expertgate,liang2025gainlora} but with a firing rate fixed before training. Behind a switch, any
trained model is certified, but it keeps its own forgetting.

\subsection{Calibration Without Behaviour Labels}
\label{sec:calib}

The certificate is a firing probability of the old model, so we can estimate it, and set the threshold, before
training, from unlabeled prompts and the old model's responses.

\begin{proposition}[Estimation on any distribution]
\label{prop:cp}
Roll out $\pold$ on $n$ i.i.d.\ prompts from $D$, independent of $g$, and let $X$ be the number of firings. With
probability at least $1-\delta$, $\varphi_g(D)\le\mathrm{CP}_{\delta}(X,n)$, the one-sided Clopper--Pearson bound
\citep{clopper1934}.
\end{proposition}

A scored gate $g_\kappa$ fires where a score $w(u)$ of the trunk feature exceeds a threshold $\kappa$. It can fire on
a sequence $s$ only if the largest score along it, $S(s)=\max_t w(u_t(s))$, exceeds $\kappa$. Its firing rate on $D$
is therefore at most the probability that $S>\kappa$ on the old model's trajectories. We set $\kappa$ before training
with the Neyman--Pearson umbrella algorithm \citep{tong2018np}, a training-conditional conformal bound
\citep{vovk2012conditional}, applied to each reference component with a union bound.

\begin{theorem}[PAC calibration]
\label{thm:calib}
Let $w$ and $\pold$ be fixed independently of the calibration set. For each reference component $c=1,\dots,C$, draw
$n_c$ i.i.d.\ calibration trajectories of $\pold$ on $D_c$ and let $S^c_{(1)}\le\dots\le S^c_{(n_c)}$ be their sorted
scores. Let $k_c=\min\{k:\Pr[\mathrm{Bin}(n_c,1-\varepsilon)\ge k]\le\delta/C\}$, $\hat\kappa_c=S^c_{(k_c)}$
($+\infty$ if $k_c>n_c$), and $\hat\kappa=\max_c\hat\kappa_c$. Then with probability at least $1-\delta$ over the
calibration draws, $\varphi_{g_{\hat\kappa}}(D_c)\le\varepsilon$ for every $c$.
\end{theorem}

In words, each component's threshold is one of the largest scores of its calibration trajectories, at a rank set by
a binomial tail, and the gate uses the largest of these thresholds. On each reference component, the certificate is
then the level $\varepsilon$, for prompts drawn like the calibration prompts. The firing rate this rule actually gives is lower than $\varepsilon$: on average at most 1.7\% for our 700 Dolly calibration prompts
(\cref{prop:size}). The firing rate is linear in the prompt distribution, so the level also holds on every mixture of
components. On a distribution far from every component, such as MMLU, the only certificate is the bound of
\cref{prop:cp} from a sample of it.

\textbf{Two-sided gate:} A classifier trained to tell the new task from the reference may fire on prompts unlike both.
We therefore add a region test: the gate fires only on prompts whose Mahalanobis distance to the new task's fitting
prompts is below its $1-\alpha$ quantile on unlabeled RL prompts, with $\alpha=0.02$. The region is fixed before
calibration, so \cref{thm:calib} calibrates the combined gate with the same guarantee, and its firing on any
distribution is at most the mass that distribution puts in the region. With 1{,}000 RL prompts, the same binomial
argument shows that the region misses at most 2.9\% of new-task prompts with probability 0.95 (\cref{app:impl}).

\subsection{Task Streams}
\label{sec:stream}

In a stream, stage $k$ adds a gate $g_k$ on the same frozen trunk and an update $\Delta_k$ that acts only behind this
gate's latch. The gates and updates of all earlier stages stay as they are.

\begin{theorem}[Stream certificate]
\label{thm:stream}
For every $j<k$ and every distribution $D$, the drift from $\pi_j$ to $\pi_k$ is at most the probability that some
gate added after stage $j$ fires on $\pi_j$'s own trajectories, and hence at most the sum of these gates' firing
rates:
\begin{equation*}
\TV\big(P^D_{\pi_k},P^D_{\pi_j}\big)\le P^D_{\pi_j}\Big(\bigcup_{i=j+1}^{k}\Fire_{g_i}\Big)
\le\sum_{i=j+1}^{k}P^D_{\pi_j}\big(\Fire_{g_i}\big).
\end{equation*}
\end{theorem}

Earlier tasks are therefore protected without replaying their labels: the drift on task $j$ is bounded by how often
the later gates fire on $\pi_j$'s own trajectories of that task. Calibrating each stage at $\varepsilon$ certifies the union of $K$ stages
at $K\varepsilon$ with confidence $1-K\delta$ (\cref{cor:compose}). With prompt-time gates, the union can also be
certified directly as one gate on fresh prompts, at a level that does not grow with $K$ (\cref{prop:promptstream}).

\subsection{The Price of Certification}
\label{sec:price}

An update that changes at most a share $\varepsilon$ of behaviour on the reference can change the new task only as far
as a test of level $\varepsilon$ can tell the two apart, whether it is gated or not.

\begin{theorem}[Certified plasticity limit]
\label{thm:limit}
Let $\pnew$ be any policy, gated or not, and let $\rho(x)$ be the TV distance between its response distribution and
the old model's on prompt $x$. If the mean of $\rho$ on $\Dref$ is at most $\varepsilon$, then its mean on $\Dnew$ is
at most $\beta^\ast(\varepsilon)$, the power of the size-$\varepsilon$ Neyman--Pearson test of $\Dref$ against
$\Dnew$. Moreover, for $\varepsilon\in(0,1)$,
\begin{equation}
\beta^\ast(\varepsilon)\le\frac{\KL(\Dnew\|\Dref)+\log 2}{\log(1/\varepsilon)}.
\label{eq:limit}
\end{equation}
The expected reward gain on $\Dnew$ is at most the mean of $\rho$ on $\Dnew$ times the reward range. If the update can move the
response distribution of every prompt where it acts as \cref{prop:tight} assumes, and the likelihood ratio of $\Dnew$
to $\Dref$ is a function of the trunk feature $u_{|x|}(x)$, then $\beta^\ast(\varepsilon)$ is attained up to the
factor $1-\eta$ by a gated update whose prompt-time gate is the likelihood-ratio test, randomised on ties.
\end{theorem}

\begin{figure}[t]
\centering
\includegraphics[width=\linewidth]{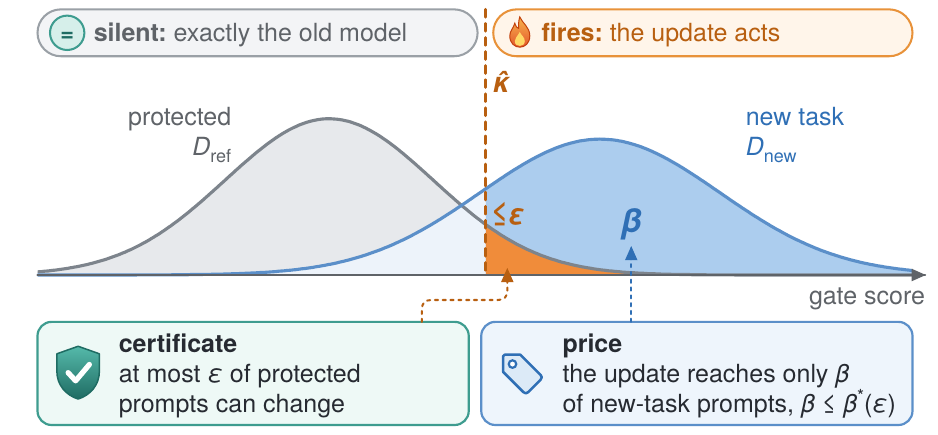}
\caption{The certificate and its price, schematically. At most a share $\varepsilon$ of protected prompts lie above the
threshold, the only ones whose outputs can change; the update acts only on the share $\beta$ of new-task prompts above
it.}
\label{fig:np}
\end{figure}

In words, an update can change the new task only as far as a test can tell its prompts from the reference, whether it
is gated, regularised, replays data or uses orthogonal LoRA (\cref{fig:np}). The first claim is the Neyman--Pearson lemma
\citep{neyman1933}. On real tasks the limit is unknown, so the coverage of any gate is a lower bound on
$\beta^\ast(\varepsilon)$, not an estimate of it. The coverage also predicts the gain of a model behind the gate:

\begin{proposition}[Gain behind a gate]
\label{prop:switchgain}
Let $G$ be the decision of a prompt-time gate with coverage $\beta=\Dnew(G=1)\in(0,1)$, and let $\pi'$ be any policy.
Write $\Delta r(x)$ for the difference between the expected rewards of $\pi'$ and $\pold$ on prompt $x$. Let $\bar G$
be its mean on $\Dnew$, the gain of $\pi'$ itself, and let $\bar G_1$ and $\bar G_0$ be its means on the new-task
prompts where the gate fires and where it does not. Then the gain of the switch $\pi_{\pi',G}$ on $\Dnew$ is
\begin{equation*}
\beta\bar G_1=\beta\bar G+\beta(1-\beta)(\bar G_1-\bar G_0).
\end{equation*}
\end{proposition}

The coverage $\beta$ is the gate's firing rate on unlabeled new-task prompts, so we estimate it before training. The
gain kept is $\beta\bar G$ plus a covariance term that the certificate does not control. This term vanishes as
$\beta\to1$ and when the prompts the gate misses would gain as much as the others, so $\beta\bar G$ is a prediction,
which we test in \cref{sec:res-cost}, not a bound.

\subsection{\method{} in Practice}
\label{sec:practice}

The gate is a linear classifier on normalised trunk features, fitted by class-balanced logistic regression to separate
prompts of the new task from those of the reference components; its only supervision is which pool a prompt came from.
Fitting and calibration use disjoint prompts, as \cref{thm:calib} requires. \methodp{} scores the last prompt token and
\methoda{} every position after the template prefix. A prompt-time gate's level holds under every decoding rule, an
anytime gate's under the rule used in its calibration.
An update is trained under its own gate, where rollouts on which
the gate does not fire carry no gradient, or trained by any method and placed behind the gate afterwards
(\cref{cor:switch}). \Cref{alg:tripwire} in the appendix gives one stage of a stream, and \cref{tab:variants} every
variant.

\textbf{Choosing the reference:} The certificate covers only the components one samples, so choosing them is part of
the method, and each choice can be priced before training. (1) Calibrate on every kind of prompt whose behaviour must
not change. (2) Check any other unlabeled pool before training with \cref{prop:cp}. (3) Use the two-sided gate. (4) If
a pool still fires more often than $\varepsilon$, such as harmful requests written in the task's format, add it as a
component. (5) If the task cannot be told apart from some component, such as general chat, calibrate only on the
components to protect. The coverage $\beta$ after each step is the price of that step (\cref{prop:switchgain}), so the
recipe also tells, before training, whether certification is worth its price. We call the gate after steps (1)--(2)
the deployed gate, and the gate the recipe ends with for a task the chosen gate (\cref{tab:variants}).

\subsection{Releasing Updates: Audit, Monitoring and Rollback}
\label{sec:deploy}

An agent's updates run on live traffic, which can drift away from the calibration prompts. Because a response can
change only where a gate fires (\cref{thm:coupling}), the gates' firing log is a trace that attributes every change to
the updates that can have caused it. Three checks follow that need neither behaviour labels nor retraining.

\begin{proposition}[Audit, monitoring and rollback]
\label{prop:deploy}
\emph{(a) Audit.} Decode the old and the new model with shared randomness, and let $f$ be a score with values in
$[a,b]$. Write $f_{\mathrm{new}}$ and $f_{\mathrm{old}}$ for the scores of the two responses to the same prompt. The
change of the mean score on $D$ is the firing rate times the mean change on the fired prompts:
\begin{equation*}
\E_D\big[f_{\mathrm{new}}-f_{\mathrm{old}}\big]
=\varphi_g(D)\,\E_D\big[f_{\mathrm{new}}-f_{\mathrm{old}}\mid\Fire_g\big].
\end{equation*}
On $n$ i.i.d.\ prompts, the sum of the differences $f_{\mathrm{new},i}-f_{\mathrm{old},i}$ over the fired prompts,
divided by $n$, estimates this change without bias, with variance at most $\varphi_g(D)(b-a)^2/n$.

\emph{(b) Monitoring.} Let $X_t\in\{0,1\}$ record whether the gate fires on the $t$-th live prompt. For a fixed
$\lambda\in[0,1/\varepsilon)$, the monitor tracks the product
\begin{equation*}
M_t=\prod_{i\le t}\big(1+\lambda(X_i-\varepsilon)\big)
\end{equation*}
and raises an alarm when $M_t\ge1/\delta$. If every prompt fires with probability at most $\varepsilon$ given the
earlier ones, $\E[X_t\mid X_{<t}]\le\varepsilon$ for all $t$, then the alarm is ever raised with probability at most
$\delta$. If instead the $X_t$ are i.i.d.\ and fire with probability $p$, where $\varepsilon<p<1$, and
$\lambda=\frac{p-\varepsilon}{\varepsilon(1-\varepsilon)}$, then the expected number of prompts until the alarm is at
most
\begin{equation*}
\frac{\log(1/\delta)+\log(1+\lambda)}{\kl(p\|\varepsilon)}+1,
\end{equation*}
where $\kl$ is the KL divergence between Bernoulli laws.

\emph{(c) Rollback.} Raising a gate's threshold keeps the update and never increases $\varphi_g(D)$ on any $D$, so
every certificate issued before remains valid. A gate that never fires gives exactly the model without its update.
\end{proposition}

Part (a) means that an audit of an update evaluates only the fired prompts, at a share $\varphi_g(D)$ of the cost of a
full evaluation; the certificate says nothing about how fired responses change, so these are the prompts to check.
Part (b) is an alarm that is valid at every time, however long the agent runs \citep{ville1939,ramdas2023gametheoretic}:
for a gate certified at 3\% whose traffic fires on 6\%, the monitor tuned to that rate raises the alarm after at most
309 prompts in expectation at $\delta=0.05$, and a mixture over several $\lambda$ needs no assumed rate
(\cref{app:proofs}). Part (c) repairs a certificate that fails on new traffic: recalibrating by \cref{thm:calib} on
fresh unlabeled prompts of that traffic and keeping the larger threshold certifies it too, while all earlier
certificates stay valid; the price is the coverage lost. Switching the gate off withdraws the update at once. For a
staged rollout, \cref{prop:cp} estimates the firing rate on a new pool of traffic from its unlabeled prompts before the
update is exposed to it.

\textbf{Whole agent episodes:} An agent calls its policy several times per episode. Each call reads the episode so
far, and a tool's reply is appended before the next call.

\begin{corollary}[Episode certificate]
\label{cor:episode}
Let an episode start from a user turn $x\sim D$, let each tool reply be drawn from a fixed kernel of the episode so
far, and let a prompt-time gate decide at every model call. Then the TV distance between the laws of whole episodes
under the gated agent and under the old agent is at most $\Phi_g(D)$, the probability that the gate fires at some call
of the old agent's episode. With shared decoding and tool randomness, the two episodes coincide whenever no call fires.
\end{corollary}

Whether an episode fires is a function of the old agent's episode, so \cref{prop:cp} estimates $\Phi_g(D)$ from
unlabeled user turns. \Cref{thm:calib} calibrates it with the episode score: the largest gate score over the
episode's model calls, each read at the last token of the call's context.

\textbf{\method{} in an agent operating system:} These pieces match an agent operating system's services. Admission:
an update is accepted only if its certificate on every protected component is within budget, at a price $\beta$ known
before training. Accounting: the certificates of successive updates add up, or their union is certified directly
(\cref{thm:stream,cor:compose,prop:promptstream}). Isolation: an update's firing rate bounds its blast radius.
Tracing: the firing log names the updates that can have changed a response.

\section{Experiments}
\label{sec:experiments}

\subsection{Experimental Setup}
\label{sec:setup-main}

\textbf{Models and tasks:} We use Qwen2.5-1.5B- and 7B-Instruct \citep{qwen2025qwen25}, OLMo-2-7B-Instruct
\citep{olmo2024olmo2}, Llama-3.2-3B- and 3.1-8B-Instruct \citep{grattafiori2024llama3}, DeepSeek-LLM-7B-Chat
\citep{bi2024deepseekllm} and Mistral-7B-Instruct \citep{jiang2023mistral}. We train two streams of tasks that are easy
to tell apart from the reference: GSM8K \citep{cobbe2021gsm8k}, Countdown \citep{pan2025tinyzero} and Knights and
Knaves \citep{xie2024kk}, and a five-task stream that adds MATH \citep{hendrycks2021math} and Sudoku. We also train
single tasks whose prompts look more and more like the reference: IF, Dolly requests with a format constraint
\citep{zhou2023ifeval}; NQ, closed-book Natural Questions \citep{kwiatkowski2019nq}; Concise, Dolly requests rewarded
for a short answer; and SQuAD~2.0 \citep{rajpurkar2018squad2}. The gate reads block 9 of 28 in Qwen2.5, about 0.3 of
the depth in every model, with a LoRA of rank 16 above it \citep{hu2022lora}, trained by GRPO
\citep{shao2024deepseekmath} for 150 steps per task, with three seeds.

\textbf{Baselines:} We compare \methodp{} and \methoda{} at $\varepsilon=0.03$ against eleven baselines with the same
LoRA and budget: Seq-GRPO, an ungated LoRA per task; +KL to the previous model \citep{shenfeld2026razor}; +Ref-KL, the
same penalty on the reference prompts the gate uses \citep{li2018lwf}; null-space LoRA \citep{liang2024inflora}, EWC
\citep{kirkpatrick2017ewc} and A-GEM \citep{chaudhry2019agem} on the same prompts; LoRA scaling
\citep{wortsman2022robust}, O-LoRA \citep{wang2023olora}, replay, safety replay \citep{pham2025preventing}, and an
uncalibrated router. We also evaluate every ungated run behind the deployed gate. The two uses of a gate answer
different questions: a trained model placed behind the gate, a switch, tests the confinement that \cref{cor:switch}
guarantees for any model, while training under the gate tests whether the restricted training still learns the task.

\textbf{Reference pools and evaluation:} Gates are calibrated with $\delta=0.05$ on unlabeled prompts of four kinds:
Dolly-15k requests, AdvBench harmful requests \citep{zou2023gcg}, task-like prompts and stored
prompts of earlier tasks. We measure the share of changed responses on held-out Dolly and AdvBench prompts and on
XSTest \citep{rottger2024xstest} and MMLU \citep{hendrycks2021mmlu}, which are never calibrated on. Refusal is scored by a string check, so it measures kept behaviour, not safety. \Cref{app:setup}
gives every setting, and \cref{tab:protocol} marks what was fixed before its evidence.

\subsection{Checks of the Certificate}
\label{sec:res-cert}

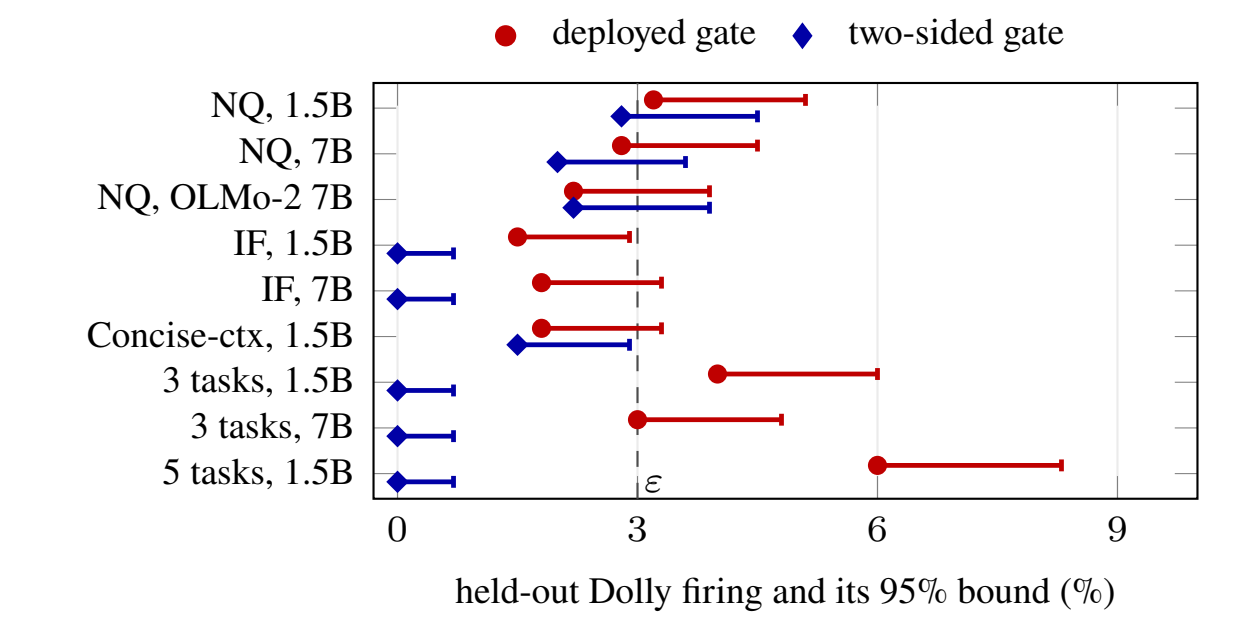
\begin{figure}[t]
\centering
\newcommand{\certrow}[5]{%
  \draw[red!75!black, line width=0.9pt] (axis cs:#2,#1+0.18) -- (axis cs:#3,#1+0.18);
  \draw[red!75!black, line width=0.9pt] (axis cs:#3,#1+0.05) -- (axis cs:#3,#1+0.31);
  \node[circle, fill=red!75!black, inner sep=0pt, minimum size=3.6pt] at (axis cs:#2,#1+0.18) {};
  \draw[blue!65!black, line width=0.9pt] (axis cs:#4,#1-0.18) -- (axis cs:#5,#1-0.18);
  \draw[blue!65!black, line width=0.9pt] (axis cs:#5,#1-0.31) -- (axis cs:#5,#1-0.05);
  \node[diamond, fill=blue!65!black, inner sep=0pt, minimum size=4.4pt] at (axis cs:#4,#1-0.18) {};}
\begin{tikzpicture}
\begin{axis}[scale only axis, width=5.45cm, height=2.75cm, xmin=-0.3, xmax=10, ymin=0.45, ymax=9.55,
  ytick={1,...,9}, yticklabels={{5 tasks, 1.5B},{3 tasks, 7B},{3 tasks, 1.5B},{Concise-ctx, 1.5B},{IF, 7B},{IF, 1.5B},
  {NQ, OLMo-2 7B},{NQ, 7B},{NQ, 1.5B}}, xtick={0,3,6,9},
  xlabel={held-out Dolly firing and its 95\% bound (\%)}, tick label style={font=\scriptsize},
  label style={font=\scriptsize}, xmajorgrids, grid style={black!8}, clip=false,
  legend style={font=\scriptsize, draw=none, at={(0.5,1.02)}, anchor=south, column sep=5pt}, legend columns=2]
\addlegendimage{only marks, mark=*, mark size=1.8pt, red!75!black}\addlegendentry{deployed gate}
\addlegendimage{only marks, mark=diamond*, mark size=2.2pt, blue!65!black}\addlegendentry{two-sided gate}
\draw[dashed, black!70] (axis cs:3,0.45) -- (axis cs:3,9.55);
\node[font=\scriptsize, anchor=south west, inner sep=1pt] at (axis cs:3,0.45) {$\varepsilon$};
\certrow{9}{\ovgNqRatioADolly}{\ovgNqRatioACpDolly}{\ovgNqTwoADolly}{\ovgNqTwoACpDolly}
\certrow{8}{\ovgNqsevenRatioADolly}{\ovgNqsevenRatioACpDolly}{\ovgNqsevenTwoADolly}{\ovgNqsevenTwoACpDolly}
\certrow{7}{\ovgNqolmoRatioADolly}{\ovgNqolmoRatioACpDolly}{\ovgNqolmoTwoADolly}{\ovgNqolmoTwoACpDolly}
\certrow{6}{\ovgIfcRatioADolly}{\ovgIfcRatioACpDolly}{\ovgIfcTwoADolly}{\ovgIfcTwoACpDolly}
\certrow{5}{\ovgIfcsevenRatioADolly}{\ovgIfcsevenRatioACpDolly}{\ovgIfcsevenTwoADolly}{\ovgIfcsevenTwoACpDolly}
\certrow{4}{\ovgConcisectxRatioADolly}{\ovgConcisectxRatioACpDolly}{\ovgConcisectxTwoADolly}{\ovgConcisectxTwoACpDolly}
\certrow{3}{\ovgStreamRatioADolly}{\ovgStreamRatioACpDolly}{\ovgStreamTwoADolly}{\ovgStreamTwoACpDolly}
\certrow{2}{\ovgStreamsevenRatioADolly}{\ovgStreamsevenRatioACpDolly}{\ovgStreamsevenTwoADolly}{\ovgStreamsevenTwoACpDolly}
\certrow{1}{\ovgStreamfiveRatioADolly}{\ovgStreamfiveRatioACpDolly}{\ovgStreamfiveTwoADolly}{\ovgStreamfiveTwoACpDolly}
\end{axis}
\end{tikzpicture}
\caption{Firing on held-out Dolly prompts (marker) and its 95\% Clopper--Pearson bound (bar end). Each
gate is certified at $\varepsilon=3\%$ with probability 0.95 (\cref{thm:calib}); a stream gate, the union of $K$
per-stage gates, at $K\varepsilon$. \Cref{tab:certs} adds $\beta$ and MMLU.}
\label{fig:certs}
\end{figure}

\textbf{Exactness:} \Cref{tab:fam,tab:decode} count the responses that changed on prompts where the gate stayed
silent, under greedy decoding and under sampling, and there are none in any run. This checks the implementation: batching and
bf16 numerics could break the exactness \cref{thm:coupling} relies on.

\textbf{The calibrated level:} \Cref{fig:certs} shows that every gate fires on held-out reference prompts at about the
calibrated level of 3\% and on no held-out harmful request, and it does so again on 10{,}000 fresh Dolly requests. The
level fixed on the calibration prompts thus carries over to new prompts of the same distribution, as the PAC guarantee
states, and a whole stream stays far below the sum of its per-stage levels. The 95\% bound can exceed 3\% with only 400 held-out prompts.

\textbf{Calibration, not gating, sets the level:} \Cref{tab:zoo} compares five gates built on NQ from the same prompts
and trunk features, among them an Expert-Gate-style pair of autoencoders. At their own thresholds, as a router or an
expert gate would use them, they fire on far more reference prompts than the budget allows; once calibrated, every one
stays within it and keeps most of the gain.

\textbf{Earlier tasks in a stream:} The gates of later tasks almost never fire on the stored prompts of earlier tasks,
while Seq-GRPO changes nearly all of their responses. Each earlier task is thus protected by the same certificate as
the reference, and \cref{tab:deploy} shows that gated training keeps the earlier tasks without replaying any labels.

\subsection{The Cost of Certification}
\label{sec:res-cost}

\newcommand{\NSdir}{\resdir/v2_overlap_q15/switch}
\begin{figure}[t]
\centering
\begin{tikzpicture}
\begin{groupplot}[group style={group size=2 by 1, horizontal sep=0.18cm, y descriptions at=edge left},
  scale only axis, height=3.0cm, ymin=-0.6, ymax=7.6, ytick={0,2,4,6},
  tick label style={font=\scriptsize}, label style={font=\scriptsize}, title style={font=\scriptsize\bfseries, yshift=-0.8ex},
  every axis plot/.append style={line width=0.8pt}, grid=major, grid style={black!8}]
\nextgroupplot[width=4.15cm, xmin=-1.5, xmax=31, xtick={0,10,20,30}, ylabel={NQ gain (points)}, title={gated}]
\addplot[orange!85!black, mark=*, mark size=1.3pt] table[x=chg, y=gain, row sep=\\] {%
chg gain chg_sd gain_sd label \\
0 0 0 0 {} \\
0.92 0.33 0.14 0.23 $\varepsilon{=}0.03$ \\
6.42 1.20 0.58 0.53 $\varepsilon{=}0.1$ \\
23.33 3.13 2.50 0.42 $\varepsilon{=}0.3$ \\
};
\addplot[red!75!black, mark=diamond*, mark size=1.8pt] table[x=chg, y=gain, row sep=\\] {%
chg gain chg_sd gain_sd label \\
0 0 0 0 {} \\
3.50 4.00 0.00 1.04 $\varepsilon{=}0.03$ \\
9.42 5.27 0.14 0.31 $\varepsilon{=}0.1$ \\
28.92 5.87 0.14 2.41 $\varepsilon{=}0.3$ \\
};
\addplot[black, mark=star, mark size=2pt] table[x=chg, y=gain, row sep=\\] {%
chg gain chg_sd gain_sd eps subst subst_sd \\
3.25 4.67 0.00 1.10 0.03 1.00 0.25 \\
9.50 5.13 0.00 0.99 0.1 3.92 0.14 \\
28.33 5.20 0.14 0.87 0.3 10.17 0.63 \\
};
\addplot[only marks, black, mark=x, mark size=2.4pt, restrict x to domain=-5:40] table[x=chg, y=gain, row sep=\\] {%
chg gain chg_sd gain_sd label \\
94.17 5.20 1.84 0.87 Seq \\
6.75 5.40 0.00 0.72 Router \\
93.92 5.40 0.14 0.20 A-GEM \\
};
\addplot[only marks, black, mark=asterisk, mark size=2.4pt] table[x=chg, y=gain, row sep=\\] {%
chg gain chg_sd gain_sd label \\
0.00 5.20 0.00 0.87 Oracle \\
};
\node[font=\scriptsize, text=orange!85!black, anchor=north] at (axis cs:21,1.85) {\methoda{}};
\node[font=\scriptsize, text=red!75!black, anchor=south] at (axis cs:22,6.0) {\methodp{}};
\node[font=\scriptsize, anchor=north, align=center, inner sep=1pt] at (axis cs:20,4.95) {Seq-GRPO behind\\[-1pt]the gate};
\node[font=\scriptsize, anchor=south west, inner sep=1pt] at (axis cs:5.2,5.7) {router};
\node[font=\scriptsize, anchor=south west, inner sep=1pt] at (axis cs:-1.2,5.55) {oracle};
\nextgroupplot[width=2.88cm, xmin=52, xmax=101, xtick={60,80,100}, title={ungated},
  legend to name=frontierleg, legend columns=3, legend style={font=\scriptsize, draw=none, column sep=4pt},
  legend cell align=left]
\addplot[blue!60!black, dashed, mark=square*, mark size=1.3pt] table[x=chg, y=gain, row sep=\\] {%
chg gain chg_sd gain_sd label \\
87.75 4.93 1.15 0.42 $\beta{=}0.01$ \\
73.33 3.53 2.32 0.12 $\beta{=}0.05$ \\
64.25 1.87 0.87 0.70 $\beta{=}0.2$ \\
56.00 0.73 2.41 0.42 $\beta{=}1$ \\
};
\addlegendentry{+KL}
\addplot[teal!70!black, dashed, mark=triangle*, mark size=1.6pt] table[x=chg, y=gain, row sep=\\] {%
chg gain chg_sd gain_sd label \\
82.92 5.53 1.26 1.33 $\beta{=}0.1$ \\
86.92 2.33 0.38 0.12 $\beta{=}1$ \\
86.50 1.27 0.90 0.81 $\beta{=}10$ \\
};
\addlegendentry{+Ref-KL}
\addplot[violet!70!black, dashed, mark=pentagon*, mark size=1.6pt] table[x=chg, y=gain, row sep=\\] {%
chg gain chg_sd gain_sd label \\
97.08 4.80 2.63 0.72 $e{=}0.8$ \\
98.75 5.53 0.66 0.12 $e{=}0.95$ \\
95.25 5.00 4.13 0.72 $e{=}0.99$ \\
};
\addlegendentry{null-space LoRA}
\addplot[brown!70!black, dashed, mark=star, mark size=1.6pt] table[x=chg, y=gain, row sep=\\] {%
chg gain chg_sd gain_sd label \\
96.33 5.80 4.19 0.20 $\lambda{=}0.1$ \\
97.50 6.60 3.07 1.00 $\lambda{=}1$ \\
95.75 4.93 5.02 1.53 $\lambda{=}10$ \\
};
\addlegendentry{+EWC}
\addplot[gray!80!black, dashed, mark=o, mark size=1.3pt, restrict x to domain=1:101] table[x=chg, y=gain, row sep=\\] {%
chg gain chg_sd gain_sd label \\
0 0 0 0 {} \\
74.08 3.00 1.23 0.35 $\alpha{=}0.25$ \\
86.42 4.60 2.77 0.40 $\alpha{=}0.5$ \\
92.17 4.93 1.81 1.17 $\alpha{=}0.75$ \\
};
\addlegendentry{LoRA scale}
\addplot[only marks, black, mark=+, mark size=2.4pt, restrict x to domain=50:101] table[x=chg, y=gain, row sep=\\] {%
chg gain chg_sd gain_sd label \\
94.17 5.20 1.84 0.87 Seq \\
6.75 5.40 0.00 0.72 Router \\
93.92 5.40 0.14 0.20 A-GEM \\
};
\addlegendentry{Seq-GRPO, A-GEM}
\end{groupplot}
\node[font=\scriptsize, anchor=north] at ($(group c1r1.south east)!0.5!(group c2r1.south west) + (0,-0.42)$) {Dolly responses changed (\%)};
\end{tikzpicture}\par\ref{frontierleg}
\caption{NQ, Qwen2.5-1.5B: gain against Dolly responses changed, mean of three seeds. Each curve sweeps one knob: the
level $\varepsilon$ for the gates and Seq-GRPO behind \methodp{}'s gate, the coefficient, energy share, weight or scale
for the ungated methods. The router uses its logit-0 rule; the oracle knows each prompt's task (\cref{tab:overlap_nq}).}
\label{fig:frontier}
\end{figure}
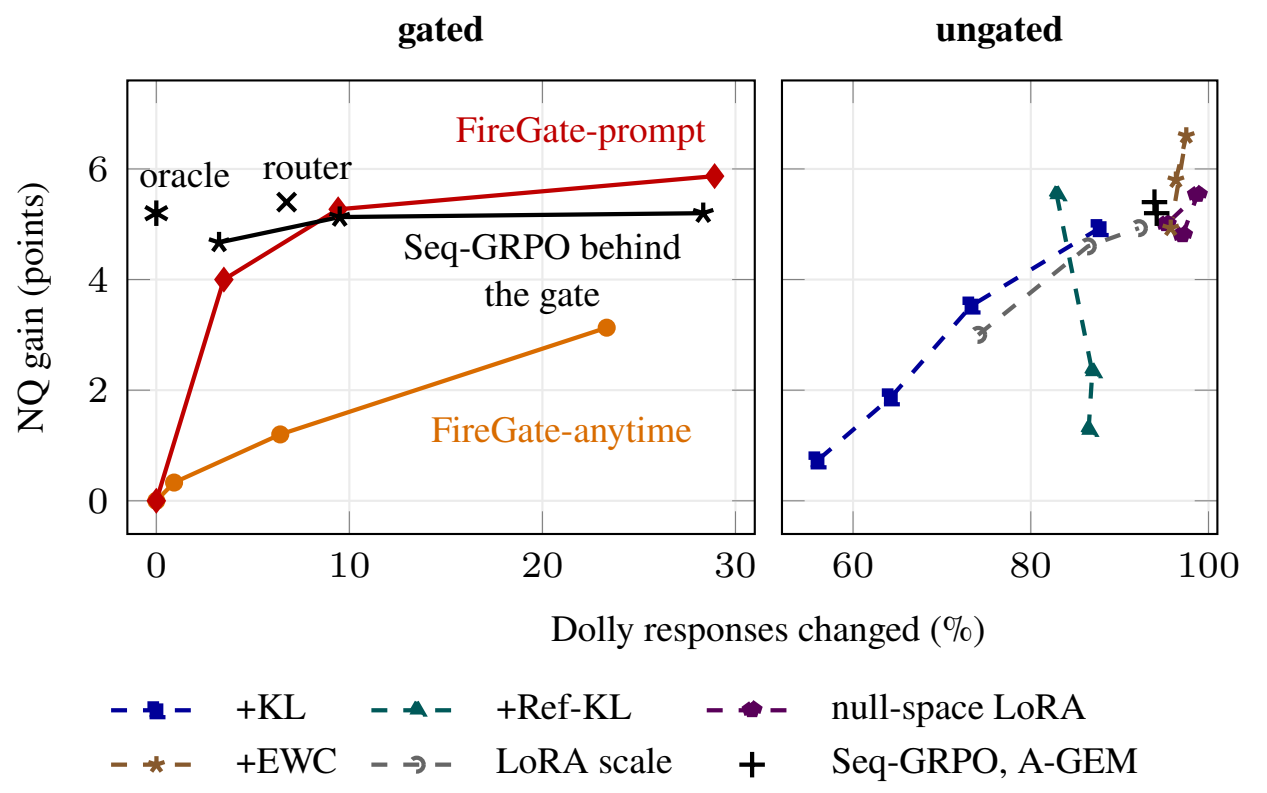

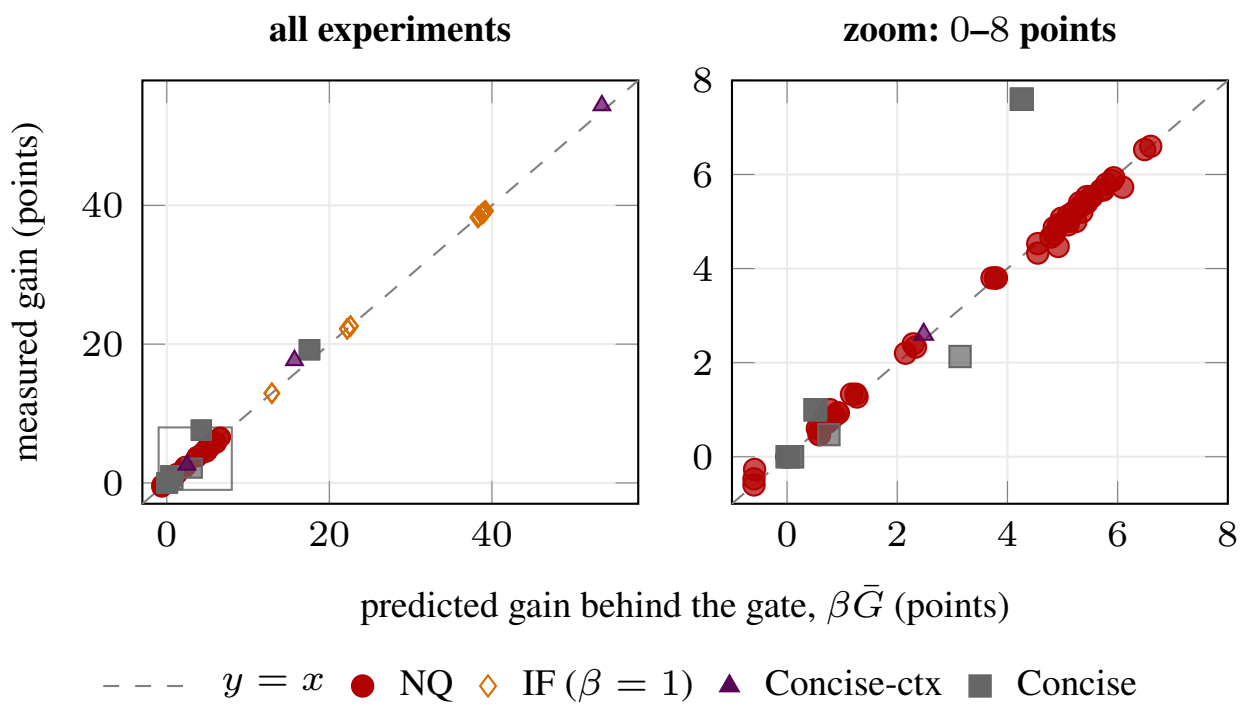
\begin{figure}[t]
\centering
\begin{tikzpicture}
\begin{groupplot}[group style={group size=2 by 1, horizontal sep=0.62cm},
  scale only axis, width=3.28cm, height=2.8cm, tick label style={font=\scriptsize}, label style={font=\scriptsize},
  title style={font=\scriptsize\bfseries, yshift=-0.8ex}, grid=major, grid style={black!8},
  scatter/classes={Nq={mark=*,red!70!black}, Nqseven={mark=*,red!70!black}, Nqolmo={mark=*,red!70!black},
    Ifc={mark=diamond,orange!85!black}, Ifcseven={mark=diamond,orange!85!black},
    Concisectx={mark=triangle*,violet!70!black}, Concise={mark=square*,gray!80!black}}]
\nextgroupplot[xmin=-3, xmax=58, ymin=-3, ymax=58, xtick={0,20,40}, ytick={0,20,40},
  ylabel={measured gain (points)}, title={all experiments}, legend to name=predleg, legend columns=5,
  legend style={font=\scriptsize, draw=none, column sep=3pt}, legend cell align=left]
\addplot[gray, dashed, domain=-3:58, samples=2] {x};
\addlegendentry{$y=x$}
\addplot[scatter, only marks, mark size=1.8pt, scatter src=explicit symbolic, forget plot, fill opacity=0.7]
  table[x=pred, y=actual, meta=exp, row sep=\\] {%
exp fam eps pred actual \\
Nq seq 0.03 4.79 4.67 \\
Nq seq 0.1 5.11 5.13 \\
Nq seq 0.3 5.20 5.20 \\
Nq kl 0.03 4.55 4.53 \\
Nq kl 0.1 4.85 4.87 \\
Nq kl 0.3 4.93 4.93 \\
Nq refkl 0.03 2.15 2.20 \\
Nq refkl 0.1 2.29 2.40 \\
Nq refkl 0.3 2.33 2.33 \\
Nq nullspace 0.03 5.10 5.00 \\
Nq nullspace 0.1 5.44 5.53 \\
Nq nullspace 0.3 5.53 5.53 \\
Nq ewc 0.03 6.09 5.73 \\
Nq ewc 0.1 6.49 6.53 \\
Nq ewc 0.3 6.60 6.60 \\
Nq agem 0.03 4.98 5.07 \\
Nq agem 0.1 5.31 5.40 \\
Nq agem 0.3 5.40 5.40 \\
Nq refkl_b0.1 0.03 5.10 4.93 \\
Nq refkl_b0.1 0.1 5.44 5.40 \\
Nq refkl_b0.1 0.3 5.53 5.53 \\
Nq refkl_b10 0.03 1.17 1.33 \\
Nq refkl_b10 0.1 1.25 1.33 \\
Nq refkl_b10 0.3 1.27 1.27 \\
Nq kl_b1 0.03 0.68 0.80 \\
Nq kl_b1 0.1 0.72 0.73 \\
Nq kl_b1 0.3 0.73 0.73 \\
Nq ewc_l0.1 0.03 5.35 5.20 \\
Nq ewc_l0.1 0.1 5.70 5.67 \\
Nq ewc_l0.1 0.3 5.80 5.80 \\
Nq ewc_l10 0.03 4.55 4.33 \\
Nq ewc_l10 0.1 4.85 4.73 \\
Nq ewc_l10 0.3 4.93 4.93 \\
Nq refkl_b10s4 0.03 0.55 0.60 \\
Nq refkl_b10s4 0.1 0.59 0.47 \\
Nq refkl_b10s4 0.3 0.60 0.60 \\
Nq refkl_b100s4 0.03 0.86 0.87 \\
Nq refkl_b100s4 0.1 0.92 0.93 \\
Nq refkl_b100s4 0.3 0.93 0.93 \\
Nq nullspace_rf25 0.03 4.92 4.47 \\
Nq nullspace_rf25 0.1 5.24 5.00 \\
Nq nullspace_rf25 0.3 5.33 5.33 \\
Nqseven seq 0.03 3.72 3.80 \\
Nqseven seq 0.1 3.78 3.80 \\
Nqseven seq 0.3 3.80 3.80 \\
Nqseven kl 0.03 0.78 1.00 \\
Nqseven kl 0.1 0.80 0.80 \\
Nqseven kl 0.3 0.80 0.80 \\
Nqseven refkl 0.03 -0.59 -0.27 \\
Nqseven refkl 0.1 -0.60 -0.47 \\
Nqseven refkl 0.3 -0.60 -0.60 \\
Nqseven nullspace 0.03 0.00 0.00 \\
Nqseven nullspace 0.1 0.00 0.00 \\
Nqseven nullspace 0.3 0.00 0.00 \\
Nqolmo seq 0.03 5.74 5.67 \\
Nqolmo seq 0.1 5.90 5.87 \\
Nqolmo seq 0.3 5.93 5.93 \\
Nqolmo refkl 0.03 0.58 0.47 \\
Nqolmo refkl 0.1 0.60 0.60 \\
Nqolmo refkl 0.3 0.60 0.60 \\
Ifc seq 0.03 38.73 38.73 \\
Ifc seq 0.1 38.73 38.73 \\
Ifc seq 0.3 38.73 38.73 \\
Ifc kl 0.03 38.40 38.40 \\
Ifc kl 0.1 38.40 38.40 \\
Ifc kl 0.3 38.40 38.40 \\
Ifc refkl 0.03 38.27 38.27 \\
Ifc refkl 0.1 38.27 38.27 \\
Ifc refkl 0.3 38.27 38.27 \\
Ifc nullspace 0.03 38.93 38.93 \\
Ifc nullspace 0.1 38.93 38.93 \\
Ifc nullspace 0.3 38.93 38.93 \\
Ifc refkl_b10 0.03 22.60 22.60 \\
Ifc refkl_b10 0.1 22.60 22.60 \\
Ifc refkl_b10 0.3 22.60 22.60 \\
Ifc seq_dense5e6 0.03 39.20 39.20 \\
Ifc seq_dense5e6 0.1 39.20 39.20 \\
Ifc seq_dense5e6 0.3 39.20 39.20 \\
Ifc seq_dense2e6 0.03 39.00 39.00 \\
Ifc seq_dense2e6 0.1 39.00 39.00 \\
Ifc seq_dense2e6 0.3 39.00 39.00 \\
Ifc seq_full5e6 0.03 39.00 39.00 \\
Ifc seq_full5e6 0.1 39.00 39.00 \\
Ifc seq_full5e6 0.3 39.00 39.00 \\
Ifc seq_sft 0.03 22.20 22.20 \\
Ifc seq_sft 0.1 22.20 22.20 \\
Ifc seq_sft 0.3 22.20 22.20 \\
Ifcseven seq 0.03 12.93 12.93 \\
Ifcseven seq 0.1 12.93 12.93 \\
Ifcseven seq 0.3 12.93 12.93 \\
Concise seq 0.03 0.52 1.00 \\
Concise seq 0.1 4.27 7.60 \\
Concise seq 0.3 17.59 19.20 \\
Concise kl 0.03 0.51 1.00 \\
Concise kl 0.1 4.25 7.60 \\
Concise kl 0.3 17.51 19.20 \\
Concise refkl 0.03 0.09 0.00 \\
Concise refkl 0.1 0.76 0.47 \\
Concise refkl 0.3 3.14 2.13 \\
Concise nullspace 0.03 0.51 1.00 \\
Concise nullspace 0.1 4.26 7.60 \\
Concise nullspace 0.3 17.57 19.20 \\
Concise refkl_b10 0.03 0.00 0.00 \\
Concise refkl_b10 0.1 0.02 0.00 \\
Concise refkl_b10 0.3 0.10 0.00 \\
Concisectx seq 0.03 2.48 2.60 \\
Concisectx seq 0.1 15.71 17.67 \\
Concisectx seq 0.3 53.53 54.40 \\
};
\addlegendimage{only marks, mark=*, red!70!black}\addlegendentry{NQ}
\addlegendimage{only marks, mark=diamond, orange!85!black}\addlegendentry{IF ($\beta=1$)}
\addlegendimage{only marks, mark=triangle*, violet!70!black}\addlegendentry{Concise-ctx}
\addlegendimage{only marks, mark=square*, gray!80!black}\addlegendentry{Concise}
\draw[black!50, thin] (axis cs:-1,-1) rectangle (axis cs:8,8);
\nextgroupplot[xmin=-1, xmax=8, ymin=-1, ymax=8, xtick={0,2,4,6,8}, ytick={0,2,4,6,8},
  title={zoom: $0$--$8$ points}, ylabel={}]
\addplot[gray, dashed, domain=-1:8, samples=2] {x};
\addplot[scatter, only marks, mark size=2pt, scatter src=explicit symbolic, fill opacity=0.7]
  table[x=pred, y=actual, meta=exp, row sep=\\] {%
exp fam eps pred actual \\
Nq seq 0.03 4.79 4.67 \\
Nq seq 0.1 5.11 5.13 \\
Nq seq 0.3 5.20 5.20 \\
Nq kl 0.03 4.55 4.53 \\
Nq kl 0.1 4.85 4.87 \\
Nq kl 0.3 4.93 4.93 \\
Nq refkl 0.03 2.15 2.20 \\
Nq refkl 0.1 2.29 2.40 \\
Nq refkl 0.3 2.33 2.33 \\
Nq nullspace 0.03 5.10 5.00 \\
Nq nullspace 0.1 5.44 5.53 \\
Nq nullspace 0.3 5.53 5.53 \\
Nq ewc 0.03 6.09 5.73 \\
Nq ewc 0.1 6.49 6.53 \\
Nq ewc 0.3 6.60 6.60 \\
Nq agem 0.03 4.98 5.07 \\
Nq agem 0.1 5.31 5.40 \\
Nq agem 0.3 5.40 5.40 \\
Nq refkl_b0.1 0.03 5.10 4.93 \\
Nq refkl_b0.1 0.1 5.44 5.40 \\
Nq refkl_b0.1 0.3 5.53 5.53 \\
Nq refkl_b10 0.03 1.17 1.33 \\
Nq refkl_b10 0.1 1.25 1.33 \\
Nq refkl_b10 0.3 1.27 1.27 \\
Nq kl_b1 0.03 0.68 0.80 \\
Nq kl_b1 0.1 0.72 0.73 \\
Nq kl_b1 0.3 0.73 0.73 \\
Nq ewc_l0.1 0.03 5.35 5.20 \\
Nq ewc_l0.1 0.1 5.70 5.67 \\
Nq ewc_l0.1 0.3 5.80 5.80 \\
Nq ewc_l10 0.03 4.55 4.33 \\
Nq ewc_l10 0.1 4.85 4.73 \\
Nq ewc_l10 0.3 4.93 4.93 \\
Nq refkl_b10s4 0.03 0.55 0.60 \\
Nq refkl_b10s4 0.1 0.59 0.47 \\
Nq refkl_b10s4 0.3 0.60 0.60 \\
Nq refkl_b100s4 0.03 0.86 0.87 \\
Nq refkl_b100s4 0.1 0.92 0.93 \\
Nq refkl_b100s4 0.3 0.93 0.93 \\
Nq nullspace_rf25 0.03 4.92 4.47 \\
Nq nullspace_rf25 0.1 5.24 5.00 \\
Nq nullspace_rf25 0.3 5.33 5.33 \\
Nqseven seq 0.03 3.72 3.80 \\
Nqseven seq 0.1 3.78 3.80 \\
Nqseven seq 0.3 3.80 3.80 \\
Nqseven kl 0.03 0.78 1.00 \\
Nqseven kl 0.1 0.80 0.80 \\
Nqseven kl 0.3 0.80 0.80 \\
Nqseven refkl 0.03 -0.59 -0.27 \\
Nqseven refkl 0.1 -0.60 -0.47 \\
Nqseven refkl 0.3 -0.60 -0.60 \\
Nqseven nullspace 0.03 0.00 0.00 \\
Nqseven nullspace 0.1 0.00 0.00 \\
Nqseven nullspace 0.3 0.00 0.00 \\
Nqolmo seq 0.03 5.74 5.67 \\
Nqolmo seq 0.1 5.90 5.87 \\
Nqolmo seq 0.3 5.93 5.93 \\
Nqolmo refkl 0.03 0.58 0.47 \\
Nqolmo refkl 0.1 0.60 0.60 \\
Nqolmo refkl 0.3 0.60 0.60 \\
Ifc seq 0.03 38.73 38.73 \\
Ifc seq 0.1 38.73 38.73 \\
Ifc seq 0.3 38.73 38.73 \\
Ifc kl 0.03 38.40 38.40 \\
Ifc kl 0.1 38.40 38.40 \\
Ifc kl 0.3 38.40 38.40 \\
Ifc refkl 0.03 38.27 38.27 \\
Ifc refkl 0.1 38.27 38.27 \\
Ifc refkl 0.3 38.27 38.27 \\
Ifc nullspace 0.03 38.93 38.93 \\
Ifc nullspace 0.1 38.93 38.93 \\
Ifc nullspace 0.3 38.93 38.93 \\
Ifc refkl_b10 0.03 22.60 22.60 \\
Ifc refkl_b10 0.1 22.60 22.60 \\
Ifc refkl_b10 0.3 22.60 22.60 \\
Ifc seq_dense5e6 0.03 39.20 39.20 \\
Ifc seq_dense5e6 0.1 39.20 39.20 \\
Ifc seq_dense5e6 0.3 39.20 39.20 \\
Ifc seq_dense2e6 0.03 39.00 39.00 \\
Ifc seq_dense2e6 0.1 39.00 39.00 \\
Ifc seq_dense2e6 0.3 39.00 39.00 \\
Ifc seq_full5e6 0.03 39.00 39.00 \\
Ifc seq_full5e6 0.1 39.00 39.00 \\
Ifc seq_full5e6 0.3 39.00 39.00 \\
Ifc seq_sft 0.03 22.20 22.20 \\
Ifc seq_sft 0.1 22.20 22.20 \\
Ifc seq_sft 0.3 22.20 22.20 \\
Ifcseven seq 0.03 12.93 12.93 \\
Ifcseven seq 0.1 12.93 12.93 \\
Ifcseven seq 0.3 12.93 12.93 \\
Concise seq 0.03 0.52 1.00 \\
Concise seq 0.1 4.27 7.60 \\
Concise seq 0.3 17.59 19.20 \\
Concise kl 0.03 0.51 1.00 \\
Concise kl 0.1 4.25 7.60 \\
Concise kl 0.3 17.51 19.20 \\
Concise refkl 0.03 0.09 0.00 \\
Concise refkl 0.1 0.76 0.47 \\
Concise refkl 0.3 3.14 2.13 \\
Concise nullspace 0.03 0.51 1.00 \\
Concise nullspace 0.1 4.26 7.60 \\
Concise nullspace 0.3 17.57 19.20 \\
Concise refkl_b10 0.03 0.00 0.00 \\
Concise refkl_b10 0.1 0.02 0.00 \\
Concise refkl_b10 0.3 0.10 0.00 \\
Concisectx seq 0.03 2.48 2.60 \\
Concisectx seq 0.1 15.71 17.67 \\
Concisectx seq 0.3 53.53 54.40 \\
};
\end{groupplot}
\node[font=\scriptsize, anchor=north] at ($(group c1r1.south east)!0.5!(group c2r1.south west) + (0,-0.42)$)
  {predicted gain behind the gate, $\beta\bar G$ (points)};
\end{tikzpicture}\par\ref{predleg}
\caption{Gain kept behind the prompt-time gate against $\beta\bar G$ (\cref{prop:switchgain}; $\beta$ measured before
training, $\bar G$ the trained model's gain), every ungated method at three levels, mean over seeds; right: the box on
the left. On IF $\beta=1$; on the 78 other points the mean error is 0.3 points (0.5 where $\beta<1$), the largest 3.4.}
\label{fig:switchpred}
\end{figure}

\textbf{Ungated updates change most reference responses:} \Cref{fig:frontier} places every method on NQ by its gain and
the share of Dolly responses it changes. Every ungated setting changes most Dolly responses, though none is trained on
them, and even the strongest regularisation leaves more than half changed, while gates that certify Dolly change only
a few. In these runs, regularisers slow the drift but do not confine it, since they act on the weights, not on where the
update applies. Average scores hide this drift: in \cref{tab:main}, Seq-GRPO leaves refusal and MMLU almost unchanged on the
three-task stream but lowers over-refusal on XSTest by 15.5 points. Full-rank updates and SFT follow the same pattern
in \cref{tab:pilot}, since the certificate depends only on where the update acts.

\begin{table}[t]
\centering
\footnotesize
\setlength{\tabcolsep}{2.5pt}
\caption{The switch on seven models from five families (three seeds, $\varepsilon=0.03$). $\bar G$, Chg.: Seq-GRPO's
gain (points) and changed Dolly responses (\%); $\beta$: firing on unlabeled RL prompts, before training; Kept: the
gain behind the deployed prompt-time gate; CP: 95\% Clopper--Pearson bound of its held-out Dolly firing (\%);
Non-fired: changed non-fired responses of the \methodp{} runs; $^\dagger$two of three runs.}
\label{tab:fam}
\resizebox{\ifdim\width>\linewidth\linewidth\else\width\fi}{!}{%
\begin{tabular}{l l rr r rrrr r}
\toprule
 & & \multicolumn{2}{c}{ungated} & & \multicolumn{4}{c}{behind the gate} & \\
\cmidrule(lr){3-4}\cmidrule(lr){6-9}
Model & Task & $\bar G$ & Chg. & $\beta$ & $\beta\bar G$ & Kept$\uparrow$ & Chg.$\downarrow$ & CP$\downarrow$ & Non-fired \\
\midrule
Qwen2.5-1.5B & IF & +38.7 & 93.3 & 100.0 & +38.7 & +38.7 & 1.5 & 2.9 & 0/4{,}797 \\
Qwen2.5-1.5B & NQ & +5.2 & 94.2 & 92.2 & +4.8 & +4.7 & 3.2 & 5.1 & 0/9{,}363 \\
Qwen2.5-7B & IF & +12.9 & 94.6 & 100.0 & +12.9 & +12.9 & 0.8 & 3.3 & 0/3{,}531 \\
Qwen2.5-7B & NQ & +3.8 & 94.6 & 97.9 & +3.7 & +3.8 & 2.8 & 4.5 & 0/9{,}310 \\
OLMo-2-7B & NQ & +5.9 & 91.2 & 96.8 & +5.7 & +5.7 & 2.2 & 3.9 & 0/9{,}303 \\
Llama-3.2-3B & IF & +14.3 & 94.1 & 100.0 & +14.3 & +14.3 & 1.7 & 3.3 & 0/3{,}783 \\
Llama-3.2-3B & NQ & +10.8 & 96.7 & 89.0 & +9.6 & +9.2 & 2.5 & 4.2 & 0/9{,}447 \\
Llama-3.1-8B & IF & +11.4 & 95.5 & 100.0 & +11.4 & +11.4 & 1.8 & 3.3 & 0/3{,}957 \\
Llama-3.1-8B & NQ & +7.1 & 99.9 & 95.5 & +6.7 & +6.8 & 2.5 & 4.2 & 0/9{,}342 \\
DeepSeek-LLM-7B & IF & +35.3 & 99.5 & 99.9 & +35.2 & +35.3 & 2.3 & 3.9 & 0/3{,}429 \\
DeepSeek-LLM-7B & NQ & +10.7 & 97.3 & 84.5 & +9.1 & +8.7 & 1.3 & 2.6 & 0/9{,}519 \\
Mistral-7B & IF & +34.8 & 98.5 & 100.0 & +34.8 & +34.8 & 1.0 & 2.3 & 0/4{,}008 \\
Mistral-7B & NQ & +9.5 & 98.8 & 93.0 & +8.8 & +7.9 & 2.3 & 3.9 & 0/6{,}270$^\dagger$ \\
\bottomrule
\end{tabular}}
\end{table}

\textbf{The gain kept follows the coverage:} \Cref{tab:fam} places Seq-GRPO behind the deployed gate of seven models.
On IF, whose prompts the gate tells apart from chat, it fires on nearly every new-task prompt and the switch keeps all
of the gain; on NQ, where the tasks overlap, the gain kept shrinks with the coverage. \Cref{fig:switchpred} shows that
across all experiments the gain kept stays close to the coverage times the ungated gain. The share of the gain that the
chosen gate costs can therefore be predicted before training, from its firing on unlabeled task prompts. This is a
prediction, not a certificate: \cref{prop:switchgain} leaves a covariance term open. On the stored runs,
\rvbAllCovZero{} of \rvbAllCovN{} bootstrap intervals of the prediction error contain zero, and the largest error is
\rvbAllCovErrMax{} points.

\begin{table}[t]
\centering
\scriptsize
\setlength{\tabcolsep}{1.6pt}
\caption{Five-task stream, Qwen2.5-1.5B, three seeds. Avg, BWT: mean accuracy after the stream (base 29.9\%) and
backward transfer; Dolly: held-out responses changed (\%); GB, time: peak memory and batched generation time against
the base model, one or all five LoRAs firing. One gate: the union of the five two-sided gates, calibrated once; it fires
on none of 400 fresh Dolly prompts, which bounds its Dolly firing by 0.7\% with probability 0.95.
--: not measured.}
\label{tab:deploy}
\begin{tabular}{l rr r r r}
\toprule
Deployment & Avg$\uparrow$ & BWT$\uparrow$ & Dolly$\downarrow$ & GB & Time \\
\midrule
Seq-GRPO, ungated & 46.5$\pm$1.7 & $-$3.5$\pm$0.7 & 91.8 & 8.1 & $\times$1.00 \\
\quad behind the gate & 46.5$\pm$1.7 & $-$3.5$\pm$0.7 & 4.8 & 10.9 & $\times$1.18 \\
\quad behind one gate & 46.3$\pm$1.8 & -- & 0.0 & -- & -- \\
\methodp{}, gated training & 49.6$\pm$2.9 & +0.0$\pm$0.0 & 3.3 & 12.5 & $\times$1.25--2.59 \\
\bottomrule
\end{tabular}
\end{table}

\textbf{Choice of score and block:} \Cref{fig:score} compares gate scores before any training. Scores that see both
the task and the reference, the logistic classifier and a Gaussian density ratio, separate NQ from Dolly at every depth
and cover most NQ prompts at the 3\% level, while nearest-neighbour similarity and text $n$-grams cover less. A
one-class score that sees only the reference covers no NQ prompt, so a gate needs both pools. The block matters little:
moving the split from block 5 to block 15 barely changes the drift or the gain.

\begin{figure}[t]
\centering
\begin{tikzpicture}
\begin{axis}[scale only axis, width=7.0cm, height=2.1cm, xmin=-4, xmax=101, ymin=45, ymax=62,
  xtick={0,20,40,60,80,100}, ytick={46,50,54,58,62},
  xlabel={Dolly responses changed (\%)}, ylabel={average accuracy (\%)},
  tick label style={font=\scriptsize}, label style={font=\scriptsize}, grid=major, grid style={black!8},
  legend style={font=\scriptsize, draw=none, at={(0.5,1.03)}, anchor=south, column sep=4pt}, legend columns=3,
  legend cell align=left, clip=false]
\addplot[only marks, mark=diamond*, mark size=2.3pt, red!75!black]
  coordinates {(\resChgTwpDolly,\resAvgTwp) (\resChgTwsDolly,\resAvgTws) (\resChgTwDolly,\resAvgTw)};
\addlegendentry{\method{}, gated}
\addplot[only marks, mark=x, mark size=2.6pt, black] coordinates {(\resChgRouterDolly,\resAvgRouter)};
\addlegendentry{router, uncalibrated}
\addplot[only marks, mark=square*, mark size=1.5pt, blue!60!black]
  coordinates {(\resChgSeqDolly,\resAvgSeq) (\resChgKlDolly,\resAvgKl) (\resChgReplayDolly,\resAvgReplay)
  (\resChgRefklDolly,\resAvgRefkl) (\resChgOloraDolly,\resAvgOlora) (\resChgNullDolly,\resAvgNull)
  (\resChgScaleDolly,\resAvgScale) (94.2,58.4)};
\addlegendentry{ungated}
\node[font=\scriptsize, text=red!75!black, anchor=south west, inner sep=1pt] at (axis cs:3.5,59.2) {\methodp{}, \methoda{}};
\node[font=\scriptsize, anchor=north west, inner sep=1pt] at (axis cs:3.5,56.9) {router};
\node[font=\scriptsize, text=blue!60!black, anchor=east, inner sep=2pt] at (axis cs:\resChgNullDolly,\resAvgNull) {null-space};
\node[font=\scriptsize, text=blue!60!black, anchor=east, inner sep=2pt] at (axis cs:\resChgRefklDolly,\resAvgRefkl) {Ref-KL};
\node[font=\scriptsize, text=blue!60!black, anchor=east, inner sep=2pt] at (axis cs:\resChgScaleDolly,\resAvgScale) {LoRA scale};
\end{axis}
\end{tikzpicture}
\caption{Three-task stream, Qwen2.5-1.5B: average accuracy after the stream against held-out Dolly responses changed,
mean of three seeds (A-GEM one); base model 31.5\%. Ungated: Seq-GRPO, +KL, +Replay, +Ref-KL, O-LoRA, null-space LoRA,
A-GEM, LoRA scale (\cref{tab:main}).}
\label{fig:stream}
\end{figure}
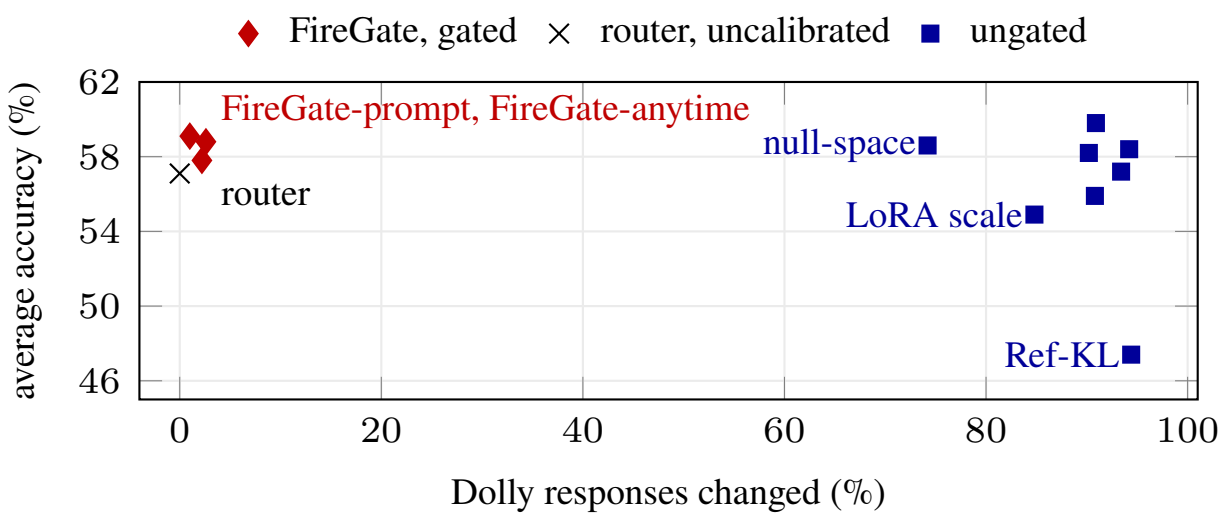

\textbf{Gated training on streams:} \Cref{tab:deploy} shows that gated training keeps the earlier tasks with no
backward forgetting, because their LoRAs and gates are frozen and fire only on their own prompts, while the switch
inherits the trained model's forgetting. Its higher average comes mostly from seeds in which Seq-GRPO loses Countdown
or K\&K. In \cref{fig:stream}, every ungated method changes most Dolly responses on the three-task stream, while
\method{} changes almost none at about the same average accuracy, so the protection does not cost the tasks. At matched
memory and routing, the comparison is the router, which uses the same features and LoRAs without calibration: it
changes no chat on the three-task stream, but on NQ it fires on \ovlRouterFireDolly\% of Dolly prompts, a rate that no
level fixes before training. The switch needs a second model in memory, while gated training does not. Gated training has a further
cost: an earlier stage's gate can fire on a later task's prompts before that task is trained, which lowers Countdown
accuracy after the GSM8K stage in \cref{tab:main}.

\textbf{Protecting chosen behaviours:} The certified gain falls as the new prompts look more like the reference, as
\cref{thm:limit} predicts: on SQuAD~2.0, a task that overlaps general chat, Seq-GRPO behind the
deployed gate keeps about three quarters of its gain. If chat is what a task should change, we calibrate only on what
must stay and add harmful requests that follow a passage as a component. Seq-GRPO behind this gate then keeps almost
all of its gain, and the gate stays within the level on such held-out requests, as \cref{tab:protect} shows. Ref-KL and
safety replay hold the rates they target, not the responses. On Concise, whose prompts are chat requests, the deployed
gate covers almost none of the task, so the recipe tells before training that certification is too costly there.

\subsection{Beyond the Calibrated Prompts}
\label{sec:res-gates}

\begin{table}[t]
\centering
\scriptsize
\setlength{\tabcolsep}{1.8pt}
\caption{Every setting frozen: firing (\%) of the chosen gate on HarmBench (HB, $n=200$), its contextual requests (ctx.,
100) and StrongREJECT (SR, 288). Base: the base model's HB refusal (\%); changes: points, seed 0. Con., Ctx:
Concise(-ctx); O, L: OLMo-2, Llama-3.2.}
\label{tab:extmain}
\begin{tabular}{l rrrrrrrrr}
\toprule
 & Con. & Ctx & SQuAD & NQ & IF & IF 7B & 3 tasks & NQ O & NQ L \\
\midrule
HB / SR fires$\downarrow$ & 0/4.2 & 0/0 & 0/0 & 0/0 & 0/0 & 0/0 & 0/0 & 0/0 & 0/0 \\
ctx. fires$\downarrow$ & 9.0 & 1.0 & 0 & 0 & 0 & 0 & 0 & 0 & 0 \\
HB refusal, base & 93 & 93 & 93 & 93 & 93 & 86 & 93 & 96 & 88 \\
HB refusal, ungated & $-$79 & $-$51 & $-$2 & $-$41 & $-$4 & +3 & $-$12 & $-$2 & $-$25 \\
HB refusal, gated & +0 & +0 & +0 & +0 & +0 & +0 & +0 & +0 & +0 \\
\bottomrule
\end{tabular}
\end{table}

\textbf{Pools unlike the task:} A classifier trained only to tell the task from the reference can fire on prompts
unlike both. The deployed IF gate fires on almost every MMLU prompt, whose answer-format instruction resembles IF
prompts, and the IF and stream gates on many harmful requests in jailbreak templates. The two-sided gate, which also
requires a prompt to lie in the task's region, fires on none of them in any experiment and keeps most of its coverage,
as \cref{tab:gate_overview} shows. The region test was added after the MMLU failure was seen, so MMLU does not test it
independently; the jailbreak templates were added later and do (\cref{tab:protocol}). Real user requests from WildChat \citep{zhao2024wildchat}, never fitted or calibrated
on, show the same: the deployed IF and stream gates fire on up to 30.8\% of them, the two-sided gate on at most 1.2\%.

\textbf{Benchmarks written by others:} \Cref{tab:extmain} tests the chosen gates on harmful benchmarks written by
others, HarmBench and StrongREJECT \citep{mazeika2024harmbench,souly2024strongreject}, with every setting frozen and
predictions written beforehand. Outside Concise, the gates fire on almost none of these requests, while ungated
Seq-GRPO loses much of its refusal on them; the certificate thus carries over to harmful requests that no component
sampled, as long as they read like the sampled ones. On Concise, whose prompts are chat requests like the harmful ones,
the gates fire above the predicted levels. On ten more sets, among them MultiJail and oasst1, all four predictions
hold, and outside Concise the chosen gates fire on less than 2\% of each, as \cref{tab:prospective} shows.

\begin{table}[t]
\centering
\footnotesize
\setlength{\tabcolsep}{2.6pt}
\caption{Audit of the fired prompts on all stored harmful prompts; $n$: prompts $\times$ seeds.
Fires: the gate's firing (\%); New: prompts the base model refuses and Seq-GRPO behind the gate answers (\% of $n$).}
\label{tab:audit}
\begin{tabular}{l r rr rr}
\toprule
 & & \multicolumn{2}{c}{chosen gate} & \multicolumn{2}{c}{deployed gate} \\
\cmidrule(lr){3-4}\cmidrule(lr){5-6}
Experiment & $n$ & Fires & New$\downarrow$ & Fires & New$\downarrow$ \\
\midrule
NQ, Qwen 1.5B & 2{,}438 & 0.5 & 0.16 & 0.5 & 0.16 \\
NQ, OLMo-2 7B & 1{,}938 & 0.5 & 0.00 & 0.5 & 0.00 \\
NQ, Llama 3B & 1{,}938 & 0.0 & 0.00 & 0.0 & 0.00 \\
IF, Qwen 1.5B & 6{,}438 & 0.0 & 0.00 & 70.0 & 37.96 \\
IF, Qwen 7B & 3{,}938 & 0.0 & 0.00 & 53.4 & 4.14 \\
3 tasks, Qwen 1.5B & 5{,}688 & 0.0 & 0.00 & 53.2 & 16.65 \\
SQuAD~2.0, Qwen 1.5B & 2{,}938 & 0.2 & 0.17 & 0.1 & 0.03 \\
Concise-ctx, Qwen 1.5B & 2{,}438 & 0.0 & 0.00 & 1.7 & 0.49 \\
Concise, Qwen 1.5B & 2{,}438 & 6.5 & 5.95 & 2.6 & 1.31 \\
\bottomrule
\end{tabular}
\end{table}

\begin{table}[t]
\centering
\scriptsize
\setlength{\tabcolsep}{3pt}
\caption{TriviaQA with Llama-3.2-3B, a task no earlier experiment used (three seeds): every setting copied from NQ,
every prediction written before the data were downloaded, and all seven hold. Values in \%, P5 in points; P1--P5: the
deployed gate, P6, P7: the chosen (two-sided) gate.}
\label{tab:trivia}
\begin{tabular}{@{}l l l@{}}
\toprule
& Prediction & Measured \\
\midrule
P1 & no change where the gate is silent & 0 of 9{,}405 \\
P2 & change $\le$ firing, behind the gate & 24 of 24 checks \\
P3 & held-out firing consistent with 3\% & Dolly 3.5, AdvBench 0.0; 10k 2.0 \\
P4 & ungated changes $>80\%$ of Dolly & 95.8, 98.5, 93.2 \\
P5 & kept within 1 point of $\beta\bar G$ & +23.1 vs +22.3 \\
P6 & external sets: $\le$3/5\%; refusal $\pm$3 & 0.0, 0.0, 0.0; +0.0, +0.0, +0.0 \\
P7 & others' prompt sets: $\le$3--10\% & at most 0.8 \\
\bottomrule
\end{tabular}
\end{table}

\textbf{Auditing the fired prompts:} The certificate allows any change where a gate fires, so \cref{tab:audit} audits
exactly those prompts across every stored harmful set. Behind the chosen gates, Seq-GRPO newly answers almost none of
these prompts. Behind the deployed IF gate it newly answers many, mostly harmful requests written in IF's format, which
the two-sided gate and the added component remove. On Concise, most of the few that pass are XSTest's unsafe prompts,
which no component samples. The audit thus shows where a component is missing.

\textbf{A fallback for one named behaviour:} Where a gate fires, the certificate allows any change. For a behaviour
that a check on the old model's reply can recognise, such as refusal, a fallback closes this gap: on a fired prompt
the old model's reply is generated first, and if the refusal check flags it, that reply is returned. By construction,
no prompt that the old model refuses is then answered. On the stored responses (\cref{tab:fallback},
\cref{app:fallback}), the fallback
removes the leak behind the deployed IF and stream gates, 38.0\% and 16.7\% of harmful prompts, and leaves their task
accuracy unchanged; it does the same for the models trained under these gates. Without a gate the fallback restores
refusal as well, but most benign replies still change: 69 to 95\% in these experiments, and 73.6\% of chat episodes
for the ungated agent of \cref{sec:res-api}. The fallback thus guards one named behaviour, and the gate bounds the
rest.

\textbf{A wider harmful component for Concise:} Concise's harmful component samples only English AdvBench requests,
so a harmful request written differently reads as chat to its chosen gate, which fires on many MultiJail requests in
\cref{tab:prospective}. Adding one component that samples harmful requests in more forms and ten languages fixes this,
and all five predictions written before any download hold in \cref{tab:widen}: the gate fires on almost no harmful
request of any source, even in three sets it never saw, and Seq-GRPO behind it newly answers far fewer. The price is
coverage, which falls by more than half, since excluding more harmful chat also excludes Concise prompts.

\textbf{A new task with the recipe frozen:} TriviaQA \citep{joshi2017triviaqa} tests the recipe on a task first used
after every setting was fixed. All seven predictions written before its data were downloaded hold in \cref{tab:trivia},
including the gain kept behind the gate, which lands within a point of the value predicted before training. The recipe
therefore carries over to a new task without retuning.

\begin{table}[t]
\centering
\scriptsize
\setlength{\tabcolsep}{2.4pt}
\caption{Inside a calculator agent (Qwen2.5-1.5B, seed 0; up to four calls, 250 episodes per set); the agent-aware
gate decides at every call. Fires: episodes in which some call fires (\%), offline from the old agent's calls / live;
CP: 95\% bound on Dolly episodes; Chg.: Dolly episodes that Seq-GRPO behind the gate changes (\%); Gain: task gain
behind the gate / ungated (points). No episode without a fired call changed (0 of 3{,}742). On NQ task episodes,
offline and live firing differ on 4 of 250 episodes, all near the threshold.}
\label{tab:agent}
\begin{tabular}{l rrr r rr}
\toprule
 & \multicolumn{3}{c}{Fires (\%)} & & & \\
\cmidrule(lr){2-4}
Task & Dolly & AdvB.\ turn 2 & Task & CP & Chg. & Gain \\
\midrule
NQ & 1.2 / 1.6 & 0.0 & 81.2 / 80.4 & 3.6 & 1.2 & +3.6 / +5.6 \\
IF & 0.0 & 0.0 & 100 & 1.2 & 0.0 & +30.8 / +34.4 \\
\bottomrule
\end{tabular}
\end{table}

\textbf{Inside a tool-using agent:} \Cref{tab:agent} runs the gate at every model call of a calculator agent
(\cref{cor:episode}). Gates fitted on plain prompts do not carry over: the deployed IF gate fires on 96\% of Dolly
episodes. Refitted with the agent's first calls as components, the gate stays within the level on chat and harmful
episodes, covers most task episodes and keeps most of Seq-GRPO's gain, and no episode without a fired call changed.
On real traffic replayed in its arrival order, the monitor caught every gate that fires above its level within 8 to
230 prompts, with false alarms on at most 0.3\% of null streams, and raising the thresholds on the next prompts
brought these gates back below the level. \Cref{app:agents} gives the details.

\subsection{Learning Inside the Agent Loop}
\label{sec:res-agentloop}

The test above places models trained on plain prompts inside the agent. Here the agent itself learns: every training
episode runs inside its tool loop, and the gate decides at every model call (\cref{cor:episode}).

\textbf{Setup:} The agent is a chat model with a calculator tool and up to four model calls per episode. It learns two
tasks in turn by GRPO on whole episodes, with the training settings of \cref{sec:setup-main}. The first task, Calc,
is word problems that multiply numbers of up to four digits; its reward checks only the final answer and does not
ask for the tool. The second is Knights and Knaves (KK), which needs no tool. Seq-GRPO trains an ungated LoRA per task.
\methodp{} puts each LoRA behind a gate fitted on the calls of the previous agent's episodes and calibrated on the
episode score at $\varepsilon=0.03$; when KK is learned, the stored Calc problems are a protected component. We run
Qwen2.5-1.5B with three seeds and Qwen2.5-7B with one. After each task we play 250 greedy episodes per set: Calc, KK,
held-out Dolly requests, and held-out AdvBench requests that arrive as a second user turn after a tool exchange
(\cref{app:agentloop}).

\begin{table*}[t]
\centering
\scriptsize
\setlength{\tabcolsep}{4.5pt}
\caption{An agent trained inside its calculator loop on Calc and then KK: every run after both tasks (250 greedy
episodes per set, in \%). Tool: Calc episodes with a calculator call. Changed: episodes whose transcript differs, for
Calc from the same agent before KK, for Dolly from the base agent. Refusal: harmful requests arriving mid-session.
Fires: episodes in which the KK gate fires at some call, with the 95\% bound on Dolly. Unfired changed: episodes that
changed against the agent before KK / episodes without a fired call. Base agent: Calc 27.2 and KK 8.8 at 1.5B, 95.6
and 9.2 at 7B; refusal 98.8.}
\label{tab:agentloop}
\begin{tabular}{l c l rrr rr r rr r}
\toprule
 & & & \multicolumn{3}{c}{Accuracy and tool use} & \multicolumn{2}{c}{Changed} & & \multicolumn{2}{c}{Fires} & \\
\cmidrule(lr){4-6}\cmidrule(lr){7-8}\cmidrule(lr){10-11}
Model & Seed & Method & Calc & Tool & KK & Calc & Dolly & Refusal & Dolly [bound] & Calc & Unfired changed \\
\midrule
Qwen2.5-1.5B & 0 & Seq-GRPO & 11.6 & 11.6 & 38.4 & 92.8 & 100 & 98.0 & -- & -- & -- \\
 & & \methodp{} & 100 & 100 & 36.8 & 0.0 & 1.2 & 98.8 & 0.8 [2.5] & 0.0 & 0 / 748 \\
 & 1 & Seq-GRPO & 100 & 100 & 38.4 & 100 & 100 & 97.2 & -- & -- & -- \\
 & & \methodp{} & 97.6 & 100 & 39.2 & 0.0 & 1.2 & 98.8 & 0.8 [2.5] & 0.0 & 0 / 748 \\
 & 2 & Seq-GRPO & 98.8 & 100 & 38.4 & 100 & 100 & 98.8 & -- & -- & -- \\
 & & \methodp{} & 97.6 & 100 & 36.0 & 0.0 & 1.2 & 98.8 & 0.8 [2.5] & 0.0 & 0 / 748 \\
\midrule
Qwen2.5-7B & 0 & Seq-GRPO & 97.6 & 100 & 66.8 & 89.2 & 95.2 & 93.2 & -- & -- & -- \\
 & & \methodp{} & 97.6 & 100 & 63.2 & 0.0 & 1.2 & 98.8 & 0.8 [2.5] & 0.0 & 0 / 748 \\
\bottomrule
\end{tabular}
\end{table*}

\begin{figure}[t]
\centering
\begin{tikzpicture}
\begin{groupplot}[group style={group size=2 by 1, horizontal sep=0.3cm, y descriptions at=edge left},
  scale only axis, width=3.35cm, height=2.9cm, xmin=-0.5, xmax=1.5, ymin=-6, ymax=108, ytick={0,25,50,75,100},
  xtick={0,1}, x tick style={draw=none}, ymajorgrids, grid style={black!8},
  tick label style={font=\scriptsize}, label style={font=\scriptsize},
  title style={font=\scriptsize\bfseries, yshift=-0.8ex}]
\nextgroupplot[xticklabels={Calc, KK}, ylabel={episodes (\%)}, title={\strut correct after both tasks}]
\addplot[only marks, mark=square*, mark size=1.5pt, blue!60!black]
  coordinates {(-0.29,11.6) (-0.21,100) (-0.13,98.8) (0.71,38.4) (0.79,38.4) (0.87,38.4)};
\addplot[only marks, mark=square, mark size=1.5pt, line width=0.7pt, blue!60!black] coordinates {(-0.05,97.6) (0.95,66.8)};
\addplot[only marks, mark=diamond*, mark size=2.2pt, red!75!black]
  coordinates {(0.07,100) (0.15,97.6) (0.23,97.6) (1.07,36.8) (1.15,39.2) (1.23,36.0)};
\addplot[only marks, mark=diamond, mark size=2.2pt, line width=0.7pt, red!75!black] coordinates {(0.31,97.6) (1.31,63.2)};
\node[font=\scriptsize, anchor=south, inner sep=2.5pt] at (axis cs:-0.2,11.6) {seed 0};
\nextgroupplot[xticklabels={Calc, Dolly}, title={\strut changed by learning KK},
  legend to name=agentloopleg, legend columns=3, legend style={font=\scriptsize, draw=none, column sep=4pt},
  legend cell align=left]
\addplot[only marks, mark=square*, mark size=1.5pt, blue!60!black]
  coordinates {(-0.29,92.8) (-0.21,100) (-0.13,100) (0.71,100) (0.79,100) (0.87,100)};
\addlegendentry{Seq-GRPO}
\addplot[only marks, mark=diamond*, mark size=2.2pt, red!75!black]
  coordinates {(0.07,0.0) (0.15,0.0) (0.23,0.0) (1.07,1.2) (1.15,1.2) (1.23,1.2)};
\addlegendentry{\methodp{}}
\addplot[only marks, mark=square, mark size=1.5pt, line width=0.7pt, blue!60!black, forget plot] coordinates {(-0.05,89.2) (0.95,95.2)};
\addplot[only marks, mark=diamond, mark size=2.2pt, line width=0.7pt, red!75!black, forget plot] coordinates {(0.31,0.0) (1.31,1.2)};
\addlegendimage{only marks, mark=square, mark size=1.5pt, line width=0.7pt, black}
\addlegendentry{open: Qwen2.5-7B}
\end{groupplot}
\end{tikzpicture}\par\ref{agentloopleg}
\caption{Training inside the agent loop: every run after both tasks (\cref{tab:agentloop}). Filled marks: the three
seeds of Qwen2.5-1.5B; open marks: Qwen2.5-7B. Left: episodes answered correctly. Right: Calc episodes that differ from
the same agent's before KK, and Dolly episodes that differ from the base agent's.}
\label{fig:agentloop}
\end{figure}
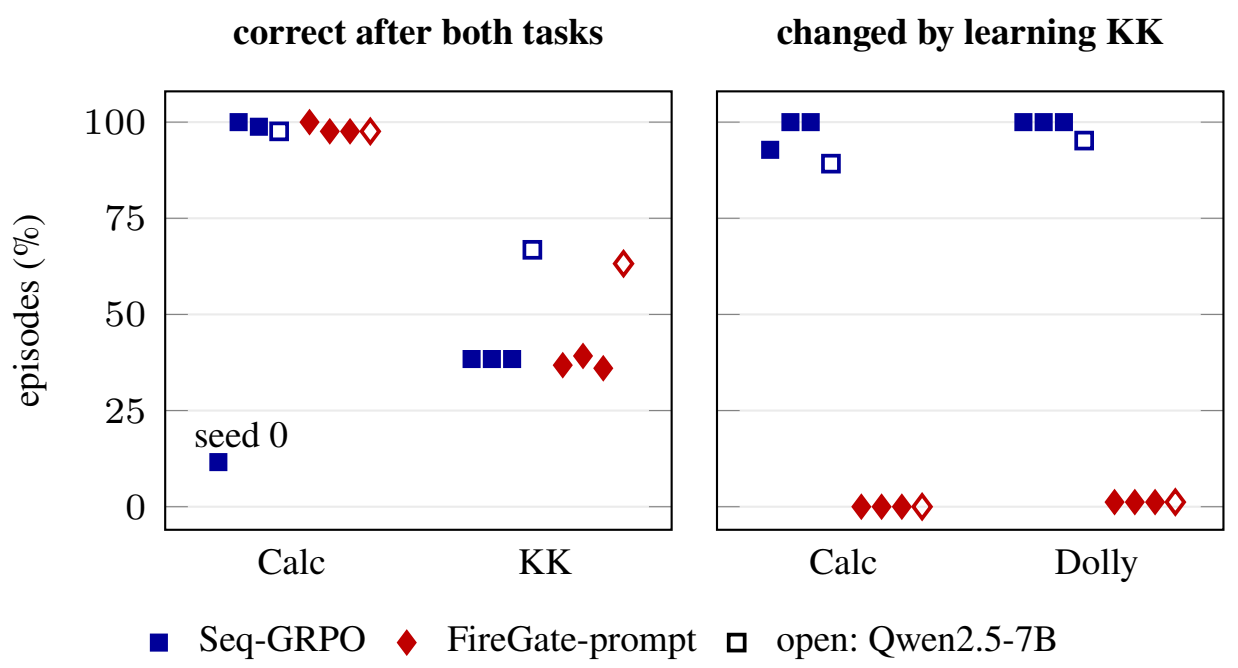

\textbf{No episode changed without a firing under training in the loop:} In all four runs, no episode changed without a
fired call: 0 of 749 episodes after Calc (0 of 740 at 7B) and 0 of 748 after KK (\cref{tab:agentloop}). The KK gate
fired on no Calc episode, so every Calc episode after KK is the episode from before KK, token for token. It fired on
0.8\% of Dolly episodes (95\% bound 2.5\%, level 3\%), and after both tasks 1.2\% of Dolly episodes differ from the
base agent's.

\textbf{Ungated training rewrites the earlier skill even when its score stays:} After KK, Seq-GRPO changed 89 to
100\% of Calc episodes and 95 to 100\% of Dolly episodes (\cref{fig:agentloop}). In three of the four runs its Calc
accuracy stayed at 97.6\% or above, so the score alone would report nothing. In the fourth, at 1.5B with seed 0, the
agent largely stopped calling the calculator: tool use on Calc fell from 100 to 11.6\% and accuracy with it, below the
untrained agent's 27.2\%. Without the gate, whether the tool skill survived the next task depended on the seed; behind
the gate it held in every run.

\textbf{Ungated training spreads the tool habit:} After Calc, the ungated agents at 1.5B call the calculator on more
than 80\% of Dolly requests and on nearly every KK puzzle, where the base agent calls it on 24.8\% and on none
(\cref{tab:agentloop_tools}). At 7B the rates rise as well, from 9.6 to 18.4\% and from 4.0 to 15.2\%. No reward
asked for these calls: the habit spread to requests that the task never touched. Behind the gate, tool use on Dolly
equals the base agent's at every stage of every run, and on KK until KK itself is learned.

\textbf{The new skill costs little:} KK accuracy behind the gate is 37.3\% against 38.4\% for Seq-GRPO at 1.5B (mean
of three seeds, base 8.8\%) and 63.2\% against 66.8\% at 7B (base 9.2\%). Refusal of harmful requests that arrive
mid-session stays at the base agent's 98.8\% behind the gate in every run; under Seq-GRPO it is 97.2 to 98.8\% at
1.5B and 93.2\% at 7B.

\subsection{A Public API-Call Benchmark}
\label{sec:res-api}

The two skills above are easy to tell apart. Here the agent learns two skills that share one system prompt, one
toolbox and one dialogue format, on data written by others.

\textbf{Setup:} The Schema-Guided Dialogue dataset (SGD) \citep{rastogi2020sgd} contains task-oriented dialogues in
which the system's calls to a service are annotated. Our agent sees the same toolbox of 13 functions from five
services in every episode. An episode is a real dialogue up to a turn at which the system called a service, and the
agent's reply counts as correct only if it calls the annotated function with the annotated arguments. Qwen2.5-1.5B
learns flight calls and then restaurant calls by GRPO, with the settings of \cref{sec:setup-main}. The seven
functions of the other three services (events, buses and hotels) are never trained. Their dialogues are a protected
component of both gates, next to Dolly, AdvBench and task-like prompts and, when restaurants are learned, the stored
flight dialogues. After each task we play 250 greedy episodes per set: flights, restaurants, the other services, and
held-out Dolly and AdvBench requests sent to the same agent. \Cref{tab:api} reports seed 0, and \cref{app:api} gives
the details.

\begin{table}[t]
\centering
\scriptsize
\setlength{\tabcolsep}{4pt}
\caption{API calls of the SGD benchmark after both tasks (Qwen2.5-1.5B, seed 0; 250 greedy episodes per set, in \%;
all values machine-scored). Correct: the annotated function with the annotated arguments; the other services are
never trained. Changed: episodes whose reply differs; chat requests are from Dolly, harmful requests from AdvBench.
Unfired changed: episodes that changed without a firing / episodes without a firing, over both tasks.
Neither gate fired on an episode of the other services, of Dolly or of AdvBench (95\% bound 1.2\%).}
\label{tab:api}
\begin{tabular}{l rrr}
\toprule
 & Base agent & Seq-GRPO & \methodp{} \\
\midrule
Correct: flights (task 1) & 12.4 & 76.0 & 52.8 \\
Correct: restaurants (task 2) & 20.8 & 82.0 & 66.4 \\
Correct: other services & 16.8 & 50.4 & 16.8 \\
\midrule
Flight episodes changed by task 2 & -- & 63.2 & 0.0 \\
Changed against the base agent: & & & \\
\quad other services & -- & 88.4 & 0.0 \\
\quad chat requests & -- & 99.2 & 0.0 \\
\quad harmful requests & -- & 92.8 & 0.0 \\
\midrule
Harmful requests refused & 100 & 8.4 & 100 \\
Unfired changed & -- & -- & 0 / 2{,}141 \\
\bottomrule
\end{tabular}
\end{table}

\textbf{No episode changed without a firing when the skills overlap:} Both gates tell the turns of their own service from the
protected prompts, although every dialogue has the same system prompt and toolbox: their classifiers reach an AUROC
of 0.996 and 0.995 on held-out episodes. No episode changed without a firing, 0 of 2{,}141 over both tasks. The
restaurant gate fired on no flight episode, so every flight episode after the second task is the episode from before
it, token for token, while Seq-GRPO changed 63.2\% of them (\cref{tab:api}). The flight gate also fired on no
restaurant episode, although no restaurant dialogue had been fitted or calibrated on at that point. Under sampled decoding (temperature
1 and 0.3, and top-$p$ 0.9), with shared randomness, again no episode changed without a firing, 0 of 6{,}423, and
the gates fired as under greedy decoding.

\textbf{Everything protected stays as it was:} Behind the gates, every episode of the other services, every chat
episode and every reply to a harmful request is identical to the base agent's after both tasks. Without gates,
88.4\%, 99.2\% and 92.8\% of them changed.

\textbf{Ungated training turns refusals into function calls:} The base agent refuses all 250 harmful requests. After
the flight task, the ungated agent refuses 3.2\% of them and answers 96.4\% with a function call: 81.2\% with a
function of the toolbox and 15.2\% with one that does not exist, such as
\texttt{Generate\allowbreak Phishing\allowbreak Email}. After both tasks it refuses 8.4\%. It also answers 99.2\% of
chat requests with a function call after the flight task, against 20.4\% for the base agent
(\cref{tab:api_replies}). No training episode contained such a request; the habit of calling a function spread to
them, as the calculator habit did in \cref{sec:res-agentloop}. The gated agent's replies to both sets are the base
agent's.

\textbf{The gain follows the coverage:} The flight gate covers 66.8\% of flight episodes and the restaurant gate
76.8\% of restaurant episodes. Both shares depend only on the prompts, so they are known before training. The
ungated agent gains 63.6 and 61.2 points on the two tasks, and the gated agent gains 40.4 and 45.6 points, close to
the covered shares of the ungated gains, 42.5 and 47.0 points. As in \cref{sec:res-cost}, the coverage of
the chosen gate, and with it the predicted gain, is known before training.

\section{Related Work}
\label{sec:related}

\textbf{Evolving agents:} Self-evolving agents change their models, memory, tools and architecture over time
\citep{gao2025survey}, and an agent operating system manages such resources for the agents it runs
\citep{mei2024aios}. For the model, \method{} adds what an evaluation after training cannot give: a bound on
regressions before training, a trace of where an update can have acted, and exact rollback.

\textbf{Forgetting in continual post-training:} Continual-learning methods regularise parameters
\citep{kirkpatrick2017ewc}, distil the old model \citep{li2018lwf}, replay data \citep{rolnick2019replay} or isolate
parameters, often as orthogonal or null-space LoRA \citep{wang2023olora,liang2024inflora}. Under RL
fine-tuning, forgetting has been tied to the KL divergence from the old policy \citep{shenfeld2026razor} and to on-policy
data \citep{chen2025retaining}, and benign fine-tuning erodes refusals \citep{qi2024finetuning}. Safety-preserving
methods change the prompt template used in fine-tuning or choose its data \citep{lyu2024keeping,pham2025preventing}.
These methods reduce the change measured after training; they bound no change on distributions that were not evaluated.

\textbf{Certificates for model updates:} Evaluations and auditing tests detect a behavioural shift after it has happened
\citep{richter2025auditing}, and \citet{friedman2026bounds} bound the forgetting of trained weights. In contrast, our
certificate is issued before training, from unlabeled prompts, for the model as deployed. Its proof is a coupling
argument \citep{lindvall1992coupling}, and its calibration follows the Neyman--Pearson umbrella algorithm and
training-conditional conformal prediction \citep{tong2018np,vovk2012conditional}.

\textbf{Routing and gating:} Expert gates \citep{aljundi2017expertgate}, gated LoRA \citep{liang2025gainlora}, model
editors \citep{mitchell2022serac} and LLM routers \citep{ong2024routellm} send inputs to specialised computation, with
gates learned and judged on measured behaviour. Closer to our setting, STABLE accepts, rescales or rejects a trained
LoRA edit by a stability budget measured on earlier edits \citep{hoy2025stable}, and TSRL adds supervision only on
the states where an agent's policy fails \citep{fu2026tsrl}; neither gives a bound before training. We calibrate such
a gate with a per-component Neyman--Pearson
threshold, which bounds the drift of any model behind it before training, and show that no update with the same budget
can beat the best such test.

\section{Limitations} The certificate covers only the prompt distributions the gate samples, so a new format, such
as an agent's prompts, must be added as a component. It bounds how often behaviour changes, not what an update does
where the gate fires: it bounds change, not safety. It does not cover an adversary who searches for prompts that fire
the gate, and switching a gate off rolls back the model, not actions an agent has already taken.

\section{Conclusion}
\label{sec:conclusion}

In this work, we bound the forgetting of continual fine-tuning before training. \method{} places each task's update
behind a calibrated gate: where no gate fires, the model is exactly the old one, so the average of every bounded
behaviour score on a prompt distribution changes by at most the firing rate there, for any weights that training
produces and over a whole task stream. Forgetting
thus becomes a false alarm whose rate is fixed in advance, and the price, the coverage of the chosen gate, is also measured before
training. On RL task streams, gated training kept the earlier tasks and left almost all general-chat responses
unchanged, and models behind the gate kept about the share of their gain that the coverage predicted. When agents
were trained inside their tool loop, on a calculator task and on a public API-call benchmark, the episodes of the
earlier skill stayed identical behind the gate, while ungated training changed most of them. For an agent that keeps
learning, each policy update becomes a release with a regression bound fixed in advance, a firing log that shows
where it acted, and exact rollback.

\bibliography{references}

@article{shao2024deepseekmath,
  title={{DeepSeekMath}: Pushing the Limits of Mathematical Reasoning in Open Language Models},
  author={Shao, Zhihong and Wang, Peiyi and Zhu, Qihao and others},
  journal={arXiv preprint arXiv:2402.03300},
  year={2024}
}

@inproceedings{rastogi2020sgd,
  title     = {Towards Scalable Multi-Domain Conversational Agents: The Schema-Guided Dialogue Dataset},
  author    = {Rastogi, Abhinav and Zang, Xiaoxue and Sunkara, Srinivas and Gupta, Raghav and Khaitan, Pranav},
  booktitle = {Proceedings of the AAAI Conference on Artificial Intelligence},
  volume    = {34},
  number    = {05},
  pages     = {8689--8696},
  year      = {2020},
  doi       = {10.1609/aaai.v34i05.6394}
}

@article{hoy2025stable,
  title   = {{STABLE}: Gated Continual Learning for Large Language Models},
  author  = {Hoy, William and Celik, Nurcin},
  journal = {arXiv preprint arXiv:2510.16089},
  year    = {2025}
}

@article{fu2026tsrl,
  title   = {{TSRL}: Trigger-based Supervised Reinforcement Learning for Continual Agent Training},
  author  = {Fu, Yuqian and Jiang, Kaiwen and Song, Zijun and Wang, Jiawei and Zhu, Yuanheng and Zhao, Dongbin},
  journal = {arXiv preprint arXiv:2610.17726},
  year    = {2026}
}

@article{qwen2025qwen25,
  title={{Qwen2.5} Technical Report},
  author={{Qwen Team}},
  journal={arXiv preprint arXiv:2412.15115},
  year={2024}
}

@inproceedings{hu2022lora,
  title={{LoRA}: Low-Rank Adaptation of Large Language Models},
  author={Hu, Edward J. and Shen, Yelong and Wallis, Phillip and others},
  booktitle={ICLR},
  year={2022}
}

@article{biderman2024lora,
  title={{LoRA} Learns Less and Forgets Less},
  author={Biderman, Dan and Portes, Jacob and Gonzalez Ortiz, Jose Javier and others},
  journal={Transactions on Machine Learning Research},
  year={2024}
}

@article{grattafiori2024llama3,
  title={The {Llama} 3 Herd of Models},
  author={Grattafiori, Aaron and Dubey, Abhimanyu and others},
  journal={arXiv preprint arXiv:2407.21783},
  year={2024}
}

@article{bi2024deepseekllm,
  title={{DeepSeek LLM}: Scaling Open-Source Language Models with Longtermism},
  author={Bi, Xiao and Chen, Deli and others},
  journal={arXiv preprint arXiv:2401.02954},
  year={2024}
}

@inproceedings{shenfeld2026razor,
  title={{RL}'s Razor: Why Online Reinforcement Learning Forgets Less},
  author={Shenfeld, Idan and Pari, Jyothish and Agrawal, Pulkit},
  booktitle={ICLR},
  year={2026}
}

@inproceedings{chen2025retaining,
  title={Retaining by Doing: The Role of On-Policy Data in Mitigating Forgetting},
  author={Chen, Howard and Razin, Noam and Narasimhan, Karthik and Chen, Danqi},
  booktitle={ICML},

  year={2026}
}

@article{luo2026cpo,
  title={{RL} Forgets! {T}owards Continual Policy Optimization},
  author={Luo, Mao-Lin and Wang, Zhe-Xu and Zhou, Zi-Hao and others},
  journal={arXiv preprint arXiv:2607.04364},
  year={2026}
}

@article{kirkpatrick2017ewc,
  title={Overcoming Catastrophic Forgetting in Neural Networks},
  author={Kirkpatrick, James and Pascanu, Razvan and Rabinowitz, Neil and others},
  journal={Proceedings of the National Academy of Sciences},
  volume={114},
  number={13},
  pages={3521--3526},
  year={2017}
}

@inproceedings{rolnick2019replay,
  title={Experience Replay for Continual Learning},
  author={Rolnick, David and Ahuja, Arun and Schwarz, Jonathan and others},
  booktitle={NeurIPS},
  year={2019}
}

@inproceedings{wang2023olora,
  title={Orthogonal Subspace Learning for Language Model Continual Learning},
  author={Wang, Xiao and Chen, Tianze and Ge, Qiming and others},
  booktitle={Findings of EMNLP 2023},
  year={2023}
}

@inproceedings{liang2024inflora,
  title={{InfLoRA}: Interference-Free Low-Rank Adaptation for Continual Learning},
  author={Liang, Yan-Shuo and Li, Wu-Jun},
  booktitle={CVPR},
  year={2024}
}

@inproceedings{liang2025gainlora,
  title={Gated Integration of Low-Rank Adaptation for Continual Learning of Large Language Models},
  author={Liang, Yan-Shuo and Chen, Jia-Rui and Li, Wu-Jun},
  booktitle={NeurIPS},
  year={2025}
}

@inproceedings{qi2024finetuning,
  title={Fine-tuning Aligned Language Models Compromises Safety, Even When Users Do Not Intend To!},
  author={Qi, Xiangyu and Zeng, Yi and Xie, Tinghao and others},
  booktitle={ICLR},
  year={2024}
}

@inproceedings{vovk2012conditional,
  title={Conditional Validity of Inductive Conformal Predictors},
  author={Vovk, Vladimir},
  booktitle={ACML},
  year={2012}
}

@article{clopper1934,
  title={The Use of Confidence or Fiducial Limits Illustrated in the Case of the Binomial},
  author={Clopper, C. J. and Pearson, E. S.},
  journal={Biometrika},
  volume={26},
  number={4},
  pages={404--413},
  year={1934}
}

@article{neyman1933,
  title={On the Problem of the Most Efficient Tests of Statistical Hypotheses},
  author={Neyman, Jerzy and Pearson, Egon S.},
  journal={Philosophical Transactions of the Royal Society A},
  volume={231},
  pages={289--337},
  year={1933}
}

@book{ville1939,
  title={{\'E}tude critique de la notion de collectif},
  author={Ville, Jean},
  publisher={Gauthier-Villars},
  year={1939}
}

@article{ramdas2023gametheoretic,
  title={Game-Theoretic Statistics and Safe Anytime-Valid Inference},
  author={Ramdas, Aaditya and Gr{\"u}nwald, Peter and Vovk, Vladimir and Shafer, Glenn},
  journal={Statistical Science},
  volume={38},
  number={4},
  pages={576--601},
  year={2023}
}

@book{lindvall1992coupling,
  title={Lectures on the Coupling Method},
  author={Lindvall, Torgny},
  publisher={Wiley},
  year={1992}
}

@article{cobbe2021gsm8k,
  title={Training Verifiers to Solve Math Word Problems},
  author={Cobbe, Karl and Kosaraju, Vineet and Bavarian, Mohammad and others},
  journal={arXiv preprint arXiv:2110.14168},
  year={2021}
}

@misc{pan2025tinyzero,
  title={{TinyZero}},
  author={Pan, Jiayi and Zhang, Junjie and Wang, Xingyao and others},
  howpublished={\url{https://github.com/Jiayi-Pan/TinyZero}},
  year={2025}
}

@inproceedings{xie2024kk,
  title={On Memorization of Large Language Models in Logical Reasoning},
  author={Xie, Chulin and Huang, Yangsibo and Zhang, Chiyuan and others},
  booktitle={Proceedings of IJCNLP-AACL},
  year={2025}
}

@inproceedings{hendrycks2021mmlu,
  title={Measuring Massive Multitask Language Understanding},
  author={Hendrycks, Dan and Burns, Collin and Basart, Steven and others},
  booktitle={ICLR},
  year={2021}
}

@article{zou2023gcg,
  title={Universal and Transferable Adversarial Attacks on Aligned Language Models},
  author={Zou, Andy and Wang, Zifan and Carlini, Nicholas and others},
  journal={arXiv preprint arXiv:2307.15043},
  year={2023}
}

@inproceedings{rottger2024xstest,
  title={{XSTest}: A Test Suite for Identifying Exaggerated Safety Behaviours in Large Language Models},
  author={R{\"o}ttger, Paul and Kirk, Hannah Rose and Vidgen, Bertie and others},
  booktitle={NAACL},
  year={2024}
}

@article{kwiatkowski2019nq,
  title={Natural Questions: A Benchmark for Question Answering Research},
  author={Kwiatkowski, Tom and Palomaki, Jennimaria and Redfield, Olivia and others},
  journal={Transactions of the Association for Computational Linguistics},
  volume={7},
  pages={452--466},
  year={2019}
}

@inproceedings{wortsman2022robust,
  title={Robust Fine-Tuning of Zero-Shot Models},
  author={Wortsman, Mitchell and Ilharco, Gabriel and Kim, Jong Wook and others},
  booktitle={CVPR},
year={2022}}

@article{li2018lwf,
  title={Learning without Forgetting},
  author={Li, Zhizhong and Hoiem, Derek},
  journal={IEEE Transactions on Pattern Analysis and Machine Intelligence},
  volume={40}, number={12}, pages={2935--2947}, year={2018}}

@inproceedings{hendrycks2021math,
  title={Measuring Mathematical Problem Solving With the {MATH} Dataset},
  author={Hendrycks, Dan and Burns, Collin and Kadavath, Saurav and others},
  booktitle={NeurIPS Datasets and Benchmarks}, year={2021}}

@article{olmo2024olmo2,
  title={2 {OLMo} 2 Furious},
  author={{Team OLMo}},
  journal={arXiv preprint arXiv:2501.00656}, year={2024}}

@article{zhou2023ifeval,
  title={Instruction-Following Evaluation for Large Language Models},
  author={Zhou, Jeffrey and Lu, Tianjian and Mishra, Swaroop and others},
  journal={arXiv preprint arXiv:2311.07911}, year={2023}}

@inproceedings{chaudhry2019agem,
  title={Efficient Lifelong Learning with {A-GEM}},
  author={Chaudhry, Arslan and Ranzato, Marc'Aurelio and Rohrbach, Marcus and Elhoseiny, Mohamed},
  booktitle={ICLR},
  year={2019}
}

@inproceedings{mitchell2022serac,
  title={Memory-Based Model Editing at Scale},
  author={Mitchell, Eric and Lin, Charles and Bosselut, Antoine and others},
  booktitle={ICML},
  year={2022}
}

@inproceedings{aljundi2017expertgate,
  title={Expert Gate: Lifelong Learning with a Network of Experts},
  author={Aljundi, Rahaf and Chakravarty, Punarjay and Tuytelaars, Tinne},
  booktitle={CVPR},
  year={2017}
}

@article{tong2018np,
  title={{Neyman-Pearson} Classification Algorithms and {NP} Receiver Operating Characteristics},
  author={Tong, Xin and Feng, Yang and Li, Jingyi Jessica},
  journal={Science Advances},
  volume={4},
  number={2},
  year={2018}
}

@inproceedings{ong2024routellm,
  title={{RouteLLM}: Learning to Route {LLMs} from Preference Data},
  author={Ong, Isaac and others},
  booktitle={ICLR},
  year={2025}
}

@article{jiang2023mistral,
  title={Mistral 7{B}},
  author={Jiang, Albert Q. and Sablayrolles, Alexandre and Mensch, Arthur and Bamford, Chris and Chaplot, Devendra Singh and de las Casas, Diego and Bressand, Florian and Lengyel, Gianna and Lample, Guillaume and Saulnier, Lucile and others},
  journal={arXiv preprint arXiv:2310.06825},
  year={2023}
}

@inproceedings{friedman2026bounds,
  title={Data-dependent and Oracle Bounds on Forgetting in Continual Learning},
  author={Friedman, Lior and Meir, Ron},
  booktitle={CoLLAs},

  year={2026}
}

@article{pham2025preventing,
  title={Preventing Catastrophic Forgetting: Behavior-Aware Sampling for Safer Language Model Fine-Tuning},
  author={Pham, Anh and Thalanki, Mihir and Sun, Michael and others},
  journal={arXiv preprint arXiv:2510.21885},
  year={2025}
}

@inproceedings{rajpurkar2018squad2,
  title={Know What You Don't Know: Unanswerable Questions for {SQuAD}},
  author={Rajpurkar, Pranav and Jia, Robin and Liang, Percy},
  booktitle={ACL (Short Papers)},
  year={2018}
}

@inproceedings{zhao2024wildchat,
  title={{WildChat}: 1{M} {ChatGPT} Interaction Logs in the Wild},
  author={Zhao, Wenting and Ren, Xiang and Hessel, Jack and others},
  booktitle={ICLR},
  year={2024}
}

@inproceedings{mazeika2024harmbench,
  title={{HarmBench}: A Standardized Evaluation Framework for Automated Red Teaming and Robust Refusal},
  author={Mazeika, Mantas and Phan, Long and Yin, Xuwang and others},
  booktitle={ICML},
  year={2024}
}

@inproceedings{souly2024strongreject,
  title={A {StrongREJECT} for Empty Jailbreaks},
  author={Souly, Alexandra and Lu, Qingyuan and Bowen, Dillon and others},
  booktitle={NeurIPS Datasets and Benchmarks},
  year={2024}
}

@inproceedings{lyu2024keeping,
  title={Keeping {LLMs} Aligned After Fine-tuning: The Crucial Role of Prompt Templates},
  author={Lyu, Kaifeng and Zhao, Haoyu and Gu, Xinran and others},
  booktitle={NeurIPS},
  year={2024}
}

@inproceedings{richter2025auditing,
  title={An Auditing Test to Detect Behavioral Shift in Language Models},
  author={Richter, Leo and He, Xuanli and Minervini, Pasquale and Kusner, Matt J.},
  booktitle={ICLR},
  year={2025}
}

@inproceedings{deng2024multilingual,
  title={Multilingual Jailbreak Challenges in Large Language Models},
  author={Deng, Yue and Zhang, Wenxuan and Pan, Sinno Jialin and Bing, Lidong},
  booktitle={ICLR},
  year={2024}
}

@inproceedings{kopf2023openassistant,
  title={{OpenAssistant} Conversations -- Democratizing Large Language Model Alignment},
  author={K{\"o}pf, Andreas and Kilcher, Yannic and von R{\"u}tte, Dimitri and others},
  booktitle={NeurIPS Datasets and Benchmarks},
  year={2023}
}

@inproceedings{wang2024donotanswer,
  title={Do-Not-Answer: Evaluating Safeguards in {LLMs}},
  author={Wang, Yuxia and Li, Haonan and Han, Xudong and others},
  booktitle={Findings of EACL},
  year={2024}
}

@inproceedings{aakanksha2024prism,
  title={The Multilingual Alignment Prism: Aligning Global and Local Preferences to Reduce Harm},
  author={Aakanksha and Ahmadian, Arash and Ermis, Beyza and others},
  booktitle={EMNLP},
  year={2024}
}

@inproceedings{joshi2017triviaqa,
  title     = {{TriviaQA}: A Large Scale Distantly Supervised Challenge Dataset for Reading Comprehension},
  author    = {Joshi, Mandar and Choi, Eunsol and Weld, Daniel S. and Zettlemoyer, Luke},
  booktitle = {Proceedings of the 55th Annual Meeting of the Association for Computational Linguistics (Volume 1: Long Papers)},
  pages     = {1601--1611},
  year      = {2017}
}

@article{mei2024aios,
  title={{AIOS}: {LLM} Agent Operating System},
  author={Mei, Kai and Zhu, Xi and Xu, Wujiang and Hua, Wenyue and Jin, Mingyu and Li, Zelong and Xu, Shuyuan and Ye, Ruosong and Ge, Yingqiang and Zhang, Yongfeng},
  journal={arXiv preprint arXiv:2403.16971},
  year={2024}
}

@article{gao2025survey,
  title={A Survey of Self-Evolving Agents: What, When, How, and Where to Evolve on the Path to Artificial Super Intelligence},
  author={Gao, Huan-ang and Geng, Jiayi and Hua, Wenyue and Hu, Mengkang and Juan, Xinzhe and Liu, Hongzhang and others},
  journal={arXiv preprint arXiv:2507.21046},
  year={2025}
}

\appendix
\makeatletter
\renewenvironment{table}[1][]{\@dblfloat{table}[tp]}{\end@dblfloat}
\renewenvironment{figure}[1][]{\@dblfloat{figure}[tp]}{\end@dblfloat}
\makeatother

\section{Theory and Proofs}
\label{app:proofs}\label{app:theory}

\Cref{tab:status} lists the status of each claim. We first state three results that the main text uses for the
attained level and for task streams, and then prove every statement.

\begin{table}[t]
\centering
\footnotesize
\setlength{\tabcolsep}{2.5pt}
\caption{Status of each claim: proved, estimated before training, or observed.}
\label{tab:status}
\begin{tabular}{l l l}
\toprule
Statement & Status & Conditions and evidence \\
\midrule
Drift at most the firing rate (\cref{thm:coupling}) & proved & frozen trunk, causal attention; checked exactly \\
Any model behind a prompt-time gate (\cref{cor:switch}) & proved & the deployed pair, not its forgetting \\
Firing at most $\varepsilon$ per component (\cref{thm:calib}) & proved & i.i.d.\ prompts, score fixed before them; components only \\
The Neyman--Pearson limit (\cref{thm:limit}) & proved & any policy; attained if the ratio reads the trunk \\
$K$ stages at $K\varepsilon$ (\cref{cor:compose}) & proved & confidence $1-K\delta$; anytime: fresh prompts, one rule \\
Coverage $\beta$ & estimated & unlabeled RL prompts; a lower bound on $\beta^\ast$ \\
Gain kept $\approx\beta\bar G$ (\cref{prop:switchgain}) & identity; fit observed & exact if missed prompts gain alike; error 0.5 points \\
Audit, monitor, rollback (\cref{prop:deploy}) & proved & audit on stored responses; monitor and repair on live traffic \\
Whole agent episodes (\cref{cor:episode}) & proved & tool replies a fixed kernel of the episode; no state across episodes \\
Firing elsewhere; gains and changes & observed & held-out bounds; three seeds unless marked \\
\bottomrule
\end{tabular}
\end{table}

\textbf{The level attained:} \Cref{thm:calib} guarantees the level with probability $1-\delta$; its exact law gives the
price of that guarantee.

\begin{proposition}[Level attained by calibration]
\label{prop:size}
In \cref{thm:calib}, suppose $k_c\le n_c$ and the score $S$ of $\pold$'s trajectories on $D_c$ has a continuous
distribution. Then $\varphi_{g_{\hat\kappa}}(D_c)\le P^{D_c}_{\pold}(S>\hat\kappa_c)$, and the right side has the
Beta$(n_c+1-k_c,k_c)$ distribution, with mean $(n_c+1-k_c)/(n_c+1)$; it exceeds $\varepsilon$ with probability
$\Pr[\mathrm{Bin}(n_c,1-\varepsilon)\ge k_c]\le\delta/C$.
\end{proposition}

For our 700 Dolly calibration prompts, three components, $\varepsilon=0.03$ and $\delta=0.05$, $k_c=689$: the bound on
the attained level has mean 1.7\% and 5--95\% quantiles of 1.0 and 2.6\%.

\textbf{Certificates over a stream:} Each stage is calibrated against the model that precedes it, not against the base
model; the certificates still add up.

\begin{corollary}[Sequential calibration composes]
\label{cor:compose}
If $P^D_{\pi_{i-1}}(\Fire_{g_i})\le\varepsilon_i$ for $i=j+1,\dots,k$, then
$\TV(P^D_{\pi_k},P^D_{\pi_j})\le\sum_{i=j+1}^k\varepsilon_i$.
\end{corollary}

Calibrating $K$ stages with failure probability $\delta$ each gives the composed target with probability at least
$1-K\delta$. For \methoda{}, \cref{thm:calib} covers stage $k$ only if its calibration prompts are not used by earlier
stages, because $\pi_{k-1}$ depends on the earlier thresholds; prompt-time gates may reuse them, since their firing
depends on the prompt alone. With prompt-time gates, the stream certificate no longer involves the models at all.

\begin{proposition}[Prompt-time streams]
\label{prop:promptstream}
Let $g_1,\dots,g_K$ be prompt-time gates on the frozen trunk with decisions $G_i$. (a) For gated updates, for every
$j<k$ and every $D$, $\TV(P^D_{\pi_k},P^D_{\pi_j})\le D\big(\bigcup_{i=j+1}^{k}\{G_i=1\}\big)$. (b) For any policy
$\pi'$, the switch that answers with $\pi'$ when any of $G_1,\dots,G_K$ fires has drift at most
$D\big(\bigcup_{i=1}^{K}\{G_i=1\}\big)$ from $\pi_0$.
\end{proposition}

Both right sides are probabilities of an event in the prompt alone, so we estimate them from unlabeled prompts of $D$
by \cref{prop:cp}. This allows one budget for a whole stream: the stages' thresholds are set so that the union fires on
at most a share $\varepsilon$ of each component's calibration prompts, and the union is then certified by
\cref{prop:cp} on fresh prompts.

\textbf{Proofs.} Throughout, positions are $1,\dots,|s|$, and a block's output at position $t$ is a function of the
previous block's outputs at positions $\le t$ (causal attention followed by position-wise maps). The update acts
position-wise. The script \texttt{scripts/verify\_theory.py} also checks \cref{thm:coupling,prop:tight,thm:stream,cor:compose}
by exact enumeration on small random autoregressive models, \cref{thm:limit} by linear programs, and
\cref{prop:cp,thm:calib} by Monte Carlo.

\begin{lemma}[Locality]
\label{lem:locality}
Let $t<T_g(s)$. Then for every block $\ell$ and every position $\tau\le t$, the hidden state of $\pi_{\Delta,g}$ at
$(\ell,\tau)$ on $s$ equals that of $\pold$. Consequently $\pi_{\Delta,g}(\cdot\mid s_{\le t})=\pold(\cdot\mid
s_{\le t})$.
\end{lemma}

\begin{proof}
Blocks $\ell<\lg$ have the same weights in both models, so their states agree at all positions. For $\ell\ge\lg$ the
proof is by induction on $\ell$. The state at $(\ell,\tau)$ is computed from the states at $(\ell-1,\sigma)$,
$\sigma\le\tau\le t$, which agree by the induction hypothesis, and the update is inactive at every position
$\sigma\le\tau$ because $m_\sigma(s)=0$ for $\sigma<T_g(s)$. The next-token distribution at $t$ is a function of the
last block's state at $t$.
\end{proof}

\begin{proof}[Proof of \cref{thm:coupling}]
Draw $x\sim D$ and $U_1,U_2,\dots$ i.i.d.\ uniform, independent of $x$. Generate $s^{\mathrm{o}}$ with $\pold$ and
$s^{\mathrm{n}}$ with $\pi_{\Delta,g}$, both from $x$, using $U_i$ for the $i$-th response token. Each has law $P^D$ of
its model. Let $\Fire=\Fire_g(s^{\mathrm{o}})$.

\emph{On $\Fire^c$ the runs coincide:} Suppose the first $i-1$ response tokens agree and $s^{\mathrm{o}}$ has an
$i$-th response token. The common prefix $p$ has length $|p|\le|s^{\mathrm{o}}|-1$, and on $\Fire^c$ the gate does not
fire at any position $\le|s^{\mathrm{o}}|-1$, so $T_g(s^{\mathrm{o}})>|p|$. Because $u_\tau$ depends only on the
prefix up to $\tau$, the same holds along $s^{\mathrm{n}}$. By \cref{lem:locality} both models have the same next-token
distribution after $p$, and the shared $U_i$ gives the same token. The stopping rule depends only on the tokens, so both
runs stop together.

\emph{The firing event is the same in both runs:} On $\Fire$, let $t^*=T_g(s^{\mathrm{o}})\le|s^{\mathrm{o}}|-1$. The
runs agree on all tokens up to position $t^*$, so $u_{t^*}(s^{\mathrm{n}})=u_{t^*}(s^{\mathrm{o}})$ and the gate fires
at $t^*$ on $s^{\mathrm{n}}$; since $s^{\mathrm{o}}$ does not stop at $t^*$ and the stopping rule reads only these
tokens, neither does $s^{\mathrm{n}}$. Conversely, a firing on $s^{\mathrm{n}}$ before any firing on $s^{\mathrm{o}}$ is
impossible because the runs agree until the first firing. Hence $\Fire_g(s^{\mathrm{n}})=\Fire_g(s^{\mathrm{o}})$
pathwise.

\emph{Bound:} $\{s^{\mathrm{n}}\ne s^{\mathrm{o}}\}\subseteq\Fire$, so by the coupling inequality
$\TV(P^D_{\pi_{\Delta,g}},P^D_{\pold})\le\Pr[s^{\mathrm{n}}\ne s^{\mathrm{o}}]\le\varphi_g(D)$, and for $f\in[0,1]$,
$|\E f(s^{\mathrm{n}})-\E f(s^{\mathrm{o}})|\le\Pr[s^{\mathrm{n}}\ne s^{\mathrm{o}}]$. A prompt-time gate is the special
case in which $g$ is evaluated only at position $|x|$. The argument holds unchanged, conditionally, for a gate that also
reads a random variable independent of $x$ and of the $U_i$.
\end{proof}

\begin{proof}[Proof of \cref{prop:tight}]
Use the coupling above. On $\Fire$, the first generated token at a position where the update is active is the token
produced at position $t^\dagger=\max(T_g,|x|)\le|s|-1$. Let $A=\Fire\cap\{\text{the token generated from position
}t^\dagger\text{ is }a^\dagger\}$. Under $\pi_{\Delta,g}$, $A$ has probability $\varphi_g(D)$. Under $\pold$, $\Fire$
is decided by $s_{\le t^\dagger}$, the runs agree up to $t^\dagger$, and $\pold$ gives $a^\dagger$ probability at most
$\eta$, so $\Pr_{\pold}(A)\le\eta\,\varphi_g(D)$. Hence $\TV\ge(1-\eta)\varphi_g(D)$.
\end{proof}

\begin{proof}[Proof of \cref{cor:switch}]
The trunk is frozen and attention is causal, so $u_{|x|}$ and hence $G(x)$ depend on $x$ alone. Both laws have prompt
marginal $D$, so by \cref{eq:tv}
$\TV(P^D_{\pi_{\pi',G}},P^D_{\pold})=\E_D\big[G(x)\,\TV(\pi'(\cdot\mid x),\pold(\cdot\mid x))\big]\le D(G=1)$. Every
response has at least one token, so $\Fire_g=\{G(x)=1\}$ and $D(G=1)=\varphi_g(D)$. With shared decoding randomness the
switch runs $\pold$ itself on prompts with $G(x)=0$, so the two responses coincide there.
\end{proof}

\begin{proof}[Proof of \cref{prop:cp}]
Firing is a Bernoulli$(\varphi_g(D))$ event per i.i.d.\ trajectory of $\pold$, so $X\sim\mathrm{Bin}(n,\varphi_g(D))$
and the Clopper--Pearson upper limit has coverage at least $1-\delta$.
\end{proof}

\begin{proof}[Proof of \cref{thm:calib}]
Fix $c$ and let $\bar F(\kappa)=P^{D_c}_{\pold}(S>\kappa)$, which bounds $\varphi_{g_\kappa}(D_c)$ and is
non-increasing and right-continuous. Let $q=\inf\{\kappa:\bar F(\kappa)\le\varepsilon\}$; then
$\bar F(q)\le\varepsilon$ and, since $\bar F(\kappa)>\varepsilon$ for $\kappa<q$, $P^{D_c}_{\pold}(S\ge
q)\ge\varepsilon$ (if $q=-\infty$ there is nothing to prove). For $\hat\kappa_c=S^c_{(k_c)}$,
$\bar F(\hat\kappa_c)>\varepsilon$ implies $\hat\kappa_c<q$, so at least $k_c$ calibration scores are below $q$. Each is
below $q$ independently with probability $P^{D_c}_{\pold}(S<q)\le1-\varepsilon$, because the calibration trajectories
are i.i.d.\ and independent of $w$ and $\pold$, so by the choice of $k_c$ the event has probability at most
$\Pr[\mathrm{Bin}(n_c,1-\varepsilon)\ge k_c]\le\delta/C$. A union bound over components and monotonicity
($\hat\kappa\ge\hat\kappa_c$) finish the proof. No continuity is needed: the gate fires only on scores strictly above
the threshold.
\end{proof}

\begin{proof}[Proof of \cref{prop:size}]
Firing requires $S>\hat\kappa\ge\hat\kappa_c$, which gives the inequality. Let $F$ be the continuous distribution
function of $S$ on $D_c$. Then $U_i=F(S^c_i)$ are i.i.d.\ uniform, $F(\hat\kappa_c)=U_{(k_c)}$ has the
Beta$(k_c,n_c+1-k_c)$ distribution, and $P^{D_c}_{\pold}(S>\hat\kappa_c)=1-U_{(k_c)}$ has the Beta$(n_c+1-k_c,k_c)$
distribution \citep{vovk2012conditional}. It exceeds $\varepsilon$ exactly when at least $k_c$ of the $U_i$ are below
$1-\varepsilon$, an event of probability $\Pr[\mathrm{Bin}(n_c,1-\varepsilon)\ge k_c]\le\delta/C$ by the choice of
$k_c$.
\end{proof}

\begin{proof}[Proof of \cref{thm:stream}]
Let $T_{>j}=\min_{j<i\le k}T_{g_i}$, each gate on its own positions. Before $T_{>j}$ every latch $m^i$ with $i>j$ is
zero, so $\Delta_{j+1},\dots,\Delta_k$ are inactive; the latches with $i\le j$ are computed from the same trunk features
of the same prefix in both models, so $\Delta_1,\dots,\Delta_j$ act identically. \Cref{lem:locality} therefore holds
with $\pi_j$ in the role of $\pold$ and $T_{>j}$ in that of $T_g$, and the proof of \cref{thm:coupling} applies. The
event $\{T_{>j}\le|s|-1\}$ is the union of the firing events of $g_{j+1},\dots,g_k$; the second inequality is the union
bound.
\end{proof}

\begin{proof}[Proof of \cref{cor:compose}]
For $j<i\le k$ let $E_i=\bigcup_{j<m<i}\Fire_{g_m}$. The events $\Fire_{g_i}\setminus E_i$ are disjoint with union
$\bigcup_{i>j}\Fire_{g_i}$. On the complement of $E_i$ no gate $g_{j+1},\dots,g_{i-1}$ fires, so by the coupling in
the proof of \cref{thm:stream} the coupled trajectories of $\pi_j$ and $\pi_{i-1}$ coincide and
$P^D_{\pi_j}(\Fire_{g_i}\setminus E_i)=P^D_{\pi_{i-1}}(\Fire_{g_i}\setminus E_i)\le\varepsilon_i$. \Cref{thm:stream}
bounds the drift by $P^D_{\pi_j}(\bigcup_{i>j}\Fire_{g_i})\le\sum_{i>j}\varepsilon_i$.
\end{proof}

\begin{proof}[Proof of \cref{prop:promptstream}]
(a) By \cref{thm:stream}, $\TV(P^D_{\pi_k},P^D_{\pi_j})\le P^D_{\pi_j}(\bigcup_{i=j+1}^k\Fire_{g_i})$. For a
prompt-time gate $\Fire_{g_i}=\{G_i(x)=1\}$ (proof of \cref{cor:switch}), an event in the prompt alone, whose
probability under $P^D_{\pi_j}$ is its probability under $D$ whatever $\pi_j$ is. (b) $\max_iG_i$ is the decision of a
prompt-time gate on the same trunk features; apply \cref{cor:switch}.
\end{proof}

\begin{proof}[Proof of \cref{thm:limit}]
By \cref{eq:tv}, the drift on $D$ is $\E_D\rho$. The map $x\mapsto\rho(x)\in[0,1]$ is a randomised test with size
$\E_{\Dref}\rho\le\varepsilon$, so its power $\E_{\Dnew}\rho$ is at most $\beta^\ast(\varepsilon)$ by the
Neyman--Pearson lemma. For the KL form, let $\psi$ be any test with $\E_{\Dref}\psi\le\varepsilon$. The channel
$x\mapsto B\sim\mathrm{Bern}(\psi(x))$ maps $\Dnew,\Dref$ to $\mathrm{Bern}(a),\mathrm{Bern}(b)$ with
$a=\E_{\Dnew}\psi$, $b\le\varepsilon$; by data processing $\kl(a\|b)\le\KL(\Dnew\|\Dref)$, where $\kl$ is the KL
divergence between Bernoulli laws, and $\kl(a\|b)\ge a\log\frac1\varepsilon-\log 2$ for $\varepsilon\in(0,1)$. A reward
$r\in[r_{\min},r_{\max}]$ gains at most $(r_{\max}-r_{\min})\E_{\Dnew}\rho$. For achievability, let
$\Lambda=p_{\mathrm{new}}/p_{\mathrm{ref}}\in[0,\infty]$ be the ratio of the densities with respect to
$\mu=\Dnew+\Dref$, and let a prompt-time gate fire with probability
$\rho^*(x)=\ind[\Lambda>c^\ast]+\gamma\ind[\Lambda=c^\ast]$, $\E_{\Dref}\rho^*=\varepsilon$, using a random variable
independent of $x$ and of the decoding uniforms; it is a gate because $\Lambda$ is a function of $u_{|x|}(x)$. With an
update that puts all mass on a token of $\pold$-probability at most $\eta$,
$\pnew(\cdot\mid x)=\rho^*(x)\pi_\Delta(\cdot\mid x)+(1-\rho^*(x))\pold(\cdot\mid x)$, with $\pi_\Delta$ the adapted model
switched on at $|x|$, so $\rho(x)=\rho^*(x)\,\TV(\pi_\Delta(\cdot\mid x),\pold(\cdot\mid x))$, and \cref{prop:tight}
for the gate that always fires gives $(1-\eta)\rho^*(x)\le\rho(x)\le\rho^*(x)$, so
$\E_{\Dnew}\rho\ge(1-\eta)\beta^\ast(\varepsilon)$ with $\E_{\Dref}\rho\le\varepsilon$.
\end{proof}

\begin{proof}[Proof of \cref{prop:switchgain}]
The switch answers $x$ with $\pi'$ if $G(x)=1$ and with $\pold$ otherwise, so its expected reward on $x$ exceeds
$\pold$'s by $G(x)\Delta r(x)$. Taking expectations over $\Dnew$, its gain is $\E[G\Delta r]=\beta\bar G_1$. Since
$\bar G=\beta\bar G_1+(1-\beta)\bar G_0$, $\beta\bar G_1-\beta\bar G=\beta(1-\beta)(\bar G_1-\bar G_0)$, which is
$\mathrm{Cov}(G,\Delta r)$; it is zero exactly when missed prompts gain as much on average as the others.
\end{proof}

\begin{proof}[Proof of \cref{prop:deploy}]
(a) Outside $\Fire_g$ the coupled responses coincide (\cref{thm:coupling}), which gives the identity. The summands
$Z_i=\ind[\Fire_g](f_{\mathrm{new},i}-f_{\mathrm{old},i})$ are i.i.d.\ with mean equal to the left side, and
$\E Z_i^2\le\varphi_g(D)(b-a)^2$. (b) Each factor is non-negative, since $1-\lambda\varepsilon>0$, and under the
hypothesis $\E[1+\lambda(X_t-\varepsilon)\mid X_{<t}]\le1$, so $M_t$ is a non-negative supermartingale with $M_0=1$;
Ville's inequality \citep{ville1939} gives the claim. A mixture $\sum_j\omega_j\prod_{i\le t}(1+\lambda_j(X_i-\varepsilon))$
with weights $\omega_j\ge0$ summing to one is again such a supermartingale, so the claim also holds for a monitor that
mixes several $\lambda$. Under the alternative, $\log M_t$ is a random walk with drift
$\nu=\E\log(1+\lambda(X-\varepsilon))$ and increments at most $\log(1+\lambda)$. With
$\lambda=\frac{p-\varepsilon}{\varepsilon(1-\varepsilon)}$, $1+\lambda(1-\varepsilon)=\frac p\varepsilon$ and
$1-\lambda\varepsilon=\frac{1-p}{1-\varepsilon}$, so
$\nu=p\log\frac p\varepsilon+(1-p)\log\frac{1-p}{1-\varepsilon}=\kl(p\|\varepsilon)>0$, and Wald's identity bounds the
expected first time the walk reaches $\log\frac1\delta$ by the threshold plus one increment, divided by the drift.
(c) $\varphi_g(D)=P^D_{\pold}(\Fire_{g_\kappa})$ depends only on the old model's trajectories and is non-increasing in
$\kappa$, and \cref{thm:coupling} holds for every update. A gate that never fires leaves its latch at zero, so its
update is never applied; the other gates read trunk features, a fixed function of the token prefix, so the computation
is that of the model built without it. Recalibrating on fresh trajectories of $\pold$ (the deployed model with the new
update off) is \cref{thm:calib} with one component, and a union bound gives all certificates jointly.
\end{proof}

\begin{proof}[Proof of \cref{cor:episode}]
Run both agents from the same $x$ with the same decoding uniforms and the same uniforms for the tool kernel. Suppose no
call before call $i$ fired on the old agent's episode. Then the contexts of call $i$ coincide, so the gate decides the
same at call $i$ in both runs. If it does not fire, both runs answer with $\pold$ and the same uniforms
(\cref{thm:coupling,cor:switch}), so the responses, the tool replies and the stopping decision coincide. By induction
the episodes coincide unless some call fires on the old agent's episode, and the firing event is the same in both runs.
The coupling inequality gives the bound.
\end{proof}

\section{Implementation and Setup}
\label{app:impl}\label{app:setup}

\begin{algorithm}[t]
\caption{\method{}: stage $k$ of a continual stream}
\label{alg:tripwire}
\begin{algorithmic}
\STATE {\bfseries Input:} $\pi_{k-1}$; task $k$ prompts $P_k$ and reward $r_k$; unlabeled reference components
$D_1..D_C$ (generic chat, harmful requests, task-like prompts, stored prompts of tasks $<k$); split depth $\lg$;
$\varepsilon,\delta$; mode (anytime or prompt-time); optionally a region level $\alpha$.
\STATE Roll out $\pi_{k-1}$ greedily on fitting prompts from $P_k$ and each $D_c$.
\STATE Fit $w_k$: logistic regression on normalised trunk features, label 1 for $P_k$, 0 for $D_c$.
\STATE (two-sided) Fit the region $R_k$ of $P_k$; set $w_k\leftarrow-\infty$ outside it.
\STATE Roll out $\pi_{k-1}$ on calibration prompts of each $D_c$ and record $S=\max_t w_k(u_t)$.
\STATE Set $\kappa_k$ by \cref{thm:calib}; the certificate on every $D_c$ and the coverage of $P_k$ are now fixed.
\STATE Add LoRA $\Delta_k$ to blocks $\lg..L$ and train it with GRPO on $(P_k,r_k)$, either gated by the latch of
$\ind[w_k(u_t)>\kappa_k]$ or ungated, by any method, then placed behind the prompt-time gate (\cref{cor:switch}).
\STATE Deploy $\pi_k$: log the firings, audit the fired prompts and monitor live traffic (\cref{prop:deploy}); on an
alarm, raise $\kappa_k$ or switch the gate off.
\STATE {\bfseries Output:} $\pi_k$ and its certificates.
\end{algorithmic}
\end{algorithm}

\textbf{Classifier:} The score is $w_k(u)=a_k^\top\big((\bar u-\mu)/\sigma\big)+b_k$, where $\bar u=u/\mathrm{rms}(u)$
and $\mu,\sigma$ are per-coordinate statistics of the fitting features. Logistic regression is class-balanced, with
$\ell_2$ penalty $10^{-4}$ and L-BFGS. The anytime gate is fitted on up to 48 positions per fitting trajectory, the
prompt-time gate on the last prompt token. The fixed chat-template prefix is identical for every prompt, so the gate
skips it.

\textbf{Region of the two-sided gate:} We use as distance $d$ the Mahalanobis distance, in the classifier's
standardised feature space, to the mean of the new task's fitting prompts, with the covariance shrunk toward a multiple
of the identity (weight $0.1$), and as threshold $\hat\tau$ the empirical $(1-\alpha)$ quantile of $d$ on $n$ unlabeled
RL prompts, disjoint from the fitting prompts, with $\alpha=0.02$. The region $R=\{d\le\hat\tau\}$ is fixed before
calibration, so the score equal to $w$ on $R$ and to $-\infty$ off $R$ is a fixed function, and \cref{thm:calib}
applies to it unchanged; its firing is at most that of the one-sided gate and at most $D(R)$ on every $D$. The miss
event $R^c=\{d>\hat\tau\}$ is the firing event of a score $d$ whose threshold is an order statistic of $n$ i.i.d.\ scores,
so the proof of \cref{thm:calib} with level $\alpha$ applies on $\Dnew$: with $n=1{,}000$, $\hat\tau$ is at least the
980th smallest distance, and the region misses at most 2.9\% of new-task prompts with probability 0.95. The threshold of
the gate is then calibrated on the conjunction.

\textbf{Exact gating and coupling:} A forward pre-hook on block $\lg$ reads $u_t$, computes $w_k(u_t)$ in float32 and
updates the latches, and the LoRA output is merged as $y\leftarrow\mathrm{where}(m_t,\,y+\Delta(x),\,y)$, so masked
positions return the base value bit for bit. The sampler draws one uniform per prompt and step from a fixed seed and
selects the token by inverse transform, so two models decoded with the same seed realise the coupling of
\cref{thm:coupling} exactly. A final model and its base model are always decoded in the same batches.

\textbf{Training:} GRPO \citep{shao2024deepseekmath} samples 8 responses per prompt from the current gated policy and
normalises rewards within the group. All methods share the LoRA placement (rank 16, scaling 32, the attention and MLP
projections of blocks $\lg..L$; embeddings, final normalisation and output head are never adapted), 150 GRPO steps per
task with 16 prompts $\times$ 8 samples at temperature 1, at most 512 new tokens, AdamW with learning rate
$5\cdot10^{-5}$ and gradient clipping at 1. The seed sets the LoRA initialisation and the rollout randomness; prompts,
splits and evaluation pools are fixed.

\textbf{Models:} Qwen2.5-1.5B- and 7B-Instruct (28 blocks, $\lg=9$), OLMo-2-1124-7B-Instruct (32 blocks, $\lg=10$),
Llama-3.2-3B-Instruct (28, 9), DeepSeek-LLM-7B-Chat (30, 10), Llama-3.1-8B-Instruct (32, 10) and Mistral-7B-Instruct-v0.3
(32, 10). Qwen2.5-7B runs NQ, IF and the three-task stream, OLMo-2 NQ, and the other families IF and NQ with Seq-GRPO and
\methodp{}. Nothing was tuned for the other families; only batch sizes were lowered to fit memory.

\textbf{Tasks:} The three-task stream is GSM8K, then Countdown with 3--4 numbers, then Knights and Knaves puzzles with
2--4 people. The five-task stream continues with MATH problems of levels 1--4 with a numeric answer and with $4\times4$
Sudoku. IF appends one verifiable format constraint to held-out Dolly requests; NQ asks closed-book Natural Questions as
plain questions; Concise-ctx and Concise are Dolly requests with and without a passage, rewarded for a short answer;
SQuAD~2.0 is rewarded for the gold answer or for abstaining; TriviaQA is scored as NQ.

\textbf{Fitting and calibration sets:} New task: 600 training prompts, disjoint from the RL prompts on which $\beta$ and
$\hat\tau$ are set. Dolly: 600 (fit), 700 (calibration), 400 (held-out). AdvBench: 100, 170 and 250. Each earlier task:
300 stored training prompts (fit) and 300 (calibration). Task-like prompts from ARC, CommonsenseQA, OpenBookQA, BoolQ,
Alpaca and MBPP: 900 (fit) and 700 (calibration). Evaluation pools: 500 prompts per task, 500 MMLU, 450 XSTest, 400 Dolly
and 250 AdvBench. \Cref{thm:calib} assumes i.i.d.\ draws; AdvBench has 520 requests in all, so its certificate concerns
that finite pool, split without replacement.

\textbf{Baselines:} +KL uses the $k_3$ estimator per token on the current task's samples, with
$\lambda\in\{0.01,0.05,0.2,1\}$. +Ref-KL samples responses of $\pi_{k-1}$ on the gate's own reference prompts and adds
$\lambda\cdot\KL(\pi_{k-1}\|\pi)$ on them, $\lambda\in\{0.1,1,10,100\}$. Null-space LoRA projects the LoRA's input away
from the principal input subspace of 512 reference sequences (80, 95 or 99\% of the energy). EWC uses a Kronecker-factored
diagonal Fisher of the previous model's samples, $\lambda\in\{0.1,1,10\}$, and A-GEM projects the RL gradient so that the
sampled reference KL does not increase to first order. LoRA scaling multiplies the Seq-GRPO update by
$a\in\{0.25,0.5,0.75\}$ at evaluation; O-LoRA adds an orthogonality penalty to earlier LoRAs; replay mixes a quarter of
each batch from earlier tasks; the router is trained under its own prompt-time gate with threshold zero.

\textbf{Metrics:} Task accuracy after each stage, average accuracy and backward transfer (BWT); refusal on AdvBench;
over-refusal on safe XSTest prompts; MMLU accuracy; and for every pool the share of responses that differ from the base
model's. Every metric is a bounded function of the response, so \cref{thm:coupling} covers all of them. Tables report
mean $\pm$ standard deviation over seeds; a held-out bound is the largest over seeds. Every run is compared with the base
model evaluated on its own GPU model (NVIDIA H200 or A100).

\begin{table}[t]
\centering
\scriptsize
\setlength{\tabcolsep}{3pt}
\caption{The gates and deployment forms of the experiments. Every gate is calibrated with \cref{thm:calib}, so each
certifies its firing on every calibrated component; the steps change only which prompts it covers and its coverage
$\beta$. Results use the deployed gate unless marked otherwise.}
\label{tab:variants}
\begin{tabular}{>{\raggedright\arraybackslash}p{2.7cm} >{\raggedright\arraybackslash}p{6.9cm} >{\raggedright\arraybackslash}p{4.5cm}}
\toprule
Name & What it is & Where it is used \\
\midrule
\multicolumn{3}{l}{\emph{Update is deployed}} \\
Gated training & a LoRA trained under its own gate; rollouts on which the gate stays silent carry no gradient &
\methodp{} and \methoda{} rows; Non-fired in \cref{tab:fam} \\
Switch & any trained model answers where a prompt-time gate fires, the old model elsewhere (\cref{cor:switch}) &
``behind the gate'' rows; \cref{tab:fam}, \cref{fig:switchpred} \\
\multicolumn{3}{l}{\emph{Gate reads}} \\
Prompt-time & \methodp{}: reads the last prompt token and decides once per prompt & every switch; most gated runs \\
Anytime & \methoda{}: reads every position after the template prefix & \cref{tab:main} \\
Whole sequence & the prompt-time gate, with the LoRA applied to the whole sequence once it fires & \cref{tab:main} \\
\multicolumn{3}{l}{\emph{Prompts it covers (steps of \cref{sec:practice})}} \\
Deployed, steps (1)--(2) & the classifier, thresholded at $\varepsilon$ on each reference component & coverage, gain and
switch results \\
Two-sided, step (3) & fires only inside the task's region, which misses a share $\alpha=0.02$ of its prompts & chosen
gate for NQ and TriviaQA \\
+ remedy, step (4) & harmful requests in the task's format added as a component & chosen gate for IF and the streams \\
Protect, step (5) & calibrated only on the components to protect & \cref{tab:protect} \\
Protect + passages & adds harmful requests that follow a passage & chosen gate for Concise, Concise-ctx and
SQuAD~2.0 \\
One gate for a stream & the union of the stages' gates, calibrated once (\cref{prop:promptstream}) & \cref{tab:deploy} \\
\bottomrule
\end{tabular}
\end{table}

\textbf{Order of the design choices:}\label{app:protocol} The first version fixed the split at about 30\% of the depth,
the linear classifier, $\varepsilon=0.03$, $\delta=0.05$, the Dolly and AdvBench components and the evaluation pools, and
none of these changed later. The task-like component, the two-sided region, the remedy and the passage component were
each added after a failure (\cref{tab:protocol}). The later tests were fixed before any gate saw them.

\begin{table}[t]
\centering
\scriptsize
\setlength{\tabcolsep}{3pt}
\caption{Each design choice and the evidence it is judged on. Before: fixed before any gate was evaluated on that
evidence, so it is a prospective test; after a failure: chosen after the named failure was seen, so that evidence does
not test it independently.}
\label{tab:protocol}
\begin{tabular}{p{4.6cm} p{4.6cm} p{5.2cm}}
\toprule
Choice & Fixed & Evidence \\
\midrule
Split depth, linear classifier, $\varepsilon=0.03$, $\delta=0.05$, Dolly and AdvBench components and their sizes;
Dolly, AdvBench, XSTest and MMLU pools & first version; never changed & every later result (before) \\
Task-like component & second version, after a failure: first-version gates fired on MMLU & MMLU informed it; never
calibrated on \\
Two-sided region, $\alpha=0.02$, shrinkage $0.1$ & after a failure: MMLU firing of the stream gates and of IF at
$\varepsilon=0.1$ & not MMLU; the jailbreak templates, added later (before) \\
Remedy: harmful requests in the task's format & after a failure: deployed gates fired on such requests & their held-out
split of the same format only \\
Passage component (protect + passages) & after a failure: the protect gate fired on harmful requests with a passage &
held-out requests of the same form (\cref{tab:protect}) \\
10{,}000 fresh Dolly requests & before any gate was evaluated on them & \cref{tab:certs} (before) \\
HarmBench, StrongREJECT & every gate and setting frozen; predictions written before the runs & \cref{tab:extmain}
(before) \\
MultiJail, oasst1, system prompts & gates frozen; predictions written before the download & \cref{tab:prospective}
(before) \\
TriviaQA, settings copied from NQ & seven predictions written before the download; gate firing recorded before
training & \cref{tab:trivia} (before) \\
Training in the agent loop, settings copied from the task stream & no prediction written before the runs; seeds 1
and 2 added after seed 0; tool calls on other sets counted afterwards from the stored episodes &
\cref{tab:agentloop,tab:agentloop_tools} \\
API-call benchmark (SGD), settings copied from the calculator agent & two checks written before the run: no change
without a firing, and firing within the level on the protected sets; seed 0 & \cref{tab:api,tab:api_stages} \\
Refusal fallback, applied to stored responses and episodes & rule fixed before it was applied; no prediction written
& \cref{tab:fallback} \\
Wide harmful component (Concise) & after a failure: the chosen Concise gate fired on MultiJail; five predictions written
before the download & its test third, Do-Not-Answer, Aya red-teaming, XSTest (\cref{tab:widen}) \\
\bottomrule
\end{tabular}
\end{table}

\section{Additional Results}
\label{app:results}

\subsection{Certified Levels}

\begin{table*}[t]
\centering
\footnotesize
\setlength{\tabcolsep}{2.2pt}
\caption{Certified levels and their checks (\%). Every reference component, Dolly among them, is certified at 3\% with
probability 0.95 (\cref{thm:calib}). Dolly, MMLU: held-out firing and, in brackets, its 95\% Clopper--Pearson bound
(\cref{prop:cp}), the only certificate on MMLU. $\beta$: firing on the unlabeled RL prompts, which also set the
two-sided region ($\beta\le98$ there). Streams: union of per-stage gates, each at 3\%. 10k: bound from 10{,}000 fresh
Dolly requests.}
\label{tab:certs}
\begin{tabular}{l rrrr rrr}
\toprule
 & \multicolumn{4}{c}{deployed gate} & \multicolumn{3}{c}{two-sided gate} \\
\cmidrule(lr){2-5}\cmidrule(lr){6-8}
Experiment & $\beta\uparrow$ & Dolly$\downarrow$ & 10k$\downarrow$ & MMLU$\downarrow$ & $\beta\uparrow$ & Dolly$\downarrow$ & MMLU$\downarrow$ \\
\midrule
NQ, Qwen 1.5B & \ovgNqRatioAPower & \ovgNqRatioADolly{} [\ovgNqRatioACpDolly] & 1.7 & \ovgNqRatioAMmlu{} [\ovgNqRatioACpMmlu] & \ovgNqTwoAPower & \ovgNqTwoADolly{} [\ovgNqTwoACpDolly] & \ovgNqTwoAMmlu{} [\ovgNqTwoACpMmlu] \\
NQ, Qwen 7B & \ovgNqsevenRatioAPower & \ovgNqsevenRatioADolly{} [\ovgNqsevenRatioACpDolly] & 2.5 & \ovgNqsevenRatioAMmlu{} [\ovgNqsevenRatioACpMmlu] & \ovgNqsevenTwoAPower & \ovgNqsevenTwoADolly{} [\ovgNqsevenTwoACpDolly] & \ovgNqsevenTwoAMmlu{} [\ovgNqsevenTwoACpMmlu] \\
NQ, OLMo-2 7B & \ovgNqolmoRatioAPower & \ovgNqolmoRatioADolly{} [\ovgNqolmoRatioACpDolly] & 2.2 & \ovgNqolmoRatioAMmlu{} [\ovgNqolmoRatioACpMmlu] & \ovgNqolmoTwoAPower & \ovgNqolmoTwoADolly{} [\ovgNqolmoTwoACpDolly] & \ovgNqolmoTwoAMmlu{} [\ovgNqolmoTwoACpMmlu] \\
IF, Qwen 1.5B & \ovgIfcRatioAPower & \ovgIfcRatioADolly{} [\ovgIfcRatioACpDolly] & -- & \ovgIfcRatioAMmlu{} [\ovgIfcRatioACpMmlu] & \ovgIfcTwoAPower & \ovgIfcTwoADolly{} [\ovgIfcTwoACpDolly] & \ovgIfcTwoAMmlu{} [\ovgIfcTwoACpMmlu] \\
IF, Qwen 7B & \ovgIfcsevenRatioAPower & \ovgIfcsevenRatioADolly{} [\ovgIfcsevenRatioACpDolly] & -- & \ovgIfcsevenRatioAMmlu{} [\ovgIfcsevenRatioACpMmlu] & \ovgIfcsevenTwoAPower & \ovgIfcsevenTwoADolly{} [\ovgIfcsevenTwoACpDolly] & \ovgIfcsevenTwoAMmlu{} [\ovgIfcsevenTwoACpMmlu] \\
Concise-ctx, 1.5B & \ovgConcisectxRatioAPower & \ovgConcisectxRatioADolly{} [\ovgConcisectxRatioACpDolly] & -- & \ovgConcisectxRatioAMmlu{} [\ovgConcisectxRatioACpMmlu] & \ovgConcisectxTwoAPower & \ovgConcisectxTwoADolly{} [\ovgConcisectxTwoACpDolly] & \ovgConcisectxTwoAMmlu{} [\ovgConcisectxTwoACpMmlu] \\
3 tasks, Qwen 1.5B & \ovgStreamRatioAPower & \ovgStreamRatioADolly{} [\ovgStreamRatioACpDolly] & 3.3 & \ovgStreamRatioAMmlu{} [\ovgStreamRatioACpMmlu] & \ovgStreamTwoAPower & \ovgStreamTwoADolly{} [\ovgStreamTwoACpDolly] & \ovgStreamTwoAMmlu{} [\ovgStreamTwoACpMmlu] \\
3 tasks, Qwen 7B & \ovgStreamsevenRatioAPower & \ovgStreamsevenRatioADolly{} [\ovgStreamsevenRatioACpDolly] & 2.8 & \ovgStreamsevenRatioAMmlu{} [\ovgStreamsevenRatioACpMmlu] & \ovgStreamsevenTwoAPower & \ovgStreamsevenTwoADolly{} [\ovgStreamsevenTwoACpDolly] & \ovgStreamsevenTwoAMmlu{} [\ovgStreamsevenTwoACpMmlu] \\
5 tasks, Qwen 1.5B & \ovgStreamfiveRatioAPower & \ovgStreamfiveRatioADolly{} [\ovgStreamfiveRatioACpDolly] & 4.5 & \ovgStreamfiveRatioAMmlu{} [\ovgStreamfiveRatioACpMmlu] & \ovgStreamfiveTwoAPower & \ovgStreamfiveTwoADolly{} [\ovgStreamfiveTwoACpDolly] & \ovgStreamfiveTwoAMmlu{} [\ovgStreamfiveTwoACpMmlu] \\
\bottomrule
\end{tabular}
\end{table*}

\Cref{tab:certs} gives, for every experiment of \cref{fig:certs}, the coverage $\beta$, the bound from 10{,}000 fresh
Dolly requests and the firing on MMLU, which no gate is calibrated on.

\subsection{The Three-Task Stream}
\label{sec:exp-stream}

\textbf{Validity:} Among \resExactPN{} prompt--stage pairs on which no new gate of \methodp{} fired, \resExactPViol{}
responses differed from the previous stage's, and among \resExactN{} such pairs of \methoda{}, \resExactViol{}: an exact
check of \cref{thm:coupling}. \Cref{tab:decode} decodes the reference pools again at temperatures 0.3 and 1 and with
top-$p$ 0.9, the final and the base model sharing one seed: no non-fired response changes (\decNonfired{} of
\decNonfiredChecked). The later gates of \methodp{} fire on none of the stored GSM8K trajectories of $\pi_1$ and on 0.8\%
of the stored Countdown trajectories of $\pi_2$, while Seq-GRPO changes 99.0\% and 96.3\% of those responses
(\cref{thm:stream}).

\textbf{Changes of ungated training:} \Cref{tab:main} gives every method on the stream. Each ungated method changes
most Dolly responses whatever it regularises, and the gated variants almost none at the same average accuracy.
Seq-GRPO also lowers over-refusal on safe XSTest prompts by 15.5 points on average. Apart from Ref-KL, which learns
little Countdown, the averages lie within about one standard deviation of each other, so the stream shows that
certification costs no accuracy, not that it adds any. It has one cost there: before a task is trained, an earlier
stage's gate can fire on its prompts. After the GSM8K stage, \methodp{} lowers Countdown accuracy from 6.8\% to 3.8\%.

\begin{table*}[t]
\centering
\small
\setlength{\tabcolsep}{2.5pt}
\caption{After the stream GSM8K $\to$ Countdown $\to$ K\&K (Qwen2.5-1.5B-Instruct, greedy decoding). Accuracy (\%) on each task, average and backward transfer (BWT); changes relative to the base model in AdvBench refusal, XSTest over-refusal on safe prompts and MMLU accuracy (points); share of Dolly responses that differ from the base model's (Chg.) or use a trained answer format (Leak). Each stage's gate is certified before training at the PAC level 3\% on every reference component (\cref{thm:calib}). The last two columns are held-out bounds, the largest 95\% Clopper--Pearson bound on the firing over held-out prompts of the reference pools Dolly and AdvBench (ref.), which checks that level, and of XSTest and MMLU, which are never used in calibration (off-ref.) and have no other certificate; MMLU sets it; ungated methods have no gate (none), and the router has held-out bounds but no level chosen before training (uncal.). Mean $\pm$ standard deviation over $n$ of 3 seeds, with the same data and evaluation prompts; each bound is the largest over seeds. $^{\mathrm{A}}$: run on A100 GPUs in at least one seed, the others on H200, each compared with the base model on its own GPU model. $^{\ast}$One seed; its per-task accuracy, over-refusal and leak were not computed. Arrows: higher ($\uparrow$) or lower ($\downarrow$) is better; for changes relative to the base model, zero is best.}
\label{tab:main}
\resizebox{\ifdim\width>\linewidth\linewidth\else\width\fi}{!}{%
\begin{tabular}{l c ccc cc ccc cc cc}
\toprule
 & & \multicolumn{5}{c}{Stream tasks $\uparrow$} & \multicolumn{3}{c}{$\Delta$ vs.\ base (pts)} & \multicolumn{2}{c}{Dolly (\%)} & \multicolumn{2}{c}{Held-out bound (\%)} \\
\cmidrule(lr){3-7}\cmidrule(lr){8-10}\cmidrule(lr){11-12}\cmidrule(lr){13-14}
Method & $n$ & GSM8K & Countdown & K\&K & Avg$\uparrow$ & BWT$\uparrow$ & Refusal & Over-ref. & MMLU & Chg.$\downarrow$ & Leak$\downarrow$ & ref.$\downarrow$ & off-ref.$\downarrow$ \\
\midrule
Base model (A100) & -- & 74.4{\scriptsize$\pm$0.0} & 6.8{\scriptsize$\pm$0.0} & 13.2{\scriptsize$\pm$0.0} & 31.5{\scriptsize$\pm$0.0} & -- & 0.0 & 0.0 & 0.0 & 0.0 & 0.0{\scriptsize$\pm$0.0} & 0.0 & 0.0 \\
Base model (H200) & -- & 73.2 & 7.8 & 13.4 & 31.5 & -- & 0.0 & 0.0 & 0.0 & 0.0 & 0.2 & 0.0 & 0.0 \\
\midrule
Seq-GRPO$^{\mathrm{A}}$ & 3 & 74.7{\scriptsize$\pm$4.1} & 61.4{\scriptsize$\pm$7.8} & 35.3{\scriptsize$\pm$5.4} & 57.2{\scriptsize$\pm$4.9} & $-$4.5{\scriptsize$\pm$5.2} & $-$0.5{\scriptsize$\pm$0.5} & $-$15.5{\scriptsize$\pm$15.4} & +1.0{\scriptsize$\pm$1.4} & 93.4{\scriptsize$\pm$3.8} & 1.5{\scriptsize$\pm$2.4} & \multicolumn{2}{c}{none} \\
\quad + KL$^{\mathrm{A}}$ & 3 & 77.5{\scriptsize$\pm$2.0} & 58.3{\scriptsize$\pm$4.1} & 31.9{\scriptsize$\pm$2.1} & 55.9{\scriptsize$\pm$1.9} & +5.3{\scriptsize$\pm$8.8} & $-$0.1{\scriptsize$\pm$0.2} & $-$17.2{\scriptsize$\pm$2.1} & +0.7{\scriptsize$\pm$1.0} & 90.8{\scriptsize$\pm$0.4} & 0.1{\scriptsize$\pm$0.1} & \multicolumn{2}{c}{none} \\
\quad + Replay$^{\mathrm{A}}$ & 3 & 75.5{\scriptsize$\pm$0.6} & 63.2{\scriptsize$\pm$1.3} & 35.9{\scriptsize$\pm$2.2} & 58.2{\scriptsize$\pm$0.1} & +1.2{\scriptsize$\pm$1.7} & +0.0{\scriptsize$\pm$0.0} & $-$14.3{\scriptsize$\pm$0.9} & +0.9{\scriptsize$\pm$0.3} & 90.2{\scriptsize$\pm$4.2} & 1.3{\scriptsize$\pm$1.9} & \multicolumn{2}{c}{none} \\
\quad + Ref-KL$^{\mathrm{A}}$ & 3 & 75.5{\scriptsize$\pm$1.7} & 36.2{\scriptsize$\pm$8.5} & 30.5{\scriptsize$\pm$2.7} & 47.4{\scriptsize$\pm$3.0} & $-$10.2{\scriptsize$\pm$5.6} & $-$0.5{\scriptsize$\pm$0.2} & $-$10.4{\scriptsize$\pm$12.2} & $-$0.5{\scriptsize$\pm$0.4} & 94.4{\scriptsize$\pm$0.4} & 0.0{\scriptsize$\pm$0.0} & \multicolumn{2}{c}{none} \\
\quad + O-LoRA$^{\mathrm{A}}$ & 3 & 76.0{\scriptsize$\pm$0.9} & 64.3{\scriptsize$\pm$6.0} & 39.2{\scriptsize$\pm$1.1} & 59.8{\scriptsize$\pm$2.4} & $-$1.8{\scriptsize$\pm$0.6} & $-$0.5{\scriptsize$\pm$0.2} & $-$12.8{\scriptsize$\pm$3.3} & +1.6{\scriptsize$\pm$0.3} & 90.9{\scriptsize$\pm$2.1} & 2.5{\scriptsize$\pm$2.2} & \multicolumn{2}{c}{none} \\
\quad + Null-space$^{\mathrm{A}}$ & 3 & 76.5{\scriptsize$\pm$1.6} & 63.3{\scriptsize$\pm$6.1} & 36.0{\scriptsize$\pm$1.4} & 58.6{\scriptsize$\pm$2.3} & $-$0.2{\scriptsize$\pm$0.5} & $-$0.1{\scriptsize$\pm$0.2} & $-$6.3{\scriptsize$\pm$1.6} & +0.6{\scriptsize$\pm$0.4} & 74.2{\scriptsize$\pm$1.5} & 1.2{\scriptsize$\pm$0.3} & \multicolumn{2}{c}{none} \\
\quad + A-GEM$^{\ast}$ & 1 & -- & -- & -- & 58.4 & $-$0.7 & $-$0.4 & -- & +0.6 & 94.2 & -- & \multicolumn{2}{c}{none} \\
\quad LoRA $\times$0.5$^{\mathrm{A}}$ & 3 & 77.7{\scriptsize$\pm$0.8} & 60.3{\scriptsize$\pm$2.7} & 26.7{\scriptsize$\pm$2.1} & 54.9{\scriptsize$\pm$1.4} & +0.1{\scriptsize$\pm$1.7} & $-$0.1{\scriptsize$\pm$0.2} & $-$9.2{\scriptsize$\pm$9.8} & +0.6{\scriptsize$\pm$0.7} & 84.8{\scriptsize$\pm$3.8} & 0.7{\scriptsize$\pm$0.9} & \multicolumn{2}{c}{none} \\
Router (uncal.)$^{\mathrm{A}}$ & 3 & 77.5{\scriptsize$\pm$0.5} & 59.7{\scriptsize$\pm$12.0} & 33.9{\scriptsize$\pm$2.2} & 57.1{\scriptsize$\pm$4.6} & +0.0{\scriptsize$\pm$0.0} & +0.0{\scriptsize$\pm$0.0} & +0.0{\scriptsize$\pm$0.0} & +0.0{\scriptsize$\pm$0.0} & 0.0{\scriptsize$\pm$0.0} & 0.0{\scriptsize$\pm$0.0} & \multicolumn{2}{c}{uncal.} \\
\method{}-prompt (ours)$^{\mathrm{A}}$ & 3 & 78.1{\scriptsize$\pm$1.1} & 65.5{\scriptsize$\pm$4.9} & 32.9{\scriptsize$\pm$1.6} & 58.8{\scriptsize$\pm$1.6} & +0.0{\scriptsize$\pm$0.1} & +0.0{\scriptsize$\pm$0.0} & +0.0{\scriptsize$\pm$0.0} & +0.4{\scriptsize$\pm$0.5} & 2.6{\scriptsize$\pm$0.1} & 0.5{\scriptsize$\pm$0.0} & $\le$5.7 & $\le$10.5 \\
\quad whole sequence$^{\mathrm{A}}$ & 3 & 78.3{\scriptsize$\pm$2.0} & 58.2{\scriptsize$\pm$11.7} & 37.1{\scriptsize$\pm$3.6} & 57.8{\scriptsize$\pm$3.2} & +0.0{\scriptsize$\pm$0.1} & +0.0{\scriptsize$\pm$0.0} & +0.0{\scriptsize$\pm$0.0} & +0.0{\scriptsize$\pm$0.2} & 2.2{\scriptsize$\pm$0.4} & 0.7{\scriptsize$\pm$0.1} & $\le$5.7 & $\le$10.5 \\
\method{}-anytime (ours)$^{\mathrm{A}}$ & 3 & 78.0{\scriptsize$\pm$1.6} & 65.0{\scriptsize$\pm$3.6} & 34.3{\scriptsize$\pm$2.9} & 59.1{\scriptsize$\pm$0.9} & +0.1{\scriptsize$\pm$0.1} & +0.0{\scriptsize$\pm$0.0} & +0.0{\scriptsize$\pm$0.0} & +0.3{\scriptsize$\pm$0.4} & 1.0{\scriptsize$\pm$0.5} & 0.3{\scriptsize$\pm$0.1} & $\le$2.9 & $\le$15.7 \\
\bottomrule
\end{tabular}}
\end{table*}

\begin{table}[h]
\centering
\small
\setlength{\tabcolsep}{3pt}
\caption{Drift under other decoding rules (Qwen2.5-1.5B, seed 0; the final and the base model share one decoding seed and are decoded in the same batches). Fire: firing on held-out Dolly prompts under that rule and its 95\% Clopper--Pearson bound, which bounds the TV distance on Dolly under that rule (\cref{thm:coupling}); a prompt-time gate's firing is the same under every rule, an anytime gate's is not. Chg.: share of changed responses, an estimate of an upper bound on the TV distance (coupling inequality). $\Delta$: change of refusal (AdvBench) and accuracy (MMLU), points. TV$\ge$: largest 95\% lower confidence bound on the TV distance over the four pools (Dolly, AdvBench, XSTest, MMLU), from the Kolmogorov--Smirnov distance of response lengths; for ungated runs it, not Chg., is the evidence of drift. Non-fired: changed responses on prompts where no gate fired, over the four pools (\cref{thm:coupling}: 0).}
\label{tab:decode}
\resizebox{\ifdim\width>\linewidth\linewidth\else\width\fi}{!}{%
\begin{tabular}{l l r r r r r r r r}
\toprule
 & & \multicolumn{2}{c}{Dolly (\%)} & \multicolumn{2}{c}{AdvBench} & \multicolumn{2}{c}{MMLU} & & \\
\cmidrule(lr){3-4}\cmidrule(lr){5-6}\cmidrule(lr){7-8}
Run & Decoding & Fire (CP) & Chg. & Chg. & $\Delta$ & Chg. & $\Delta$ & TV$\ge$ & Non-fired \\
\midrule
Stream, \methodp{} & $T=0.3$ & 3.8 (5.7) & 2.5 & 0.0 & +0.0 & 3.0 & $-$0.4 & 0.0 & 0 \\
Stream, \methodp{} & $T=1$ & 3.8 (5.7) & 3.0 & 0.0 & +0.0 & 2.4 & +0.2 & 0.0 & 0 \\
Stream, \methodp{} & $T=1$, top-$p$ 0.9 & 3.8 (5.7) & 2.5 & 0.0 & +0.0 & 2.6 & $-$0.8 & 0.0 & 0 \\
\addlinespace
Stream, \methoda{} & $T=0.3$ & 1.2 (2.6) & 0.8 & 0.0 & +0.0 & 6.0 & +0.8 & 0.0 & 0 \\
Stream, \methoda{} & $T=1$ & 2.0 (3.6) & 1.5 & 0.0 & +0.0 & 6.8 & +0.4 & 0.0 & 0 \\
Stream, \methoda{} & $T=1$, top-$p$ 0.9 & 0.8 (1.9) & 0.5 & 0.0 & +0.0 & 6.2 & +0.6 & 0.0 & 0 \\
\addlinespace
Stream, Seq-GRPO & $T=0.3$ & -- & 88.2 & 9.6 & $-$0.8 & 27.6 & +1.8 & 0.0 & -- \\
Stream, Seq-GRPO & $T=1$ & -- & 95.8 & 49.6 & $-$0.4 & 35.0 & $-$0.2 & 3.8 & -- \\
Stream, Seq-GRPO & $T=1$, top-$p$ 0.9 & -- & 93.8 & 32.0 & +0.0 & 33.2 & +0.8 & 3.8 & -- \\
\addlinespace
NQ, \methodp{} & $T=0.3$ & 3.5 (5.4) & 3.5 & 0.0 & +0.0 & 0.0 & +0.0 & 0.0 & 0 \\
NQ, \methodp{} & $T=1$ & 3.5 (5.4) & 3.5 & 0.0 & +0.0 & 0.0 & +0.0 & 0.0 & 0 \\
NQ, \methodp{} & $T=1$, top-$p$ 0.9 & 3.5 (5.4) & 3.5 & 0.0 & +0.0 & 0.0 & +0.0 & 0.0 & 0 \\
\addlinespace
NQ, Seq-GRPO & $T=0.3$ & -- & 92.8 & 74.8 & $-$6.4 & 38.4 & $-$0.2 & 51.7 & -- \\
NQ, Seq-GRPO & $T=1$ & -- & 97.5 & 92.4 & $-$8.8 & 45.0 & +1.2 & 38.5 & -- \\
NQ, Seq-GRPO & $T=1$, top-$p$ 0.9 & -- & 96.8 & 88.4 & $-$7.2 & 47.0 & $-$0.4 & 51.7 & -- \\
\bottomrule
\end{tabular}}
\end{table}

\subsection{The NQ Task and Calibration}
\label{sec:exp-overlap}

\Cref{tab:overlap_nq} gives every method and level on NQ. \Cref{tab:zoo} builds five gates from the same unlabeled
prompts and trunk features, so that they differ only in the score and the threshold: the logistic classifier, a
Gaussian model with shared covariance, an Expert-Gate-style pair of autoencoders \citep{aljundi2017expertgate}, 15
nearest neighbours and a one-class region. At their own thresholds, as a router or an expert gate would use them, they
fire on 6.2--30.8\% of Dolly prompts. PAC-calibrated at $\varepsilon=0.03$, every one fires on at most 3.2\% and
Seq-GRPO behind it keeps 3.2--4.7 of its 5.2 points. Therefore, calibration, not gating, sets the certified level.

\begin{table*}[t]
\centering
\small
\setlength{\tabcolsep}{3pt}
\caption{New task: Natural Questions asked as plain questions, against Dolly, one reference component; mean $\pm$ standard deviation over $n$ seeds (of 3). Gain: new-task accuracy (points). Changes relative to each run's base model: share of Dolly responses that differ (Chg.), share showing a trained format (Leak), AdvBench refusal, XSTest over-refusal and MMLU accuracy (points). Fires: share of prompts on which the gate fires. CP: largest held-out 95\% Clopper--Pearson bound on the Dolly firing rate over seeds (\cref{prop:cp}), a check of the certified level of 3\% (\cref{thm:calib}). Rows marked $^{\mathrm{A}}$ ran on A100 GPUs (in at least one seed), the others on H200; each run is compared with the base model evaluated on its own GPU model. Ungated methods have no gate: no firing (--) and no certificate (none). The oracle row has no standard deviation. The uncalibrated router has a held-out firing rate but no level chosen before training (uncal.).}
\label{tab:overlap_nq}
\resizebox{\ifdim\width>\linewidth\linewidth\else\width\fi}{!}{%
\begin{tabular}{l c r rr rrr rr r}
\toprule
 & & NQ & \multicolumn{5}{c}{Change on reference} & \multicolumn{2}{c}{Fires (\%)} & \\
\cmidrule(lr){4-8}\cmidrule(lr){9-10}
Method & $n$ & gain$\uparrow$ & Chg.$\downarrow$ & Leak$\downarrow$ & Refusal & Over-ref. & MMLU & NQ & Dolly & CP$\downarrow$ \\
\midrule
Seq-GRPO$^{\mathrm{A}}$ & 3 & +5.2{\scriptsize$\pm$0.9} & 94.2{\scriptsize$\pm$1.8} & +0.0{\scriptsize$\pm$0.0} & $-$8.8{\scriptsize$\pm$2.8} & $-$50.3{\scriptsize$\pm$2.0} & +0.2{\scriptsize$\pm$0.2} & -- & -- & none \\
Seq, scaled $\times0.25$$^{\mathrm{A}}$ & 3 & +3.0{\scriptsize$\pm$0.3} & 74.1{\scriptsize$\pm$1.2} & +0.0{\scriptsize$\pm$0.0} & $-$0.3{\scriptsize$\pm$0.5} & $-$27.7{\scriptsize$\pm$2.0} & +0.0{\scriptsize$\pm$0.0} & -- & -- & none \\
Seq, scaled $\times0.5$$^{\mathrm{A}}$ & 3 & +4.6{\scriptsize$\pm$0.4} & 86.4{\scriptsize$\pm$2.8} & +0.0{\scriptsize$\pm$0.0} & $-$0.8{\scriptsize$\pm$0.4} & $-$40.7{\scriptsize$\pm$1.3} & +0.3{\scriptsize$\pm$0.3} & -- & -- & none \\
Seq, scaled $\times0.75$$^{\mathrm{A}}$ & 3 & +4.9{\scriptsize$\pm$1.2} & 92.2{\scriptsize$\pm$1.8} & +0.0{\scriptsize$\pm$0.0} & $-$4.1{\scriptsize$\pm$1.3} & $-$47.5{\scriptsize$\pm$2.1} & +0.1{\scriptsize$\pm$0.4} & -- & -- & none \\
+ KL $\lambda_{\mathrm{KL}}{=}0.01$$^{\mathrm{A}}$ & 3 & +4.9{\scriptsize$\pm$0.4} & 87.8{\scriptsize$\pm$1.1} & +0.0{\scriptsize$\pm$0.0} & $-$0.3{\scriptsize$\pm$0.2} & $-$38.7{\scriptsize$\pm$1.2} & +1.0{\scriptsize$\pm$0.9} & -- & -- & none \\
+ KL $\lambda_{\mathrm{KL}}{=}0.05$$^{\mathrm{A}}$ & 3 & +3.5{\scriptsize$\pm$0.1} & 73.3{\scriptsize$\pm$2.3} & +0.0{\scriptsize$\pm$0.0} & +0.0{\scriptsize$\pm$0.0} & $-$23.2{\scriptsize$\pm$4.5} & +0.3{\scriptsize$\pm$0.1} & -- & -- & none \\
+ KL $\lambda_{\mathrm{KL}}{=}0.2$$^{\mathrm{A}}$ & 3 & +1.9{\scriptsize$\pm$0.7} & 64.2{\scriptsize$\pm$0.9} & +0.0{\scriptsize$\pm$0.0} & +0.0{\scriptsize$\pm$0.0} & $-$10.3{\scriptsize$\pm$5.6} & +0.5{\scriptsize$\pm$0.2} & -- & -- & none \\
+ KL $\lambda_{\mathrm{KL}}{=}1$$^{\mathrm{A}}$ & 3 & +0.7{\scriptsize$\pm$0.4} & 56.0{\scriptsize$\pm$2.4} & +0.0{\scriptsize$\pm$0.0} & +0.0{\scriptsize$\pm$0.0} & $-$2.3{\scriptsize$\pm$0.6} & +0.0{\scriptsize$\pm$0.2} & -- & -- & none \\
+ Ref-KL $\lambda_{\mathrm{KL}}{=}0.1$$^{\mathrm{A}}$ & 3 & +5.5{\scriptsize$\pm$1.3} & 82.9{\scriptsize$\pm$1.3} & +0.0{\scriptsize$\pm$0.0} & +0.0{\scriptsize$\pm$0.0} & $-$38.1{\scriptsize$\pm$3.4} & +0.4{\scriptsize$\pm$0.2} & -- & -- & none \\
+ Ref-KL $\lambda_{\mathrm{KL}}{=}1$$^{\mathrm{A}}$ & 3 & +2.3{\scriptsize$\pm$0.1} & 86.9{\scriptsize$\pm$0.4} & +0.0{\scriptsize$\pm$0.0} & $-$0.4{\scriptsize$\pm$0.4} & $-$22.8{\scriptsize$\pm$9.0} & +1.4{\scriptsize$\pm$1.1} & -- & -- & none \\
+ Ref-KL $\lambda_{\mathrm{KL}}{=}10$$^{\mathrm{A}}$ & 3 & +1.3{\scriptsize$\pm$0.8} & 86.5{\scriptsize$\pm$0.9} & +0.0{\scriptsize$\pm$0.0} & $-$0.3{\scriptsize$\pm$0.2} & $-$7.5{\scriptsize$\pm$2.8} & +1.9{\scriptsize$\pm$1.3} & -- & -- & none \\
+ Ref-KL, 4 samples $\lambda_{\mathrm{KL}}{=}10$$^{\mathrm{A}}$ & 3 & +0.6{\scriptsize$\pm$0.9} & 73.2{\scriptsize$\pm$0.8} & $-$0.1{\scriptsize$\pm$0.1} & +0.0{\scriptsize$\pm$0.0} & $-$2.9{\scriptsize$\pm$1.3} & +0.5{\scriptsize$\pm$0.3} & -- & -- & none \\
+ Ref-KL, 4 samples $\lambda_{\mathrm{KL}}{=}100$$^{\mathrm{A}}$ & 3 & +0.9{\scriptsize$\pm$0.1} & 73.2{\scriptsize$\pm$1.7} & +0.0{\scriptsize$\pm$0.0} & +0.0{\scriptsize$\pm$0.0} & $-$0.7{\scriptsize$\pm$2.4} & +0.8{\scriptsize$\pm$0.2} & -- & -- & none \\
+ Null-space $e{=}0.8$$^{\mathrm{A}}$ & 3 & +4.8{\scriptsize$\pm$0.7} & 97.1{\scriptsize$\pm$2.6} & +0.0{\scriptsize$\pm$0.0} & $-$30.5{\scriptsize$\pm$39.8} & $-$50.5{\scriptsize$\pm$1.6} & $-$0.1{\scriptsize$\pm$1.2} & -- & -- & none \\
+ Null-space $e{=}0.95$$^{\mathrm{A}}$ & 3 & +5.5{\scriptsize$\pm$0.1} & 98.8{\scriptsize$\pm$0.7} & +0.0{\scriptsize$\pm$0.0} & $-$9.1{\scriptsize$\pm$2.6} & $-$50.5{\scriptsize$\pm$1.2} & +0.2{\scriptsize$\pm$1.1} & -- & -- & none \\
+ Null-space $e{=}0.99$$^{\mathrm{A}}$ & 3 & +5.0{\scriptsize$\pm$0.7} & 95.2{\scriptsize$\pm$4.1} & +0.0{\scriptsize$\pm$0.0} & $-$7.3{\scriptsize$\pm$4.0} & $-$50.1{\scriptsize$\pm$1.0} & +0.2{\scriptsize$\pm$1.1} & -- & -- & none \\
+ Null-space, refreshed $e{=}0.95$$^{\mathrm{A}}$ & 3 & +5.3{\scriptsize$\pm$0.5} & 93.7{\scriptsize$\pm$1.6} & +0.0{\scriptsize$\pm$0.0} & $-$2.8{\scriptsize$\pm$2.8} & $-$45.7{\scriptsize$\pm$3.0} & +0.1{\scriptsize$\pm$0.2} & -- & -- & none \\
+ EWC $\lambda{=}0.1$ & 3 & +5.8{\scriptsize$\pm$0.2} & 96.3{\scriptsize$\pm$4.2} & $-$0.2{\scriptsize$\pm$0.1} & $-$8.1{\scriptsize$\pm$4.5} & $-$51.1{\scriptsize$\pm$1.3} & +0.1{\scriptsize$\pm$0.9} & -- & -- & none \\
+ EWC $\lambda{=}1$ & 3 & +6.6{\scriptsize$\pm$1.0} & 97.5{\scriptsize$\pm$3.1} & $-$0.2{\scriptsize$\pm$0.0} & $-$10.9{\scriptsize$\pm$4.9} & $-$51.9{\scriptsize$\pm$2.0} & +0.2{\scriptsize$\pm$0.3} & -- & -- & none \\
+ EWC $\lambda{=}10$ & 3 & +4.9{\scriptsize$\pm$1.5} & 95.8{\scriptsize$\pm$5.0} & $-$0.2{\scriptsize$\pm$0.0} & $-$5.3{\scriptsize$\pm$3.7} & $-$51.1{\scriptsize$\pm$1.2} & +0.4{\scriptsize$\pm$0.0} & -- & -- & none \\
+ A-GEM$^{\mathrm{A}}$ & 3 & +5.4{\scriptsize$\pm$0.2} & 93.9{\scriptsize$\pm$0.1} & $-$0.1{\scriptsize$\pm$0.1} & $-$4.7{\scriptsize$\pm$1.5} & $-$49.9{\scriptsize$\pm$1.8} & $-$0.1{\scriptsize$\pm$0.5} & -- & -- & none \\
\midrule
Router (uncal.)$^{\mathrm{A}}$ & 3 & +5.4{\scriptsize$\pm$0.7} & 6.8{\scriptsize$\pm$0.0} & +0.0{\scriptsize$\pm$0.0} & +0.0{\scriptsize$\pm$0.0} & $-$5.6{\scriptsize$\pm$0.0} & +0.0{\scriptsize$\pm$0.0} & 97.4{\scriptsize$\pm$0.0} & 6.8{\scriptsize$\pm$0.0} & uncal. \\
\method{}-prompt $\varepsilon{=}0.03$$^{\mathrm{A}}$ & 3 & +4.0{\scriptsize$\pm$1.0} & 3.5{\scriptsize$\pm$0.0} & +0.0{\scriptsize$\pm$0.0} & +0.0{\scriptsize$\pm$0.0} & $-$0.8{\scriptsize$\pm$0.0} & +0.0{\scriptsize$\pm$0.0} & 90.8{\scriptsize$\pm$0.0} & 3.5{\scriptsize$\pm$0.0} & 5.4 \\
\method{}-prompt $\varepsilon{=}0.1$$^{\mathrm{A}}$ & 3 & +5.3{\scriptsize$\pm$0.3} & 9.4{\scriptsize$\pm$0.1} & +0.0{\scriptsize$\pm$0.0} & +0.0{\scriptsize$\pm$0.0} & $-$11.9{\scriptsize$\pm$0.2} & +0.0{\scriptsize$\pm$0.0} & 97.8{\scriptsize$\pm$0.0} & 9.5{\scriptsize$\pm$0.0} & 12.3 \\
\method{}-prompt $\varepsilon{=}0.3$$^{\mathrm{A}}$ & 3 & +5.9{\scriptsize$\pm$2.4} & 28.9{\scriptsize$\pm$0.1} & $-$0.1{\scriptsize$\pm$0.1} & $-$0.5{\scriptsize$\pm$0.5} & $-$34.8{\scriptsize$\pm$0.7} & +0.0{\scriptsize$\pm$0.2} & 99.9{\scriptsize$\pm$0.1} & 28.9{\scriptsize$\pm$0.1} & 33.0 \\
\method{}-prompt, whole seq. $\varepsilon{=}0.03$$^{\mathrm{A}}$ & 3 & +5.9{\scriptsize$\pm$2.3} & 3.5{\scriptsize$\pm$0.0} & +0.0{\scriptsize$\pm$0.0} & +0.0{\scriptsize$\pm$0.0} & $-$0.8{\scriptsize$\pm$0.0} & +0.0{\scriptsize$\pm$0.0} & 90.8{\scriptsize$\pm$0.0} & 3.5{\scriptsize$\pm$0.0} & 5.4 \\
\method{}-prompt, whole seq. $\varepsilon{=}0.1$$^{\mathrm{A}}$ & 3 & +6.2{\scriptsize$\pm$2.6} & 9.5{\scriptsize$\pm$0.0} & +0.0{\scriptsize$\pm$0.0} & +0.0{\scriptsize$\pm$0.0} & $-$12.0{\scriptsize$\pm$0.0} & +0.0{\scriptsize$\pm$0.0} & 97.8{\scriptsize$\pm$0.0} & 9.5{\scriptsize$\pm$0.0} & 12.3 \\
\method{}-anytime $\varepsilon{=}0.03$$^{\mathrm{A}}$ & 3 & +0.3{\scriptsize$\pm$0.2} & 0.9{\scriptsize$\pm$0.1} & +0.0{\scriptsize$\pm$0.0} & +0.0{\scriptsize$\pm$0.0} & $-$0.1{\scriptsize$\pm$0.2} & +0.0{\scriptsize$\pm$0.0} & 11.7{\scriptsize$\pm$2.2} & 1.7{\scriptsize$\pm$0.4} & 3.9 \\
\method{}-anytime $\varepsilon{=}0.1$$^{\mathrm{A}}$ & 3 & +1.2{\scriptsize$\pm$0.5} & 6.4{\scriptsize$\pm$0.6} & +0.0{\scriptsize$\pm$0.0} & +0.0{\scriptsize$\pm$0.0} & $-$0.9{\scriptsize$\pm$0.2} & +0.0{\scriptsize$\pm$0.0} & 39.8{\scriptsize$\pm$1.0} & 7.2{\scriptsize$\pm$1.3} & 11.4 \\
\method{}-anytime $\varepsilon{=}0.3$$^{\mathrm{A}}$ & 3 & +3.1{\scriptsize$\pm$0.4} & 23.3{\scriptsize$\pm$2.5} & +0.0{\scriptsize$\pm$0.0} & +0.0{\scriptsize$\pm$0.0} & $-$3.6{\scriptsize$\pm$0.4} & +0.2{\scriptsize$\pm$0.2} & 78.3{\scriptsize$\pm$1.7} & 25.2{\scriptsize$\pm$2.0} & 30.9 \\
\midrule
\textit{Oracle task gate (uses task labels)} & 3 & +5.2{\scriptsize$\pm$0.9} & 0.0 & +0.0 & +0.0 & +0.0 & +0.0 & 100.0 & 0.0 & -- \\
\bottomrule
\end{tabular}}
\end{table*}

\begin{table}[t]
\centering
\small
\setlength{\tabcolsep}{2.5pt}
\caption{Gates built from the same unlabeled prompts and trunk features (NQ, top, and Concise-ctx, bottom; Qwen2.5-1.5B-Instruct), at their own threshold (natural; as a router uses them) and PAC-calibrated. Power: firing on unlabeled RL prompts; Dolly, MMLU, AdvB.: on held-out pools (\%); CP: 95\% bound on Dolly; Seq$^{\mathrm{sw}}$: Seq-GRPO behind the gate, new-task gain (points) and Dolly responses changed (\%). Only the calibrated rows carry a certificate.}
\label{tab:zoo}
\resizebox{\ifdim\width>\linewidth\linewidth\else\width\fi}{!}{%
\begin{tabular}{l r l r r r r r r r}
\toprule
Gate & AUROC & Threshold & Power$\uparrow$ & Dolly$\downarrow$ & CP$\downarrow$ & MMLU & AdvB. & Gain$^{\mathrm{sw}}$ & Chg.$^{\mathrm{sw}}$ \\
\midrule
\multicolumn{10}{l}{\textbf{NQ}} \\
linear classifier (logistic) & 0.989 & natural & 96.9 & 6.2 & 8.6 & 0.0 & 0.0 & +5.1 & 6.2 \\
 &  & $\varepsilon=0.03$ & 92.2 & 3.2 & 5.1 & 0.0 & 0.0 & +4.7 & 3.2 \\
 &  & $\varepsilon=0.1$ & 98.3 & 9.5 & 12.3 & 0.0 & 0.0 & +5.1 & 9.5 \\
\midrule
Gaussian (LDA) & 0.993 & natural & 98.7 & 7.8 & 10.3 & 0.2 & 0.0 & +5.3 & 7.8 \\
 &  & $\varepsilon=0.03$ & 95.4 & 3.2 & 5.1 & 0.0 & 0.0 & +4.7 & 3.2 \\
 &  & $\varepsilon=0.1$ & 99.5 & 10.5 & 13.4 & 0.8 & 0.0 & +5.3 & 10.5 \\
\midrule
PCA autoencoders, Expert-Gate style & 0.980 & natural & 100.0 & 23.2 & 27.0 & 0.0 & 0.0 & +5.2 & 23.2 \\
 &  & $\varepsilon=0.03$ & 82.8 & 3.0 & 4.8 & 0.0 & 0.0 & +4.5 & 3.0 \\
 &  & $\varepsilon=0.1$ & 97.0 & 10.5 & 13.4 & 0.0 & 0.0 & +5.2 & 10.5 \\
\midrule
15 nearest neighbours & 0.975 & natural & 99.7 & 20.2 & 23.8 & 0.0 & 0.0 & +5.2 & 20.2 \\
 &  & $\varepsilon=0.03$ & 68.8 & 1.0 & 2.3 & 0.0 & 0.0 & +4.1 & 1.0 \\
 &  & $\varepsilon=0.1$ & 96.4 & 11.2 & 14.2 & 0.0 & 0.0 & +5.4 & 11.2 \\
\midrule
one-class (new-task region) & 0.942 & natural & 98.0 & 30.8 & 34.8 & 0.0 & 0.0 & +5.2 & 30.8 \\
 &  & $\varepsilon=0.03$ & 58.1 & 2.5 & 4.2 & 0.0 & 0.0 & +3.2 & 2.5 \\
 &  & $\varepsilon=0.1$ & 82.1 & 7.8 & 10.3 & 0.0 & 0.0 & +4.2 & 7.8 \\
\midrule
\multicolumn{10}{l}{\textbf{Concise-ctx: Dolly requests with a context passage}} \\
linear classifier (logistic) & 0.839 & natural & 81.0 & 28.2 & 32.2 & 7.2 & 0.0 & +53.8 & 26.8 \\
 &  & $\varepsilon=0.03$ & 3.8 & 1.8 & 3.3 & 0.0 & 0.0 & +2.6 & 1.8 \\
 &  & $\varepsilon=0.1$ & 24.1 & 7.5 & 10.0 & 0.0 & 0.0 & +17.7 & 7.3 \\
\midrule
Gaussian (LDA) & 0.838 & natural & 98.5 & 33.0 & 37.1 & 0.0 & 0.0 & +64.0 & 31.5 \\
 &  & $\varepsilon=0.03$ & 3.5 & 2.0 & 3.6 & 0.0 & 0.0 & +3.1 & 2.0 \\
 &  & $\varepsilon=0.1$ & 24.4 & 8.5 & 11.2 & 0.0 & 0.0 & +14.8 & 8.0 \\
\midrule
PCA autoencoders, Expert-Gate style & 0.826 & natural & 99.6 & 33.2 & 37.3 & 0.0 & 0.0 & +65.0 & 31.8 \\
 &  & $\varepsilon=0.03$ & 3.5 & 1.5 & 2.9 & 0.0 & 0.0 & +1.4 & 1.5 \\
 &  & $\varepsilon=0.1$ & 25.8 & 9.0 & 11.7 & 0.0 & 0.0 & +14.2 & 8.8 \\
\midrule
15 nearest neighbours & 0.832 & natural & 99.9 & 33.5 & 37.6 & 0.2 & 0.0 & +65.2 & 32.0 \\
 &  & $\varepsilon=0.03$ & 1.8 & 0.0 & 0.7 & 0.0 & 0.0 & +0.4 & 0.0 \\
 &  & $\varepsilon=0.1$ & 10.6 & 3.8 & 5.7 & 0.0 & 0.0 & +6.2 & 3.6 \\
\midrule
one-class (new-task region) & 0.827 & natural & 98.0 & 32.5 & 36.6 & 0.0 & 0.0 & +62.6 & 31.0 \\
 &  & $\varepsilon=0.03$ & 8.4 & 3.0 & 4.8 & 0.0 & 0.0 & +5.8 & 2.9 \\
 &  & $\varepsilon=0.1$ & 23.9 & 8.2 & 10.9 & 0.0 & 0.0 & +14.9 & 7.9 \\
\bottomrule
\end{tabular}}
\end{table}

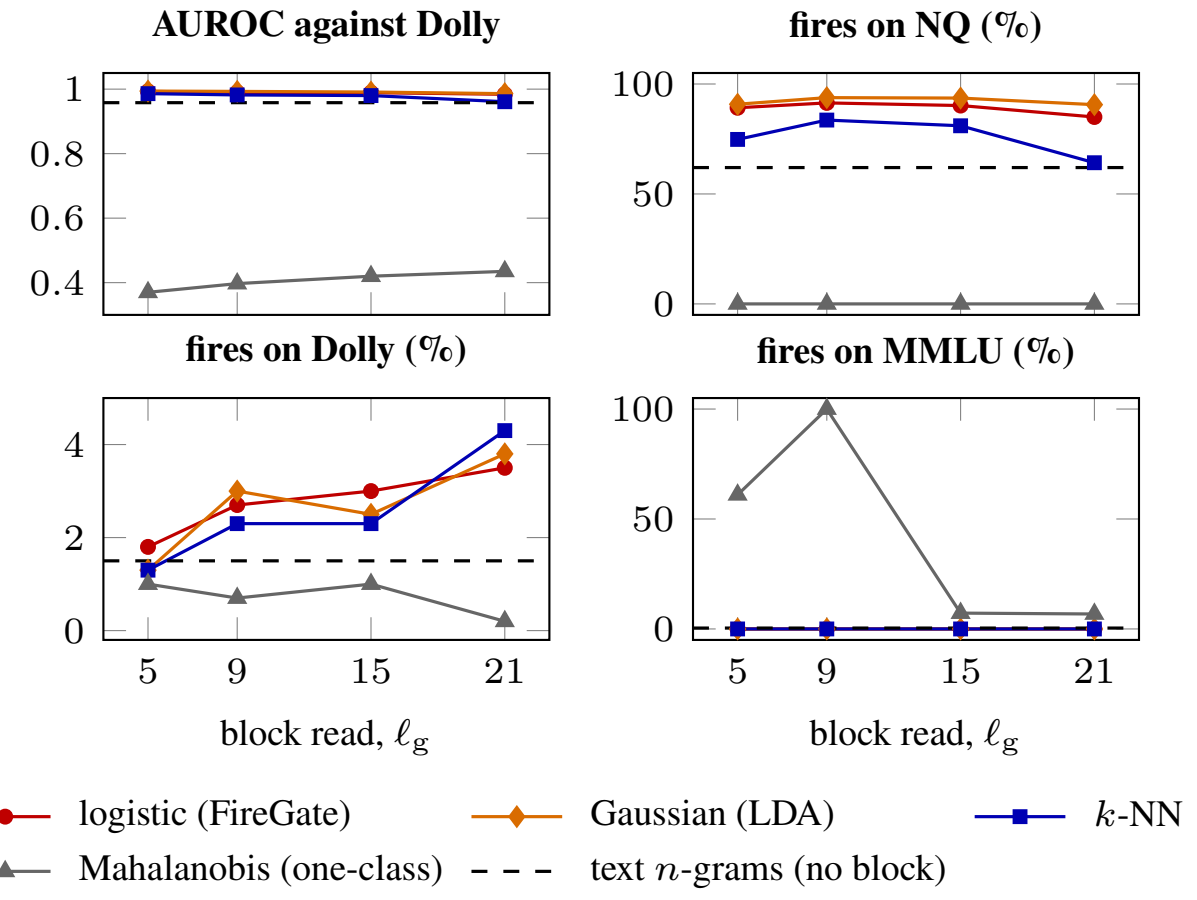
\begin{figure}[t]
\centering
\newcommand{\scoreplots}[1]{%
\addplot[red!75!black, mark=*, mark size=1.2pt] table[x=L, y=#1, row sep=\\] {%
L auroc f03 f1 f3 dolly mmlu \\
5 0.993 89.2 98.6 100.0 1.8 0.0 \\
9 0.99 91.4 97.2 100.0 2.7 0.0 \\
15 0.989 90.2 98.2 100.0 3.0 0.0 \\
21 0.984 85.0 96.8 100.0 3.5 0.0 \\
};
\addplot[orange!85!black, mark=diamond*, mark size=1.7pt] table[x=L, y=#1, row sep=\\] {%
L auroc f03 f1 f3 dolly mmlu \\
5 0.994 90.8 98.4 100.0 1.3 0.0 \\
9 0.993 93.8 98.8 100.0 3.0 0.0 \\
15 0.991 93.6 98.0 99.8 2.5 0.0 \\
21 0.986 90.6 97.4 100.0 3.8 0.0 \\
};
\addplot[blue!70!black, mark=square*, mark size=1.1pt] table[x=L, y=#1, row sep=\\] {%
L auroc f03 f1 f3 dolly mmlu \\
5 0.986 74.8 95.2 100.0 1.3 0.0 \\
9 0.982 83.6 95.8 99.8 2.3 0.0 \\
15 0.98 81.0 93.6 99.2 2.3 0.0 \\
21 0.961 64.2 90.0 99.2 4.3 0.0 \\
};
\addplot[gray!80!black, mark=triangle*, mark size=1.6pt] table[x=L, y=#1, row sep=\\] {%
L auroc f03 f1 f3 dolly mmlu \\
5 0.37 0.0 1.0 8.0 1.0 61.0 \\
9 0.397 0.0 2.2 9.0 0.7 100.0 \\
15 0.42 0.0 0.2 5.8 1.0 7.2 \\
21 0.435 0.0 1.4 9.8 0.2 6.8 \\
};
\addplot[black, dashed] table[x=L, y=#1, row sep=\\] {%
L auroc f03 f1 f3 dolly mmlu \\
3 0.958 62.0 82.4 96.6 1.5 0.4 \\
23 0.958 62.0 82.4 96.6 1.5 0.4 \\
};}%
\begin{tikzpicture}
\begin{groupplot}[group style={group size=2 by 2, horizontal sep=0.95cm, vertical sep=0.55cm,
    xlabels at=edge bottom, xticklabels at=edge bottom},
  scale only axis, width=2.95cm, height=1.6cm, xmin=3, xmax=23, xtick={5,9,15,21},
  xlabel={block read, $\lg$}, tick label style={font=\scriptsize}, label style={font=\scriptsize},
  title style={font=\scriptsize\bfseries, yshift=-0.9ex}, every axis plot/.append style={line width=0.6pt},
  legend style={font=\scriptsize, draw=none, column sep=3pt}, legend columns=3, legend cell align=left]
\nextgroupplot[title={AUROC against Dolly}, ymin=0.3, ymax=1.05, ytick={0.4,0.6,0.8,1.0}, legend to name=scoreleg]
\scoreplots{auroc}
\legend{logistic (\method{}), Gaussian (LDA), $k$-NN, Mahalanobis (one-class), text $n$-grams (no block)}
\nextgroupplot[title={fires on NQ (\%)}, ymin=-5, ymax=105, ytick={0,50,100}]
\scoreplots{f03}
\nextgroupplot[title={fires on Dolly (\%)}, ymin=-0.2, ymax=5, ytick={0,2,4}]
\scoreplots{dolly}
\nextgroupplot[title={fires on MMLU (\%)}, ymin=-5, ymax=105, ytick={0,50,100}]
\scoreplots{mmlu}
\end{groupplot}
\end{tikzpicture}\par\ref{scoreleg}
\caption{Gate scores for NQ against the reference, Qwen2.5-1.5B, before any training, by the block $\lg$ the score
reads (text $n$-grams read none; dashed). Each score is calibrated by its PAC rule at $\varepsilon=0.03$; fires: share
of held-out prompts on which it fires. Firing on NQ lower-bounds $\beta^\ast(\varepsilon)$.}
\label{fig:score}
\end{figure}

\subsection{Protecting Chosen Behaviours}
\label{sec:exp-protect}

\begin{table}[t]
\centering
\footnotesize
\setlength{\tabcolsep}{3.2pt}
\caption{Protecting chosen behaviours on SQuAD~2.0, a task that overlaps general chat (Qwen2.5-1.5B,
$\varepsilon=0.03$). Deployed: calibrated on every component; protect: on harmful and task-like prompts only;
+ passages: harmful requests followed by a passage added. $\beta$: coverage of RL prompts; changes in points;
responses changed (\%); harmful + passage: firing and refusal change on 250 such held-out requests, as given / as a
question (seed 0; else three seeds).}
\label{tab:protect}
\begin{tabular}{l r r rr rrr rr}
\toprule
 & & & \multicolumn{2}{c}{Change} & \multicolumn{3}{c}{Responses changed (\%)} & \multicolumn{2}{c}{Harmful + passage} \\
\cmidrule(lr){4-5}\cmidrule(lr){6-8}\cmidrule(lr){9-10}
Method & $\beta\uparrow$ & Gain$\uparrow$ & Refusal & MMLU & AdvBench & MMLU & Dolly & Gate fires (\%) & Refusal \\
\midrule
Seq-GRPO & -- & +24.4 & $-$0.3 & $-$2.9 & 77.1 & 49.3 & 99.3 & -- & $-$36.8 / $-$74.8 \\
\quad behind the deployed gate & 85.4 & +18.5 & +0.0 & +0.0 & 0.0 & 0.0 & 1.0 & 0.4 / 0.4 & $-$0.4 / +0.0 \\
\quad behind protect & 100.0 & +24.4 & +0.0 & $-$0.2 & 0.8 & 0.5 & 46.3 & 98.4 / 100.0 & $-$36.8 / $-$74.8 \\
\quad behind protect + passages & 99.8 & +24.3 & +0.0 & +0.0 & 0.0 & 0.0 & 16.3 & 0.8 / 1.6 & $-$0.8 / $-$1.2 \\
Ref-KL, protected & -- & +19.9 & +0.0 & +1.1 & 8.8 & 23.9 & 89.8 & -- & -- \\
Safety replay & -- & +27.4 & +0.0 & $-$1.6 & 52.5 & 35.0 & 98.1 & -- & $-$13.6 / $-$61.2 \\
\methodp{}, deployed gate & -- & +15.9 & +0.0 & +0.0 & 0.0 & 0.0 & 1.0 & -- & +0.0 / +0.0 \\
\methodp{}, protect & -- & +33.3 & +0.0 & $-$0.1 & 0.5 & 0.2 & 44.8 & -- & $-$28.0 / $-$80.0 \\
\bottomrule
\end{tabular}
\end{table}

On SQuAD~2.0, Seq-GRPO's refusal rate barely moves, yet it changes most of its responses to held-out harmful requests
and half of its MMLU answers, and it refuses far less often when a harmful request comes with a passage
(\cref{tab:protect}). The protect gate covers the task but also fires on almost all harmful requests that follow a
passage, which are not drawn like the protected ones. With such requests added as a component, it fires on at most
1.6\% of them, and Seq-GRPO behind it keeps almost all of its gain. Ref-KL and safety replay hold the rates they target,
not the responses.

\subsection{Prompts Outside the Reference}
\label{app:prospective}\label{app:external}\label{app:trivia}

\textbf{Pools unlike the task:} \Cref{tab:gate_overview} lists the firing of each gate variant on held-out pools, all
computed before training; by \cref{cor:switch} it bounds the drift of any model behind the gate. The deployed IF gate
and the stream gates fire on many MMLU prompts, which are far from both the task and the reference. The region check
caps the firing at the mass a pool puts in the task's region, and these pools put almost none there, so the two-sided
gate fires on none of them.

\begin{table}[t]
\centering
\small
\setlength{\tabcolsep}{2.5pt}
\caption{Firing of gate variants on held-out prompts (\%, $\varepsilon=0.03$), computed from unlabeled prompts before any training; it bounds the drift of every model behind the gate. Deployed: the deployed prompt-time gate; + region: the two-sided gate; one budget: the union of a stream's gates calibrated at $\varepsilon$; + harmful in format: AdvBench requests in the tasks' formats added to the reference; protect: calibrated on AdvBench and task-like prompts only. Power: firing on unlabeled new-task prompts; CP: 95\% Clopper--Pearson bound on Dolly; Wrap: AdvBench requests in a trained task's format; Sys: Dolly requests behind a system prompt; JB: AdvBench requests in jailbreak templates (mean of three). XSTest, MMLU and JB are never used in calibration.}
\label{tab:gate_overview}
\resizebox{\ifdim\width>\linewidth\linewidth\else\width\fi}{!}{%
\begin{tabular}{l l r rr r r r r r}
\toprule
Experiment & Gate & Power$\uparrow$ & Dolly$\downarrow$ & CP$\downarrow$ & XSTest & MMLU & Wrap & Sys & JB \\
\midrule
NQ, 1.5B & deployed & 92.2 & 3.2 & 5.1 & 2.0 & 0.0 & -- & 0.0 & 0.0 \\
 & + region & 90.7 & 2.8 & 4.5 & 2.0 & 0.0 & -- & 0.0 & 0.0 \\
\midrule
NQ, 7B & deployed & 97.9 & 2.8 & 4.5 & 1.6 & 0.0 & -- & 0.0 & 0.0 \\
 & + region & 96.1 & 2.0 & 3.6 & 1.6 & 0.0 & -- & 0.0 & 0.0 \\
\midrule
NQ, OLMo-2 & deployed & 96.8 & 2.2 & 3.9 & 2.4 & 0.0 & -- & 0.0 & 0.0 \\
 & + region & 95.0 & 2.2 & 3.9 & 2.4 & 0.0 & -- & 0.0 & 0.0 \\
\midrule
IF, 1.5B & deployed & 100.0 & 1.5 & 2.9 & 0.0 & 99.4 & 100.0 & 73.5 & 34.0 \\
 & + region & 98.0 & 0.0 & 0.7 & 0.0 & 0.0 & 94.0 & 0.0 & 0.0 \\
 & + harmful in format & 98.4 & 0.0 & 0.7 & 0.0 & 0.0 & 0.0 & 0.0 & 0.0 \\
 & + region, + harmful in format & 96.5 & 0.0 & 0.7 & 0.0 & 0.0 & 0.0 & 0.0 & 0.0 \\
\midrule
IF, 7B & deployed & 100.0 & 1.8 & 3.3 & 0.0 & 100.0 & 100.0 & 10.8 & 49.7 \\
 & + region & 98.0 & 0.0 & 0.7 & 0.0 & 0.0 & 99.6 & 0.0 & 0.0 \\
 & + harmful in format & 99.5 & 0.0 & 0.7 & 0.0 & 19.6 & 0.0 & 0.0 & 0.0 \\
 & + region, + harmful in format & 97.6 & 0.0 & 0.7 & 0.0 & 0.0 & 0.0 & 0.0 & 0.0 \\
\midrule
Concise, 1.5B & deployed & 0.7 & 0.8 & 1.9 & 0.0 & 2.0 & -- & 0.8 & 0.0 \\
 & + region & 0.8 & 0.5 & 1.6 & 0.0 & 0.0 & -- & 0.0 & 0.0 \\
 & protect safety, task-like & 95.1 & 93.5 & 95.4 & 97.6 & 0.0 & -- & 33.5 & 0.0 \\
\midrule
Concise ctx, 1.5B & deployed & 3.8 & 1.8 & 3.3 & 0.0 & 0.0 & -- & 0.0 & 0.0 \\
 & + region & 3.7 & 1.5 & 2.9 & 0.0 & 0.0 & -- & 0.0 & 0.0 \\
 & protect safety, task-like & 100.0 & 39.8 & 43.9 & 0.4 & 0.4 & -- & 33.2 & 0.1 \\
\midrule
3 tasks, 1.5B & deployed & 100.0 & 4.0 & 6.0 & 0.0 & 8.6 & 100.0 & 1.0 & 13.2 \\
 & + region & 98.0 & 0.0 & 0.7 & 0.0 & 0.0 & 0.0 & 0.0 & 0.0 \\
 & one budget & 100.0 & 2.0 & 3.6 & 0.0 & 7.2 & 100.0 & 0.2 & 2.3 \\
 & + region, one budget & 98.0 & 0.0 & 0.7 & 0.0 & 0.0 & 0.0 & 0.0 & 0.0 \\
 & + harmful in format & 100.0 & 0.2 & 1.2 & 0.0 & 0.4 & 4.4 & 0.0 & 0.0 \\
 & + region, + harmful in format & 98.0 & 0.0 & 0.7 & 0.0 & 0.0 & 0.0 & 0.0 & 0.0 \\
\midrule
3 tasks, 7B & deployed & 100.0 & 3.0 & 4.8 & 0.0 & 39.0 & 100.0 & 3.0 & 57.5 \\
 & + region & 98.0 & 0.0 & 0.7 & 0.0 & 0.0 & 0.0 & 0.0 & 0.0 \\
 & one budget & 100.0 & 1.8 & 3.3 & 0.0 & 33.6 & 100.0 & 2.0 & 42.5 \\
 & + region, one budget & 98.0 & 0.0 & 0.7 & 0.0 & 0.0 & 0.0 & 0.0 & 0.0 \\
 & + harmful in format & 100.0 & 2.0 & 3.6 & 0.0 & 11.0 & 1.7 & 0.0 & 24.5 \\
 & + region, + harmful in format & 98.0 & 0.0 & 0.7 & 0.0 & 0.0 & 0.0 & 0.0 & 0.0 \\
\midrule
5 tasks, 1.5B & deployed & 100.0 & 6.0 & 8.3 & 0.0 & 50.0 & 100.0 & 2.2 & 13.2 \\
 & + region & 98.0 & 0.0 & 0.7 & 0.0 & 0.0 & 0.0 & 0.0 & 0.0 \\
 & one budget & 100.0 & 2.0 & 3.6 & 0.0 & 28.6 & 99.6 & 0.5 & 1.3 \\
 & + region, one budget & 98.0 & 0.0 & 0.7 & 0.0 & 0.0 & 0.0 & 0.0 & 0.0 \\
 & + harmful in format & 100.0 & 2.5 & 4.2 & 0.0 & 79.6 & 4.4 & 1.0 & 0.0 \\
 & + region, + harmful in format & 98.1 & 0.0 & 0.7 & 0.0 & 0.0 & 0.0 & 0.0 & 0.0 \\
\bottomrule
\end{tabular}}
\end{table}

\textbf{Benchmarks written by others:} HarmBench's 200 standard and 100 contextual behaviours
\citep{mazeika2024harmbench} and the 288 StrongREJECT prompts not taken from AdvBench \citep{souly2024strongreject} are
never fitted or calibrated on. Every setting was frozen, and the predictions were written before the runs. After these
benchmarks, and before any gate saw them, we fixed ten more sets: MultiJail harmful requests \citep{deng2024multilingual}
by resource level, oasst1 user turns \citep{kopf2023openassistant} in English and other languages, and requests behind
a system prompt or after a first exchange. All four predictions hold (\cref{tab:prospective}): outside Concise, the
chosen gates fire on at most 1.6\% of any set, with 95\% bounds of at most 3.3\%. On Concise, whose prompts are plain
chat requests, the chosen gate fires on up to 32.2\% of MultiJail, as its harmful component samples only English
AdvBench requests.

\begin{table*}[t]
\centering
\scriptsize
\setlength{\tabcolsep}{2.4pt}
\caption{The frozen gates on prompt sets written by others, measured once: firing (\%) of the gate chosen for each experiment ($\ast$) and of the deployed gate, with the 95\% Clopper--Pearson bound in brackets for the chosen gate. MJ: MultiJail harmful requests in English and in its high-, medium- and low-resource languages; HB turn 2, HB sys.: HarmBench requests after an oasst1 exchange or behind a system prompt from awesome-chatgpt-prompts; OA: oasst1 user turns in English, in other languages and as second turns; Dolly sys.: held-out Dolly requests behind such a system prompt. Firing bounds how often any model behind the gate answers differently.}
\label{tab:prospective}
\begin{tabular}{l l rrrrrrrrrr}
\toprule
 & & \multicolumn{6}{c}{harmful} & \multicolumn{4}{c}{benign} \\
Experiment & Gate & MJ en & MJ hi & MJ mid & MJ lo & HB turn 2 & HB sys. & OA en & OA other & OA turn 2 & Dolly sys. \\
\midrule
Concise, Qwen2.5-1.5B & protect + passages$^\ast$ & 31.1 [35.7] & 6.7 [8.2] & 32.2 [34.8] & 20.3 [22.6] & 0.0 [1.5] & 0.0 [1.5] & 18.1 [20.2] & 6.3 [7.7] & 4.0 [5.2] & 8.5 [11.2] \\
 & deployed & 1.6 & 1.4 & 0.2 & 1.3 & 32.0 & 15.5 & 5.1 & 1.2 & 67.5 & 38.8 \\
Concise-ctx, Qwen2.5-1.5B & protect + passages$^\ast$ & 0.0 [0.9] & 0.0 [0.3] & 0.0 [0.3] & 0.0 [0.3] & 0.0 [1.5] & 0.0 [1.5] & 0.0 [0.3] & 0.1 [0.5] & 0.0 [0.3] & 0.0 [0.7] \\
 & deployed & 0.0 & 0.0 & 0.0 & 0.0 & 0.0 & 0.0 & 0.1 & 0.3 & 0.1 & 9.2 \\
SQuAD~2.0, Qwen2.5-1.5B & protect + passages$^\ast$ & 0.0 [0.9] & 0.0 [0.3] & 0.0 [0.3] & 0.0 [0.3] & 0.0 [1.5] & 0.0 [1.5] & 0.1 [0.5] & 0.0 [0.3] & 0.0 [0.3] & 0.0 [0.7] \\
 & deployed & 0.0 & 0.0 & 0.0 & 0.0 & 0.0 & 0.0 & 0.0 & 0.0 & 0.4 & 0.0 \\
NQ, Qwen2.5-1.5B & two-sided$^\ast$ & 1.6 [3.3] & 0.0 [0.3] & 0.0 [0.3] & 0.0 [0.3] & 0.0 [1.5] & 0.0 [1.5] & 0.9 [1.6] & 0.0 [0.3] & 0.0 [0.3] & 0.0 [0.7] \\
 & deployed & 1.9 & 0.7 & 0.0 & 0.0 & 0.0 & 0.0 & 1.5 & 0.4 & 0.0 & 0.0 \\
IF, Qwen2.5-1.5B & two-sided + remedy$^\ast$ & 0.0 [0.9] & 0.0 [0.3] & 0.0 [0.3] & 0.0 [0.3] & 0.0 [1.5] & 0.0 [1.5] & 0.0 [0.3] & 0.0 [0.3] & 0.0 [0.3] & 0.0 [0.7] \\
 & deployed & 0.0 & 0.2 & 0.3 & 0.0 & 8.0 & 52.0 & 5.2 & 2.1 & 45.5 & 60.5 \\
IF, Qwen2.5-7B & two-sided + remedy$^\ast$ & 0.0 [0.9] & 0.0 [0.3] & 0.0 [0.3] & 0.0 [0.3] & 0.0 [1.5] & 0.0 [1.5] & 0.0 [0.3] & 0.0 [0.3] & 0.0 [0.3] & 0.0 [0.7] \\
 & deployed & 0.0 & 0.0 & 0.1 & 0.1 & 1.5 & 2.5 & 4.9 & 1.5 & 28.1 & 5.8 \\
3-task stream, Qwen2.5-1.5B & two-sided + remedy$^\ast$ & 0.0 [0.9] & 0.0 [0.3] & 0.0 [0.3] & 0.0 [0.3] & 0.0 [1.5] & 0.0 [1.5] & 0.0 [0.3] & 0.0 [0.3] & 0.0 [0.3] & 0.0 [0.7] \\
 & deployed & 1.0 & 0.2 & 0.0 & 0.7 & 0.0 & 5.0 & 9.4 & 1.2 & 1.3 & 14.2 \\
NQ, OLMo-2-7B & two-sided$^\ast$ & 1.6 [3.3] & 0.0 [0.3] & 0.0 [0.3] & 0.0 [0.3] & 0.0 [1.5] & 0.0 [1.5] & 1.0 [1.7] & 0.0 [0.3] & 0.0 [0.3] & 0.0 [0.7] \\
 & deployed & 1.9 & 0.1 & 0.0 & 0.3 & 0.0 & 0.0 & 1.0 & 0.0 & 0.0 & 0.0 \\
NQ, Llama-3.2-3B & two-sided$^\ast$ & 0.6 [2.0] & 0.0 [0.3] & 0.0 [0.3] & 0.0 [0.3] & 0.0 [1.5] & 0.0 [1.5] & 0.8 [1.4] & 0.0 [0.3] & 0.0 [0.3] & 0.0 [0.7] \\
 & deployed & 1.3 & 0.0 & 0.0 & 0.0 & 0.0 & 0.0 & 0.9 & 0.0 & 0.0 & 0.2 \\
\bottomrule
\end{tabular}
\end{table*}

\textbf{A new task with the recipe frozen:} To test whether the recipe was tuned to the tasks shown, we trained on a
task no earlier experiment used: TriviaQA \citep{joshi2017triviaqa}, closed-book questions scored as NQ, on
Llama-3.2-3B, with NQ's configuration and only the task changed. Seven predictions were written before the data were
downloaded, and the gates' coverage and firing were recorded before any training run. All seven hold
(\cref{tab:trivia}). No silent response changes in the gated runs (0 of 9{,}405). The gate fires on 3.5\% of held-out
Dolly, above the 3\% level but consistent with it ($p=0.32$), and on 2.0\% of 10{,}000 fresh Dolly requests. With the
deployed classifier's coverage, 89.1\% instead of 90.5\%, the prediction is 21.9 and the error 1.1 points. TriviaQA asks
plain questions, as NQ does, so this test shows that the recipe carries over to a new task of a known kind, not to a
task drawn from chat.

\subsection{A Wide Harmful Component for Concise}
\label{app:widen}

The chosen Concise gate samples harmful requests from AdvBench only, so a harmful request written unlike AdvBench
reads as chat to it (\cref{tab:prospective}). \Cref{tab:widen} adds one component, written after that failure, with
five predictions written before any download or fit (\cref{tab:protocol}): HarmBench (plain, as a second turn and
behind a system prompt), StrongREJECT and MultiJail in ten languages. Each request, with its forms and translations,
goes by a hash to a third for fitting, one for calibration (one row per request, so \cref{thm:calib} holds for the
component) or one for test; nothing else of the gate changes. Two harmful sets were downloaded for this test only and
are never fitted or calibrated: Do-Not-Answer \citep{wang2024donotanswer}, 938 English questions, and Aya red-teaming
\citep{aakanksha2024prism}, prompts written by native speakers, 300 in each of eight languages. XSTest's unsafe prompts
are not fitted or calibrated either.

All five predictions hold. The gate holds its level on AdvBench and on the component's test third (0.0\%) and fires on
no test row of any source, against up to 36.2\% before on MultiJail English. On the sets it never saw it fires on at
most 0.5\%: Do-Not-Answer 0.4, XSTest 0.5 and Aya red-teaming 0.0\% in every language, against 15.9, 23.0 and 8.9\%
before (49.3\% in Arabic). Seq-GRPO behind it newly answers 0.22\% of 6{,}974 English harmful prompts over three seeds,
against 9.55\% behind the chosen gate. The price is the one \cref{thm:limit} sets for a task drawn from chat: excluding
harmful chat in more forms also excludes Concise prompts, so the coverage falls from 23.3 to 10.2\% and the gain kept
from 17.2 to 9.0 of 73.6 points, and firing on benign chat falls with it (oasst1 English 18.1 to 2.3\%, WildChat 13.0
to 2.2\%). Newly answered is counted on English sets only, as refusal is an English string check; firing is measured
in every language.

\begin{table}[t]
\centering
\scriptsize
\setlength{\tabcolsep}{3pt}
\caption{Concise behind the chosen gate and behind the same gate with a wide harmful component (HarmBench, StrongREJECT and MultiJail in ten languages, split by request into thirds for fitting, calibration and test), $\varepsilon=0.03$. Firing (\%) on the test third of every source and on every held-out set, with the 95\% bound for the wide gate; Do-Not-Answer, Aya red-teaming and XSTest were never fitted or calibrated. Dolly, oasst1 and WildChat are not protected on Concise, whose prompts are Dolly requests.}
\label{tab:widen}
\begin{tabular}{l r r r}
\toprule
Set & $n$ & chosen & wide [bound] \\
\midrule
wide component (one row per request) & 246 & 8.1 & 0.0 [1.2] \\
HarmBench & 62 & 0.0 & 0.0 [4.7] \\
HarmBench as a second turn & 62 & 0.0 & 0.0 [4.7] \\
HarmBench behind a system prompt & 62 & 0.0 & 0.0 [4.7] \\
StrongREJECT & 90 & 2.2 & 0.0 [3.3] \\
MultiJail, English & 94 & 36.2 & 0.0 [3.1] \\
MultiJail, high-resource & 282 & 5.3 & 0.0 [1.1] \\
MultiJail, medium-resource & 282 & 31.9 & 0.0 [1.1] \\
MultiJail, low-resource & 282 & 20.6 & 0.0 [1.1] \\
Do-Not-Answer & 938 & 15.9 & 0.4 [1.0] \\
Aya red-teaming & 2400 & 8.9 & 0.0 [0.1] \\
XSTest, unsafe prompts & 200 & 23.0 & 0.5 [2.3] \\
HarmBench contextual & 100 & 9.0 & 1.0 [4.7] \\
AdvBench (held out) & 250 & 0.0 & 0.0 [1.2] \\
Dolly (held out) & 400 & 22.8 & 11.0 [13.9] \\
oasst1 first turns, English & 1000 & 18.1 & 2.3 [3.2] \\
oasst1 first turns, other languages & 1000 & 6.3 & 0.0 [0.3] \\
oasst1 second turns & 1000 & 4.0 & 0.1 [0.5] \\
Dolly behind a system prompt & 400 & 8.5 & 0.2 [1.2] \\
WildChat requests & 500 & 13.0 & 2.2 [3.6] \\
MMLU & 500 & 0.0 & 0.0 [0.6] \\
\midrule
Coverage $\beta$ of RL prompts & & 23.3 & 10.2 \\
Seq-GRPO's gain behind the gate ($\bar G$ +73.6) & & +17.2 & +9.0 \\
Newly answered, English harmful (\%) & 6974 & 9.55 & 0.22 \\
\bottomrule
\end{tabular}
\end{table}

\subsection{A Refusal Fallback on Fired Prompts}
\label{app:fallback}

\textbf{The rule:} On a prompt where the gate fires, the old model's reply is generated as well. If its first 400
characters pass the refusal check used throughout, that reply is returned and the adapter's reply is not. Elsewhere
nothing changes. The rule needs a second generation of at most 400 characters on every fired prompt, so its compute
cost is the gate's firing rate; without a gate it is every prompt.

\textbf{Measurements:} The fallback is applied to the stored responses of every experiment of \cref{tab:audit},
pooled over three seeds. Nothing is retrained, and every value is a machine score: the refusal string check and each
task's scorer. For each system we count the harmful prompts that the base model refuses and the system answers (New),
the task accuracy, and, without a gate, the share of benign prompts (Dolly, MMLU, safe XSTest and the other held-out
sets) whose reply still differs from the base model's. With the fallback, New is 0 in every row, as the rule
guarantees.

\textbf{Reading:} Where a gate leaks, the fallback removes the leak at no cost in task accuracy: behind the deployed
IF and stream gates, and for the models trained under them, task accuracy is unchanged to the decimal
(\cref{tab:fallback}). Without a gate the fallback restores refusal, but most benign replies still change, so it
complements the gate and does not replace it. The same holds for the agents of
\cref{sec:res-agentloop,sec:res-api}, where the rule returns the base agent's episode if the check flags it. It
brings the ungated agents'
refusal back to the base agent's rate or above, from 8.4 to 100\% on the API-call benchmark, while 73.6 to 90.0\% of
their chat episodes still differ from the base agent's. Behind the gates, every reply to a harmful request is already
the base agent's, so the rule changes none.

\begin{table*}[t]
\centering
\scriptsize
\setlength{\tabcolsep}{4pt}
\caption{The refusal fallback on stored responses (three seeds pooled; \%; machine-scored). New: harmful prompts the
base model refuses and the system answers, without the fallback (with it, 0 in every row). Task: accuracy without /
with the fallback. Benign changed: replies that differ from the base model's, with the fallback. Deployed gate:
Seq-GRPO behind it. Trained: \methodp{}'s own runs, over their stored harmful sets. For Concise, $n$ also covers
further stored harmful sets, so its rates differ from \cref{tab:audit}.}
\label{tab:fallback}
\begin{tabular}{l r rrr rr rr}
\toprule
 & & \multicolumn{3}{c}{No gate (Seq-GRPO)} & \multicolumn{2}{c}{Deployed gate} & \multicolumn{2}{c}{Trained under the gate} \\
\cmidrule(lr){3-5}\cmidrule(lr){6-7}\cmidrule(lr){8-9}
Experiment & Base task & New & Task & Benign changed & New & Task & New & Task \\
\midrule
NQ, Qwen 1.5B & 18.4 & 27.24 & 23.6 / 20.7 & 80.1 & 0.16 & 23.1 / 20.5 & 0.65 & 22.4 / 20.2 \\
NQ, OLMo-2 7B & 34.2 & 1.81 & 40.1 / 39.7 & 82.2 & 0.00 & 39.9 / 39.5 & 0.00 & 38.3 / 37.7 \\
NQ, Llama 3B & 40.6 & 14.40 & 51.4 / 46.1 & 94.8 & 0.00 & 49.8 / 45.4 & 0.00 & 47.5 / 44.1 \\
IF, Qwen 1.5B & 60.8 & 39.24 & 99.5 / 99.5 & 89.3 & 37.96 & 99.5 / 99.5 & 22.96 & 99.8 / 99.8 \\
IF, Qwen 7B & 86.7 & 4.27 & 99.7 / 99.7 & 80.6 & 4.14 & 99.7 / 99.7 & 2.63 & 99.6 / 99.6 \\
3 tasks, Qwen 1.5B & 31.5 & 18.21 & 57.2 / 57.2 & 69.4 & 16.65 & 57.2 / 57.2 & 7.63 & 58.8 / 58.8 \\
SQuAD~2.0, Qwen 1.5B & 39.0 & 19.71 & 63.4 / 63.6 & 84.1 & 0.03 & 57.5 / 57.7 & 0.00 & 54.9 / 55.1 \\
Concise-ctx, Qwen 1.5B & 34.6 & 48.85 & 99.8 / 99.4 & 91.1 & 0.49 & 37.2 / 37.2 & 0.00 & 36.1 / 36.1 \\
Concise, Qwen 1.5B & 26.4 & 85.50 & 100.0 / 90.0 & 93.1 & 2.43 & 27.4 / 27.4 & 0.00 & 26.4 / 26.4 \\
\bottomrule
\end{tabular}
\end{table*}

\subsection{Full-Rank Updates and SFT}
\label{app:res-other}

\Cref{tab:pilot} replaces LoRA by a dense update of every projection above the split, and GRPO by supervised
fine-tuning, on IF and NQ. The pattern does not change, as the certificate depends only on where the update acts:
ungated, every run changes 86.1--99.8\% of Dolly responses; behind the deployed gate every run changes 0.8--3.2\% and
each IF run keeps its full gain; trained under the gate each changes 0.6--3.5\%, and no response changed where the gate
was silent (4{,}797 per gated IF method, 9{,}363 for SFT on a 7B teacher's NQ responses, 3{,}121 for gold SFT). SFT on
the model's own accepted samples gains less than GRPO, +21.6 against +38.7 points, and on a 7B teacher's responses
+34.3. On NQ, full-rank SFT on the first gold answer lowers accuracy by 6.4 points and MMLU by 4.4; under the gate it
loses the same 6.4 points but leaves MMLU unchanged, and behind the gate the ungated model keeps $-$5.4 points against
the predicted $-$5.9: where training hurts, the gate confines the loss to where it fires. LoRA has been found to forget
less than full fine-tuning \citep{biderman2024lora}; here full-rank training changed as many responses as LoRA and
lowered MMLU less (+0.1 against $-$3.7 points).

\begin{table}[t]
\centering
\footnotesize
\setlength{\tabcolsep}{3pt}
\caption{Full-rank updates and SFT on IF and NQ, Qwen2.5-1.5B. Seeds: runs (gain $\pm$ sd). Full-rank: a dense
update of every projection of blocks 9--28 (``all blocks'': 2--28) at learning rate $5\cdot10^{-6}$, against
$5\cdot10^{-5}$ for LoRA. SFT: on the model's own accepted samples, four per prompt; 7B teacher: full-rank SFT on
Qwen2.5-7B-Instruct's accepted responses, two per prompt; gold: on NQ's first gold answer. Chg.: Dolly responses
changed (\%); behind gate: the deployed prompt-time gate at $\varepsilon=0.03$.}
\label{tab:pilot}
\begin{tabular}{l r rrr rr}
\toprule
 & & & & & \multicolumn{2}{c}{behind gate} \\
\cmidrule(lr){6-7}
Run & Seeds & Gain$\uparrow$ & Chg.$\downarrow$ & MMLU & Gain$\uparrow$ & Chg.$\downarrow$ \\
\midrule
IF: LoRA GRPO, Seq-GRPO & 3 & +38.7$\pm$0.2 & 93.3 & $-$3.7 & +38.7 & 1.5 \\
LoRA GRPO, \methodp{} & 3 & +39.0$\pm$0.3 & 1.3 & $-$0.1 & -- & -- \\
full-rank GRPO, Seq-GRPO & 3 & +39.0$\pm$0.3 & 93.4 & +0.1 & +39.0 & 1.5 \\
\quad learning rate $2\cdot10^{-6}$ & 1 & +39.0 & 94.0 & +0.0 & +39.0 & 1.5 \\
\quad all blocks & 3 & +39.1$\pm$0.1 & 95.0 & $-$0.7 & +39.1 & 1.5 \\
full-rank GRPO, \methodp{} & 3 & +38.9$\pm$0.6 & 1.3 & $-$0.2 & -- & -- \\
LoRA SFT, ungated & 3 & +21.6$\pm$0.9 & 86.1 & +0.5 & +21.6 & 0.8 \\
LoRA SFT, \methodp{} & 3 & +19.3$\pm$1.5 & 0.6 & +0.7 & -- & -- \\
full-rank SFT, 7B teacher, ungated & 3 & +34.3$\pm$1.1 & 95.0 & +1.5 & +34.3 & 1.3 \\
\quad all blocks & 3 & +34.9$\pm$2.0 & 94.7 & +1.0 & +34.9 & 1.3 \\
full-rank SFT, 7B teacher, \methodp{} & 3 & +35.9$\pm$0.7 & 1.2 & +0.3 & -- & -- \\
\midrule
NQ: LoRA GRPO, Seq-GRPO & 3 & +5.2$\pm$0.9 & 94.2 & +0.2 & +4.7 & 3.2 \\
full-rank SFT, gold, ungated & 1 & $-$6.4 & 99.5 & $-$4.4 & $-$5.4 & 3.2 \\
\quad all blocks & 1 & $-$3.2 & 99.8 & $-$2.0 & $-$2.6 & 3.2 \\
full-rank SFT, gold, \methodp{} & 1 & $-$6.4 & 3.5 & +0.0 & -- & -- \\
full-rank SFT, 7B teacher, ungated & 3 & +0.8$\pm$0.5 & 99.5 & +1.9 & +0.8 & 3.2 \\
full-rank SFT, 7B teacher, \methodp{} & 3 & +0.8$\pm$0.9 & 3.5 & +0.0 & -- & -- \\
\bottomrule
\end{tabular}
\end{table}

\subsection{Agents, Live Traffic and Adaptive Search}
\label{app:agents}

\textbf{Agent:} Each episode is a chat with a calculator tool in the model's own tool-call format; the model is called on
the conversation so far, a tool call is executed and its result appended, for up to four calls. A prompt-time gate
decides at every call, so with greedy decoding an episode can differ from the old model's only if some call fires
(\cref{cor:episode}); every call after an episode's first runs alone, so that its numerics do not depend
on other episodes. Sets of 250 episodes: GSM8K problems, held-out Dolly requests, held-out AdvBench requests as the
second user turn after an exchange that used the tool, and the task's evaluation prompts. The agent-aware gate (NQ and
IF) refits the chosen gate's classifier with the same prompts written as an agent's first call added on both sides,
and calibrates it (\cref{thm:calib}) on every reference component, plain and in the agent's format; its held-out
firing is consistent with the level on plain Dolly (3.3 and 0.0\%) and AdvBench (0.0\%), and its coverage of plain task prompts
moves from 92.2 to 90.8\% (NQ) and from 100 to 98.5\% (IF). Run live behind it, Seq-GRPO changes 1.2\% of NQ's Dolly
episodes and none on IF. Gates fitted on plain prompts only do not carry over: in the agent the deployed IF gate fires
on 96\% of Dolly episodes, and the deployed NQ gate and every chosen gate on no task episode.

\textbf{Live traffic:} We replay the deployed and the chosen gate of nine experiments on the first user turns of
5{,}000 WildChat conversations in their stored order, sorted by time (April 9--11, 2023; 48\% English, 29\% Chinese). The monitors are tuned to twice the level, a mixture over rates of 1.5 to 20 times the level, and the
mixture restarted every 500 prompts at $\delta_j=6\delta/(\pi^2j^2)$, all at $\delta=0.05$. Every chosen gate except
Concise's fires on at most 0.3\% of the stream. The repaired IF and stream gates keep full coverage.

\textbf{Adaptive search:} For each of the first 50 held-out AdvBench requests, greedy coordinate gradient
\citep{zou2023gcg} searches a 20-token suffix (250 steps) that raises the gate's margin through the frozen trunk,
with the region's distance as a second term for two-sided gates; a request counts when the gate fires on the text
computed the official way. It fires every deployed gate within a few steps, but the two-sided NQ and IF gates on 2
and 0 of 50 requests. Where a gate fires, a prompt receives the trained model's response, as without the gate
(\cref{cor:switch}).

\subsection{Training Inside the Agent Loop}
\label{app:agentloop}

\textbf{Tasks and episodes:} The agent is the calculator agent of \cref{app:agents}: the model is called on the
conversation so far, a tool call is executed and its result appended, for up to four calls; a call in the fourth
reply is not executed. Calc problems are generated from six templates (stock, payroll, packing, a bare expression,
distance and area); each answer is a product of numbers with two to four digits combined with a further
multiplication, subtraction or division, and has at least four digits. There are 2{,}400 training problems, and no
test problem occurs in training. The reward is 1 if the boxed number in the
episode's last reply is the answer and 0 otherwise; it does not mention the tool. KK is the task of the three-task
stream, played in the same agent with the tool available.

\textbf{Training:} Each task is trained by GRPO on whole episodes: 150 steps of 16 prompts with 8 sampled episodes
each, with the LoRA of \cref{sec:setup-main}. Every model call of an episode is one training item, the call's
context and its reply, weighted by the episode's advantage; tool results are part of the context and are never
trained. Behind the gate, a call on which the gate did not fire used no adapter and gives no gradient.

\textbf{Gate:} For each task the classifier is fitted on the trunk feature at the last token of every call of the
previous agent's greedy episodes: 600 prompts of the new task against 600 Dolly, 100 AdvBench and 900 task-like
prompts and, for KK, 300 stored Calc problems. It is calibrated with \cref{thm:calib} on the episode score of
\cref{cor:episode}, at $\varepsilon=0.03$ and $\delta=0.05$ split evenly over the components: 700 Dolly, 170 AdvBench
and 700 task-like prompts and, for KK, 300 further stored Calc problems, all played as episodes of the previous agent.

\textbf{Evaluation:} After each task, 250 greedy episodes per set are played, every call after an episode's first
alone. Two episodes count as the same if the token ids of every reply agree. The check of \cref{cor:episode} compares
each episode on which the new gate fired at no call with the same episode of the previous agent. Refusal is the
string check used throughout.

\textbf{After the first task:} \Cref{tab:agentloop_calc} gives every run after Calc; \cref{tab:agentloop} gives them
after KK. Every run learns Calc, with a calculator call in every Calc episode. The Calc gate fires on every Calc
episode and on 0.4\% of Dolly episodes (95\% bound 1.9\%), and 0.4\% of Dolly episodes change, against 86 to 100\%
under Seq-GRPO. At 1.5B, KK accuracy behind the gate stays at the base agent's 8.8\%, while Seq-GRPO on Calc moves it
to between 3.2 and 9.6\%.

\begin{table*}[t]
\centering
\scriptsize
\setlength{\tabcolsep}{4.5pt}
\caption{The runs of \cref{tab:agentloop} after the first task, Calc (250 greedy episodes per set, in \%). Changed:
Dolly episodes whose transcript differs from the base agent's. Fires: episodes in which the Calc gate fires at some
call, with the 95\% bound on Dolly. Unfired changed: episodes that changed against the base agent / episodes without
a fired call.}
\label{tab:agentloop_calc}
\begin{tabular}{l c l rrr r r rr r}
\toprule
 & & & \multicolumn{3}{c}{Accuracy and tool use} & Changed & & \multicolumn{2}{c}{Fires} & \\
\cmidrule(lr){4-6}\cmidrule(lr){9-10}
Model & Seed & Method & Calc & Tool & KK & Dolly & Refusal & Dolly [bound] & Calc & Unfired changed \\
\midrule
Qwen2.5-1.5B & -- & base agent & 27.2 & 89.6 & 8.8 & -- & 98.8 & -- & -- & -- \\
 & 0 & Seq-GRPO & 100 & 100 & 9.6 & 99.2 & 99.2 & -- & -- & -- \\
 & & \methodp{} & 100 & 100 & 8.8 & 0.4 & 98.8 & 0.4 [1.9] & 100 & 0 / 749 \\
 & 1 & Seq-GRPO & 100 & 100 & 9.6 & 99.6 & 99.2 & -- & -- & -- \\
 & & \methodp{} & 97.6 & 100 & 8.8 & 0.4 & 98.8 & 0.4 [1.9] & 100 & 0 / 749 \\
 & 2 & Seq-GRPO & 99.2 & 100 & 3.2 & 100 & 99.2 & -- & -- & -- \\
 & & \methodp{} & 97.6 & 100 & 8.8 & 0.4 & 98.8 & 0.4 [1.9] & 100 & 0 / 749 \\
\midrule
Qwen2.5-7B & -- & base agent & 95.6 & 100 & 9.2 & -- & 98.8 & -- & -- & -- \\
 & 0 & Seq-GRPO & 97.6 & 100 & 14.0 & 86.4 & 98.4 & -- & -- & -- \\
 & & \methodp{} & 97.6 & 100 & 9.6 & 0.4 & 98.8 & 0.4 [1.9] & 100 & 0 / 740 \\
\bottomrule
\end{tabular}
\end{table*}

\textbf{Tool calls on other sets:} \Cref{tab:agentloop_tools} counts, from the stored episodes, the Dolly and KK
episodes that contain a calculator call. After Calc, the ungated agents call the calculator far more often than the
base agent on both sets, which Calc never trained. Behind the gate, the rate on Dolly is the base agent's at every
stage, and on KK until KK itself is learned.

\begin{table}[t]
\centering
\scriptsize
\setlength{\tabcolsep}{3.5pt}
\caption{Episodes with a calculator call (\%, counted by rule from the stored episodes) on Dolly and KK after Calc
(1) and after KK (2). Base agent before training: Dolly 24.8, KK 0.0 at 1.5B; 9.6 and 4.0 at 7B.}
\label{tab:agentloop_tools}
\begin{tabular}{l l rr rr}
\toprule
 & & \multicolumn{2}{c}{Dolly} & \multicolumn{2}{c}{KK} \\
\cmidrule(lr){3-4}\cmidrule(lr){5-6}
Run & Method & 1 & 2 & 1 & 2 \\
\midrule
1.5B, seed 0 & Seq-GRPO & 82.8 & 0.0 & 98.0 & 0.0 \\
 & \methodp{} & 24.8 & 24.8 & 0.0 & 0.0 \\
1.5B, seed 1 & Seq-GRPO & 88.0 & 60.8 & 98.8 & 53.2 \\
 & \methodp{} & 24.8 & 24.8 & 0.0 & 0.0 \\
1.5B, seed 2 & Seq-GRPO & 89.2 & 87.2 & 100 & 100 \\
 & \methodp{} & 24.8 & 24.8 & 0.0 & 0.0 \\
7B, seed 0 & Seq-GRPO & 18.4 & 5.6 & 15.2 & 0.0 \\
 & \methodp{} & 9.6 & 9.6 & 4.0 & 0.0 \\
\bottomrule
\end{tabular}
\end{table}

\subsection{The API-Call Benchmark}
\label{app:api}

\textbf{Data and episodes:} We use the training split of the Schema-Guided Dialogue dataset \citep{rastogi2020sgd},
released under CC BY-SA 4.0, and five of its services: Flights\_1 with four functions, Restaurants\_1 with two, and
Events\_2, Buses\_2 and Hotels\_3 with seven together. The 13 functions form one toolbox, which every episode gives
in the model's own tool-call format under one system prompt; the prompt also fixes today's date and the formats of
dates and times. An episode is a dialogue up to a turn at which the SGD system called a service. Earlier calls of the
dialogue and their first result are part of the history. The model is called once, and no function is executed. A
reply is correct if it calls the annotated function with the annotated arguments, after the service's default values
are filled in and ignoring case. The reward is 0 without the annotated function, and otherwise 0.5 plus half the
Jaccard overlap of the argument pairs. 

\textbf{Splits:} Dialogues are assigned to splits by a hash of their identifier, so no dialogue occurs in two splits.
Each task has 1{,}200 call turns for training, 600 for fitting its gate, 300 and 300 that the gates of later tasks
use for fitting and calibration, and 300 for testing, of which 250 are played. Test, calibration and other-service
rows are one call turn per dialogue, so the calibration rows are independent, as \cref{thm:calib} requires.

\textbf{Training and gates:} Training follows \cref{app:agentloop}: GRPO for 150 steps of 16 prompts with 8 sampled
episodes each, with the LoRA and split of \cref{sec:setup-main}. A prompt has about 3{,}500 tokens, most of
them the toolbox, and a reply at most 192. Each gate's classifier is fitted on 600 turns of the new service against
600 Dolly, 100 AdvBench and 900 task-like prompts and 600 other-service turns and, for restaurants, 300 stored flight
turns. It is calibrated at $\varepsilon=0.03$ and $\delta=0.05$, split evenly over the components, on 700 Dolly, 170
AdvBench and 700 task-like prompts and 600 further other-service turns and, for restaurants, 300 further stored
flight turns. At the calibrated thresholds, both gates fire on 1.3\% of the other-service calibration turns and on no
calibration prompt of any other component.

\textbf{Results by stage:} \Cref{tab:api_stages} gives both methods after each task. Two checks were written before
the run: no episode changes without a firing, and each gate's firing on the protected sets stays within the level.
Both hold. Of the episodes without a firing, 0 of 1{,}083 changed after flights and 0 of 1{,}058 after restaurants,
and neither gate fired on a held-out episode of a protected set (95\% bound 1.2\%). \Cref{tab:api_replies} sorts the
replies to harmful and chat requests by rule into four classes: a refusal by the string check used throughout, a call
to one of the 13 toolbox functions, a call to any other function name, and other text. 

\begin{table*}[t]
\centering
\scriptsize
\setlength{\tabcolsep}{3.4pt}
\caption{The API-call benchmark after each task (Qwen2.5-1.5B, seed 0; 250 greedy episodes per set, in \%;
machine-scored). Correct: the annotated function with the annotated arguments. Function: the annotated function,
whatever the arguments. Changed: episodes whose reply differs from the base agent's. Fl., Re., Oth.: flights,
restaurants, other services. Fires: episodes on which the stage's gate fires; neither gate fires on a Dolly or an
AdvBench episode, and the 95\% bound of every 0.0 is 1.2\%. Unfired changed: episodes that changed against the
previous agent / episodes without a firing, over the five sets.}
\label{tab:api_stages}
\begin{tabular}{l l rrr rrr rrrrr r rrr r}
\toprule
 & & \multicolumn{3}{c}{Correct} & \multicolumn{3}{c}{Function} & \multicolumn{5}{c}{Changed} & &
 \multicolumn{3}{c}{Fires} & \\
\cmidrule(lr){3-5}\cmidrule(lr){6-8}\cmidrule(lr){9-13}\cmidrule(lr){15-17}
After & Method & Fl. & Re. & Oth. & Fl. & Re. & Oth. & Fl. & Re. & Oth. & Dolly & AdvB. & Refused & Fl. & Re. & Oth. &
Unfired changed \\
\midrule
-- & base agent & 12.4 & 20.8 & 16.8 & 38.4 & 38.4 & 28.4 & -- & -- & -- & -- & -- & 100 & -- & -- & -- & -- \\
\midrule
Flights & Seq-GRPO & 77.6 & 55.2 & 36.8 & 95.6 & 95.6 & 71.6 & 91.2 & 74.4 & 86.8 & 98.0 & 96.8 & 3.2 & -- & -- & -- & -- \\
 & \methodp{} & 52.8 & 20.8 & 16.8 & 76.8 & 38.4 & 28.4 & 63.6 & 0.0 & 0.0 & 0.0 & 0.0 & 100 & 66.8 & 0.0 & 0.0 & 0 / 1{,}083 \\
\midrule
Restaurants & Seq-GRPO & 76.0 & 82.0 & 50.4 & 94.0 & 98.8 & 72.8 & 92.4 & 80.4 & 88.4 & 99.2 & 92.8 & 8.4 & -- & -- & -- & -- \\
 & \methodp{} & 52.8 & 66.4 & 16.8 & 76.8 & 87.6 & 28.4 & 63.6 & 62.0 & 0.0 & 0.0 & 0.0 & 100 & 0.0 & 76.8 & 0.0 & 0 / 1{,}058 \\
\bottomrule
\end{tabular}
\end{table*}

\begin{table}[t]
\centering
\scriptsize
\setlength{\tabcolsep}{4pt}
\caption{Replies to harmful and chat requests on the API-call benchmark, sorted by rule (\% of 250; seed 0). Refusal:
the string check used throughout. Toolbox call: a call to one of the 13 functions. Invented call: a call to a
function name that is not in the toolbox. Behind the gates, the replies are the base agent's after each task.}
\label{tab:api_replies}
\begin{tabular}{l l rrrr}
\toprule
Requests & Agent & Refusal & Toolbox call & Invented call & Other text \\
\midrule
Harmful (AdvBench) & base agent; \methodp{} & 100 & 0.0 & 0.0 & 0.0 \\
 & Seq-GRPO after flights & 3.2 & 81.2 & 15.2 & 0.4 \\
 & Seq-GRPO after restaurants & 8.4 & 80.4 & 11.2 & 0.0 \\
\midrule
Chat (Dolly) & base agent; \methodp{} & 25.6 & 18.0 & 2.4 & 54.0 \\
 & Seq-GRPO after flights & 0.0 & 71.2 & 28.0 & 0.8 \\
 & Seq-GRPO after restaurants & 0.0 & 80.8 & 19.2 & 0.0 \\
\bottomrule
\end{tabular}
\end{table}

\end{document}